\pdfoutput=1
\documentclass{article}
\usepackage{iclr2027_conference,times}

\usepackage[utf8]{inputenc}
\usepackage[T1]{fontenc}
\usepackage{xurl}
\usepackage{hyperref}
\usepackage{url}
\usepackage[htt]{hyphenat}
\usepackage{booktabs}
\usepackage{amsmath}
\usepackage{amssymb}
\usepackage{amsfonts}
\usepackage{graphicx}
\usepackage{microtype}
\usepackage{multirow}
\usepackage{nicefrac}
\usepackage{tabularx}
\usepackage{longtable}
\usepackage{float}
\usepackage{xcolor}
\usepackage{textcomp}
\usepackage{caption}
\usepackage{enumitem}
\usepackage{placeins}

\graphicspath{{figures/}{generated/}}

\title{A minimal probe of task modeling in language models}

\author{
Tianxiang Dai \& Jonathan A. Fan\thanks{Corresponding author.}\\
Department of Electrical Engineering\\
Stanford University\\
Stanford, CA 94305, USA\\
\texttt{\{txdai,jonfan\}@stanford.edu}
}

\newcommand{\preprintcopy}{\iclrfinalcopy\def\preprintheader{Preprint}}
\preprintcopy

\begin{document}

\maketitle
\ifdefined\preprintheader\lhead{\preprintheader}\fi

\begin{abstract}
Language models read long contexts and score well on knowledge and reasoning benchmarks, but neither shows whether a model can do work whose next step depends on everything that came before, as a long-running agent must. We call this ability task modeling and propose counting identical items as its simplest test: every item adds to the total and no item records it. We define counting capacity (CC) as the sequence length up to which a model returns the exact count more often than not, and measure it for 205 models. Every resolved measurement loses count abruptly, at a median of 50 items and at most 627; across release years the median rose 1.6-fold while context windows grew about 80-fold. Changing the counted symbol or the delimiter moves a model's CC severalfold, yet the model accounts for about three quarters of the variance across 16 models and 11 encodings. Reasoning leaves the median CC unchanged across 77 matched pairs. In six open-weight models the count stays readable in early layers past the boundary but does not reach the output, and an activation edit that carries the count forward also carries the state of an object-tracking task. CC tracks the METR task horizon after controlling for general capability, and rises with capacity on grid-execution tasks and, with reasoning off, on object tracking. We do not show that counting is necessary or sufficient for long-horizon work; that remains a hypothesis, which this cheap measurement can test.
\end{abstract}

\section{Introduction}

\begin{figure}[htbp]
\centering
\includegraphics[width=\textwidth]{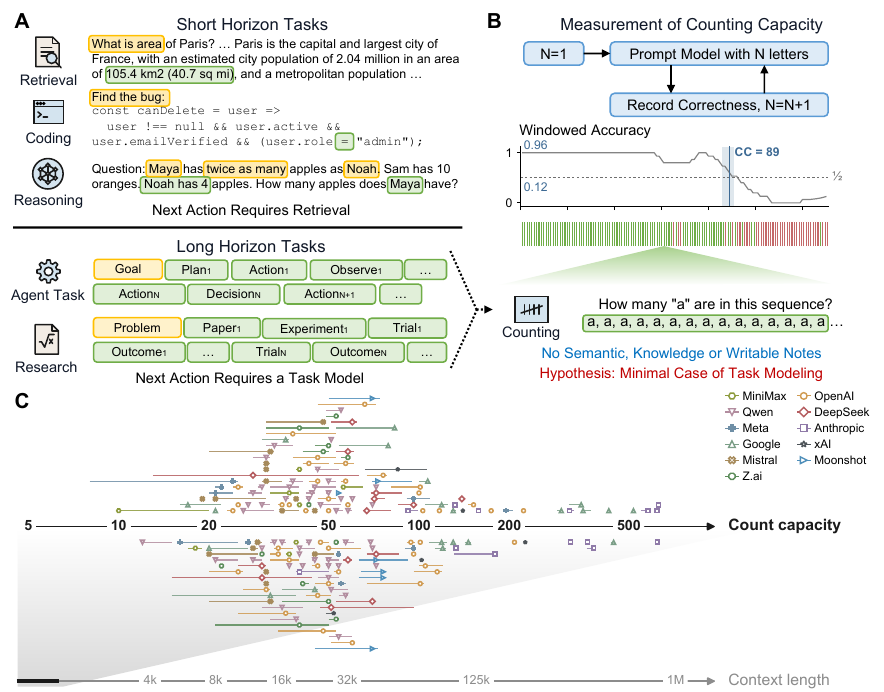}
\caption{\textbf{Counting as a candidate probe of task modeling.}
(\textbf{a}) Illustrative short-horizon tasks (retrieval, coding, a reasoning problem) are resolved by retrieval from a few attended tokens, whereas long-horizon agentic and research tasks depend on a model of the task built from the whole context, to which every step and every prior result contributes and that no token records.
(\textbf{b}) Measurement of CC on the hypothesized minimal case of task modeling, with no semantics, knowledge, or writable notes: the model must return the exact number of items in a homogeneous sequence, every length is probed once in ascending order, and a two-regime fit to the exact/inexact record locates the boundary (CC) at which exact counting stops being the majority behavior (shown: \texttt{claude-sonnet-5}, CC${=}89$).
(\textbf{c}) Fitted CC with its calibrated within-scan interval for every evaluated model (rescans vary more; Appendix~\ref{app:replicates}), with the same scale restated in prompt tokens below and compared with advertised context windows: the median model stops counting after 0.045\% of its window.}
\label{fig:framework}
\end{figure}

\noindent
Large language models (LLMs) work in ways we cannot inspect, and we hope they can execute reliably over long horizons \citep{hai2024repoexec,wu2024longgenbench,yao2022react,metachat}. Long horizon means more than a long input, a long output, or a long chain of steps. Retrieval, writing a paragraph, or following a pointer chain of a thousand links can each be long, but the state they carry at any step is a single thing that can be written down and found again, so the work can be checkpointed and resumed (Fig.~\ref{fig:framework}a). Other work depends on the whole context at once: a research hypothesis drawn from every result obtained, failures included, or Sherlock Holmes naming the one remaining possibility because he kept every eliminated one in view. We call the ability to do such work \emph{task modeling}. By a task model we mean a running summary of the task with two properties: every step so far contributes to it, and no single token in the context records it. In formal terms, a task model is a global aggregate over the context: no sparse subset of the context recovers it, and one built from half the context misses half of what happened. Writing it down as it forms helps only if the context offers a landmark to tie each written value to, because a chain of thought carries what can be named and found again.

Standard benchmarks do not isolate this ability. A score can reflect reasoning, memorized knowledge, or task-specific heuristics \citep{chen2025benchmarking}, and long-horizon suites mix task modeling with many other demands (Section~\ref{sec:related}). The headline benchmarks are also being climbed quickly \citep{phan2025hle,jimenez2024swebench}, and few text tasks remain on which every model fails while a person with a pencil has no limit.

Counting a sequence of identical items is the simplest task with the two properties (Fig.~\ref{fig:framework}b). It has no semantics and needs no knowledge. Every item adds one to the total and no item records the total, and because the items are identical, a note written partway cannot be tied back to the item where it was written. We therefore take counting as a candidate minimal case of task modeling, and we treat that choice as a hypothesis. We do not show that a model that counts well can run an agent, or that one that counts badly cannot; we measure the limit on counting and test the simpler explanations of that limit.

Whether models can count is not settled by theory. Difficulty with repeated tokens is known \citep{yona2025repeatedtoken,fu2024why,barbero2024glasses}, though our sequence interleaves two token types, item and delimiter. In theory a transformer with a chain of thought computes far more than it can in one pass \citep{merrill2024expressive,feng2023towards}, but served models differ from the idealized ones in training, quantization, and architecture. No result predicts or forbids counting at any length, so we measure it. We give the model a string of repeated items and ask how many there are, at every length in ascending order, and score each reply as exactly right or not. The \emph{counting capacity} (CC) is the length at which the exact answer stops being the model's majority behavior. Formally, the replies form a binary record, and CC is the cut \(\tau\) that maximizes the likelihood of two exact-answer rates, \(p_1>\nicefrac{1}{2}\) at lengths up to \(\tau\) and \(p_2<\nicefrac{1}{2}\) beyond it; a boundary is declared only when two rates fit better than one by the Bayesian information criterion. The fit uses the whole record, has no tuned threshold, and comes with a simulation-calibrated interval (Appendix~\ref{app:estimation}). CC is not the first wrong answer, which is an isolated slip in 284 of 296 measurements, and it is conditional on the prompt, encoding, parser, and decoding settings.

Section~\ref{sec:dynamics} measures how far models count and tests simple explanations of the limit. Across 205 LLMs (296 measurements, with reasoning-enabled and reasoning-disabled configurations counted separately; Appendix~\ref{app:catalogue}), every resolved measurement loses count, at a median of 50 items and at best 627 against a ceiling of 3000. Across release-year cohorts the median rose 1.6-fold while the cohorts' median context window grew about 80-fold (Fig.~\ref{fig:framework}c; Table~\ref{tab:cohorts}). Accuracy stays near one and then collapses within a narrow band, at a length that rescans reproduce and that, together with the wrong answers around it, identifies the served model (Appendix~\ref{app:fingerprint}). The wrong answers are structured, with errors of exactly one hundred and preferred round values. Changing the counted item or the delimiter moves one model's capacity severalfold, but across 16 models and 11 encodings the model accounts for about three quarters of the variance. Reasoning gives no median gain across 77 matched pairs, although the same models carry tasks whose running value has a name at loads well beyond their counting capacities. Past a finite length the models cannot copy the sequence either, and a second task carried at the same time lowers CC beyond its controls for one load in each of two models.

Section~\ref{sec:probes} looks for the count inside six open-weight models. A linear readout recovers it at every probed depth. Past the boundary the early-layer readout still rises with the true count, more slowly than before, while the last-layer readout falls. Editing the readable coordinate at the final token does not change the answer; shifting it across the item positions at mid-depth does, in three of the models. Replacing the internal states of the last few items with those of another sequence carries that sequence's count forward, and the same edit applied to an object-tracking history moves the tracked state while the text stays readable.

Section~\ref{sec:evaluations} compares CC with other evaluations. CC correlates with scale but is not explained by it. Among public benchmarks, the METR task horizon is the one whose correlation with CC survives conditioning on general capability; knowledge and coding benchmarks correlate with CC weakly or only through general capability. CC scales with a ladder of grid-execution tasks, and across vendor families it rises with the length at which an object-tracking task fails when reasoning is off; with reasoning on, tracking improves severalfold and counting does not. ARC-AGI-2 scores split into two release cohorts, and serving price follows CC (Appendix~\ref{app:benchmark}). Section~\ref{sec:discussion} states the hypothesis these results leave open and what they do not establish.

\section{Related work}\label{sec:related}

\paragraph{Benchmark evaluation.}
Progress is tracked through suites of knowledge, reasoning, and coding benchmarks \citep{hendrycks2021mmlu,liang2022helm,bigbench2023,rein2023gpqa,jimenez2024swebench,white2024livebench}, increasingly summarized by aggregate indices \citep{epochcapabilities2025} or by the duration of tasks a model can carry to completion \citep{kwa2025metr}.

\paragraph{Long context and counting.}
Long-context evaluation is dominated by retrieval: needle-in-a-haystack probes, multi-hop lookups, and positional-bias studies test whether a model can find sparse landmarks in a long input \citep{hsieh2024ruler,li2025needlebench,bai2024longbench,liu2023lostmiddle}. RULER \citep{hsieh2024ruler} also aggregates word frequencies, but over distinguishable words among distractors, the heterogeneous case that the first task of our ladder covers; none of them measures a quantity to which every token contributes. Counting failures themselves have been noted at the scale of single words and short lists and traced to tokenization \citep{cosma2025strawberry,zhang2024counting,xu2025llm}, repeated-token inputs destabilize attention \citep{yona2025repeatedtoken} and raise letter-counting error \citep{fu2024why}, representational collapse over repeated tokens produces counting and copying errors \citep{barbero2024glasses}, and transformers need positional inductive biases to count beyond their training lengths \citep{chang2024inductive}. Chain-of-thought prompting, test-time scaling, verifier-guided search, and tool use improve a broad range of reasoning tasks \citep{wei2022chain,kojima2022zeroshot,snell2024scaling,lightman2024verify,deepseek2025r1,yao2022react,schick2023toolformer,qin2023toolllmfacilitatinglargelanguage}; Section~\ref{sec:dynamics} asks what they do for this one. Recent studies probe counting mechanistically: \citet{venkatesh2026repeated} finds near-exact linear count readouts in Llama-3.2 and Qwen2.5 models that a late MLP block overwrites, \citet{datta2026early} find character counts encoded early and suppressed by the final MLPs, and \citet{hasani2025counting} locate an internal counter that updates per item and leans on separators as shortcuts. Our suite agrees on a count that is readable but not delivered, and adds the item-position and boundary-window interventions, the six-model comparison, and a calibrated capacity across 205 served models. Swap tracking is permutation composition, whose one-pass limits are known \citep{liu2023shortcuts,merrill2024illusion,kim2023entity}. Thresholded metrics can manufacture sharp transitions \citep{schaeffer2023mirage}, so CC is fitted to the whole record.

\paragraph{Limits and mechanisms of transformer state.}
Theoretical work bounds what fixed-depth attention can compute over long inputs, including counting and related formal-language families \citep{hahn2020selfattention,yehudai2024when,deletang2023chomsky,strobl2024formallanguages}, and mechanistic studies read structured quantities directly from the residual stream \citep{elhage2021framework,olsson2022induction,shai2024belief}, including periodic codes for numbers in arithmetic \citep{nanda2023progress,zhou2024fourier,kantamneni2025trigonometry}; explicit memory and recurrence are the proposed architectural remedies \citep{dai2019transformerxl,bulatov2022rmt,wu2022memorizing,borgeaud2022retro}.

\section{Model dynamics when counting fails}\label{sec:dynamics}

\begin{figure}[!h]
\centering
\includegraphics[width=\textwidth]{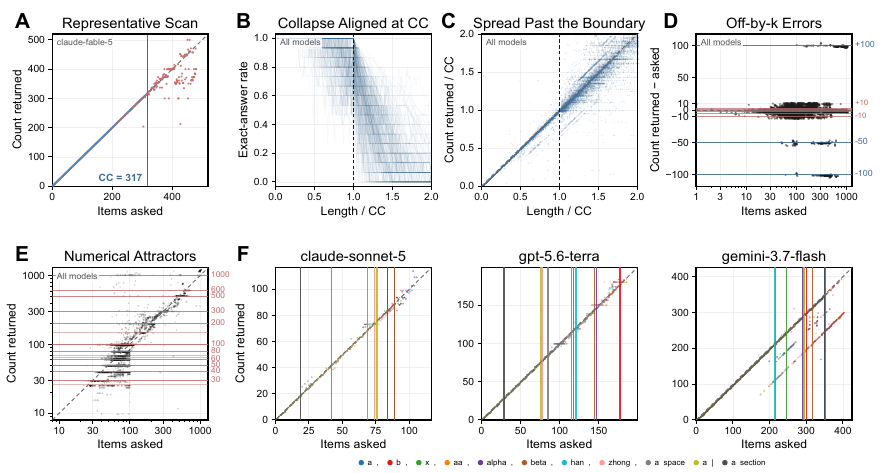}
\caption{\textbf{Model dynamics near the CC threshold.}
(\textbf{A}) A representative scan (\texttt{claude-fable-5}): the exact count is returned at every length but a few isolated slips up to the fitted CC of 317, and counting fails within a narrow band beyond it.
(\textbf{B}) Exact-answer rate against length rescaled by each measurement's CC, overlaid across all measurements: the collapse is aligned at each model's own boundary.
(\textbf{C}) All predictions on the rescaled axes: errors beyond the boundary form diagonal streaks parallel to the identity line and horizontal runs at fixed values.
(\textbf{D}) The diagonal structure: wrong answers offset from the true count by a constant amount, dominated by small slips at every scale with round offsets emerging beyond roughly one hundred items.
(\textbf{E}) The horizontal structure: wrong answers landing on preferred round integers regardless of the true count.
(\textbf{F}) Changing the counted item or the delimiter shifts the fitted CC of a single model severalfold (vertical rules mark each variant's boundary), yet the ordering of the three models is the same under every encoding; Table~\ref{tab:motif_panel} extends the sweep to 16 models.}
\label{fig:attractors}
\end{figure}

\begin{figure}[!h]
\centering
\includegraphics[width=\textwidth]{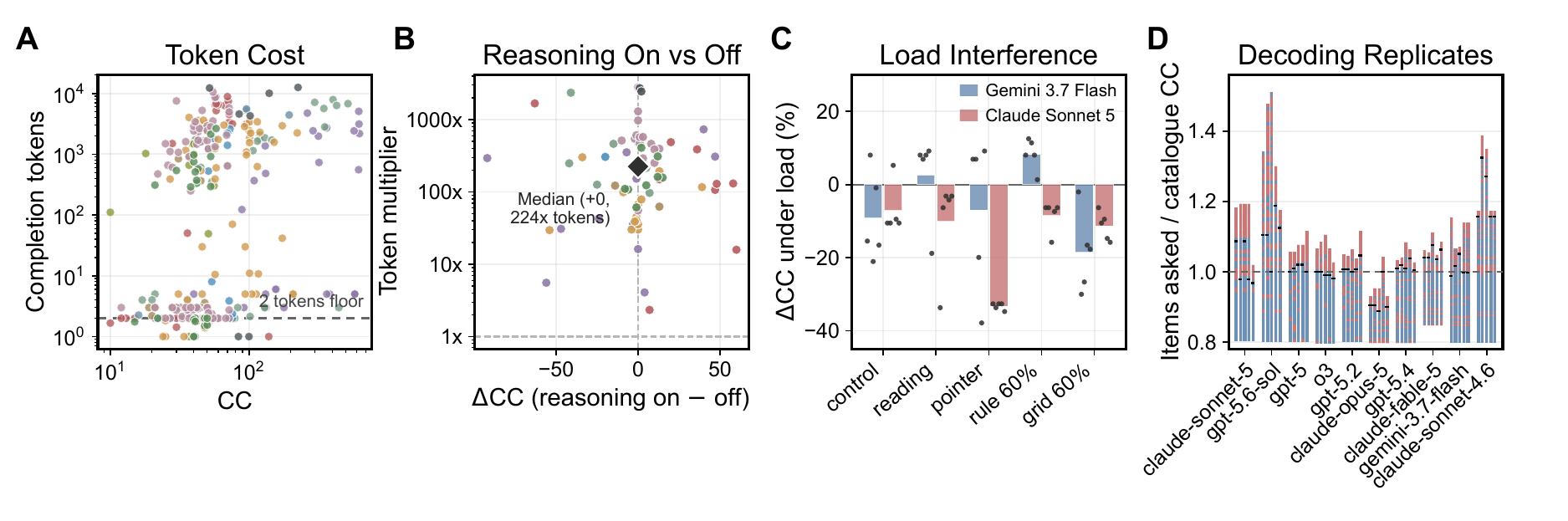}
\caption{\textbf{Impact of token consumption and test-time compute on task modeling.}
(\textbf{A}) Completion tokens per probe near each measurement's boundary against CC: reasoning configurations spend two to four orders of magnitude more tokens than non-reasoning ones over the same range of capacities.
(\textbf{B}) Matched reasoning-enabled and reasoning-disabled configurations of the same 77 hybrid models: the median change in CC is zero at a median 224-fold token cost.
(\textbf{C}) Change in CC when counting is paired with a second task, relative to the mean of five solo scans; bars are means of five scans and dots the individual scans. The controls are a grid of matched length whose answer is printed and a reading question (protocol in Appendix~\ref{app:ladder}).
(\textbf{D}) Five rescans of the transition region for ten models, nine without temperature control and one (\texttt{claude-sonnet-4.6}) pinned at zero as a control: models along the horizontal axis, items asked relative to the catalogue CC along the vertical axis, and for each model one vertical strip per scan, exact replies in blue and inexact in red, with a tick at its fitted boundary (Appendix~\ref{app:replicates}).}
\label{fig:resources}
\end{figure}

\paragraph{The collapse is abrupt and reproducible.}
A representative frontier model returns the exact count at every length up to its CC of 317 apart from a few isolated slips, then fails within a narrow band (Fig.~\ref{fig:attractors}A). When every measurement is placed on its own CC-scaled axis, the same shape appears in all 294 fitted records: accuracy stays near one through the stable regime and falls from above 0.8 to below 0.2 over a median of one tenth of CC (0.05 to 0.43 in the ten rescanned models; Fig.~\ref{fig:attractors}B; Appendix~\ref{app:estimation}). The boundary belongs to the model: five rescans of the transition region reproduce it with a standard deviation of 0.7 to 6\% of CC in nine models that offer no temperature control (Fig.~\ref{fig:resources}D). Scans that start from short lengths spread more, 5\% and 12\% in the two models we repeated, and a single scan can fall 20 to 30\% from the others (Appendix~\ref{app:replicates}). We do not interpret differences of that size between two models; the catalogue spans two orders of magnitude.

\paragraph{The errors are structured.}
Beyond CC the wrong answers fall into two patterns, diagonal streaks parallel to the identity line and horizontal runs at fixed values (Fig.~\ref{fig:attractors}C). The streaks are counts that survive with a constant offset: beyond small slips of one to five items, the offsets cluster at round values near \(\pm 100\) and \(-50\), in vendor-specific patterns (Fig.~\ref{fig:attractors}D). The runs are guesses: answers off by more than ten form bands at round values (Fig.~\ref{fig:attractors}E), and it is there that the single-number format itself starts to fail (Appendix~\ref{app:instruction}). A unit of a hundred can be lost while the rest of the count survives.

The round offsets could arise when a number of several digits is written out. We repeated the scan with the answer requested in English words, as four space-separated digits, and as the last digit alone (Appendix~\ref{app:format}). In the seven configurations that followed the formats the boundary does not move in a consistent direction: with words it is 0.80 to 1.28 of its integer value (median 1.00). In \texttt{gemini-3.7-flash}, the one model of the seven that makes hundred-item errors, the reply is exactly 100 short in 32 of 47 errors past the boundary with integers, 31 of 62 with words, and 42 of 58 with spaced digits, and its last digit stays right to the end of the scanned range. The hundred is therefore not lost in writing the final answer. Because \texttt{gemini-3.7-flash} reasons before it answers, the test does not exclude a slip inside the reasoning text.

\paragraph{The capacity depends on the encoding, and the ordering of models mostly does not.}
A counter that worked on abstract items would count any symbol equally well. Swapping the counted character or the delimiter instead shifts a single model's CC severalfold (Fig.~\ref{fig:attractors}F): one frontier model counts comma-delimited \texttt{a} to 89 but the same items delimited by a section sign only to 19. To measure how much of CC belongs to the model, we ran the eleven encodings on 16 models from seven vendors (176 scans; Appendix~\ref{app:motifs}). The model accounts for 77\% of the variance in \(\log\mathrm{CC}\) (95\% interval 50 to 88\%), the encoding for 6\%, and their interaction together with scan noise for 18\%, and the models keep their order across encodings (Kendall's \(W=0.85\)). The share depends on how far apart the models are: among the six open-weight models of Section~\ref{sec:probes}, whose capacities lie within a factor of three, the model accounts for 39\% and the encoding for 30\%. CC is a property of a model under an encoding, and the model component dominates when models differ widely.

\paragraph{Reasoning does not raise the median.}
Can a model buy capacity with tokens? Hybrid models, the same weights with reasoning on and off, give a matched comparison (Fig.~\ref{fig:resources}A, B): across the 77 pairs in which disabling verifiably took effect, the median change in CC is zero, while the median pair spends 224 times as many completion tokens. The output budget is ten times the largest capacity we measure; raising it fourfold for the sixteen reasoning members that exhausted it lifts them by a median of a dozen items and moves the median on/off difference from zero to one, though ten of those rescans remain bounded (Appendix~\ref{app:pairs}). The lack of gain is specific to counting. Three models execute pointer chains, update streams, and a small automaton, tasks whose running value can be written down, at loads up to several times their counting capacities (Appendix~\ref{app:anchored}); these tasks are long, but their state can be checkpointed at every step. A chain of thought works as external memory: it carries any running value that can be named and found again in the context. A count over identical items cannot be, so for counting, reasoning supplies a different mechanism with its own capacity, and individual models gain or lose by large amounts (Section~\ref{sec:probes}).

\paragraph{Copying fails at a finite length too.}
Asked only to copy the sequence verbatim into a tool call, a task with no counting in it and a per-token anchor throughout, each of the two models tested reproduces the sequence exactly up to a boundary of a few hundred items and then corrupts it with the same signatures counting shows (Appendix~\ref{app:copy}). The copy boundary lies above CC, so the two tasks do not fail at one length.

\paragraph{A second task lowers the count for one load per model.}
If the limit belonged to counting alone, other work done at the same time would leave it untouched. We built a small ladder of execution tasks in the grammar of abstract-reasoning grids, from counting cells of a target digit among distractors to applying a stated recoloring rule to a grid; the first, whose cells are not all alike, controls for the homogeneous stimulus (Appendix~\ref{app:ladder}). We then scanned counting while the model carried a second task placed before the sequence: a rule on a row or on a grid at 60\% of the model's own capacity for it, a pointer chain, or one of two controls of matched length, a grid whose answer is printed in the prompt and a reading question (Fig.~\ref{fig:resources}C). Each condition was scanned five times in \texttt{gemini-3.7-flash} and \texttt{claude-sonnet-5}. The controls leave CC at 0.90 to 1.03 of its solo value. One load in each model lowers it further: the pointer chain in \texttt{claude-sonnet-5}, to 0.67 of solo in all five scans, and the grid rule in \texttt{gemini-3.7-flash}, to 0.81 (\(p=0.03\) against solo, and not separable from the matched-length control at 0.91). Under the grid rule \texttt{claude-sonnet-5} keeps its count (0.89) but answers the grid correctly in only 65\% of probes. The row rule, which lowered CC by 36\% and 28\% in single scans (Table~\ref{tab:interference}), does not lower it beyond the controls in five (1.08 and 0.92). The evidence that a second task draws on the same capacity is therefore limited to one load per model, and it does not distinguish a shared resource from other forms of dual-task interference.

\section{Internal probes of the counting state}\label{sec:probes}

\begin{figure}[!htbp]
\centering
\includegraphics[width=\textwidth]{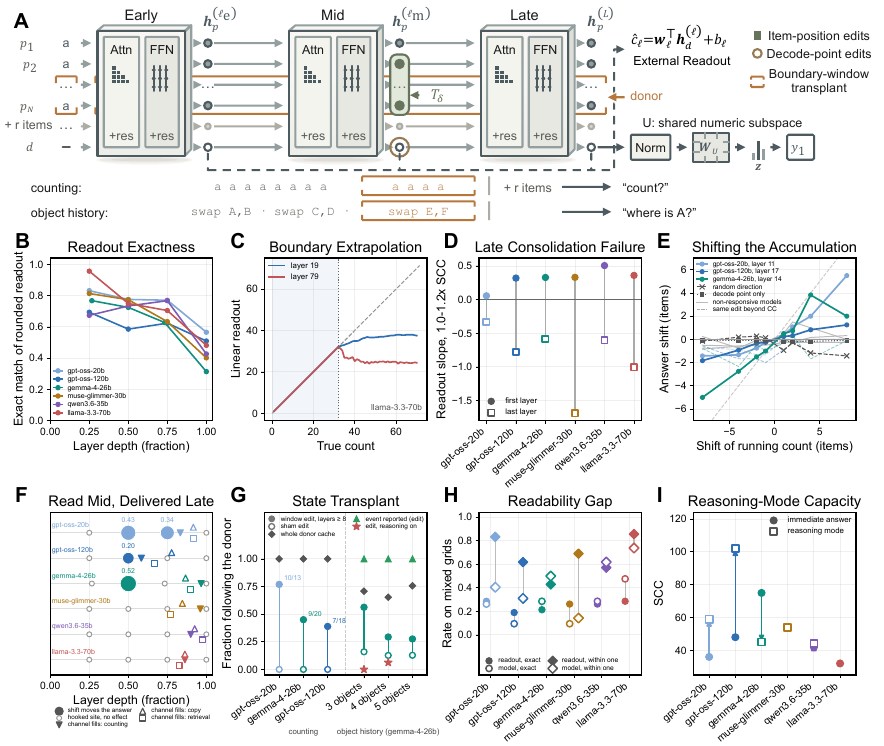}
\caption{\textbf{The counting state across six open-weight models.}
(\textbf{A}) The probe: linear readouts at every depth, edits at the decode point or across item positions, the boundary-window transplant from layer 8 upward on counting and on an object history, and the decoder's numeric subspace.
(\textbf{B}) Exact match of the rounded readout against depth.
(\textbf{C}) Readout against true count at the first and last probed layers of \texttt{llama-3.3-70b}; stable regime shaded.
(\textbf{D}) Readout slope just past the boundary (1.0 to 1.2\(\times\) CC): the first probed layer still rises at a reduced slope, the last layer falls.
(\textbf{E}) Answer shift when the count is shifted at every item position from mid-depth; random-direction controls and decode-point edits move little, and beyond CC the slope falls to at most 0.1.
(\textbf{F}) The shift moves the answer only at mid-depth; the output channel fills late for counting, copy, and retrieval alike.
(\textbf{G}) The boundary window transplanted from a donor: on counting, held-out pairs continuing from the donor's count (each bar drawn from the sham edit to the window edit; diamonds, the whole donor cache); on two-swap object histories in \texttt{gemma-4-26b}, locations answered with the donor's state, the first swap reported correctly throughout (triangles), the edit undone under native reasoning (stars).
(\textbf{H}) Rounded readout against the model's own exact match and within-one rate on mixed sequences.
(\textbf{I}) Capacity with immediate answering against native reasoning: reasoning helps, does nothing, or hurts, by model.}
\label{fig:mechanistic_probe}
\end{figure}

The count has to be carried in some internal state. We probe that state in six open-weight models from five vendors, 20B to 120B parameters, with linear readouts at four depths and interventions at the final prompt token, across the item positions, and at a prefix boundary (Fig.~\ref{fig:mechanistic_probe}A; Appendix~\ref{app:mech}).

\paragraph{The count is readable at every depth.} A held-out linear readout recovers the running count from the state at the final prompt token at every probed depth of every model. Read as a rounded integer, the readout is most exact in early layers, where it matches the true count 0.69 to 0.96 of the time, and loses 0.2 to 0.5 of that by the last layer (Fig.~\ref{fig:mechanistic_probe}B, C).

\paragraph{Past the boundary the early readout weakens and the last one reverses.} When the stable-regime readout is extrapolated past each model's CC, the readout at the earliest probed layer still rises with the true count, at 0.3 to 0.5 of its earlier slope in five models and 0.06 in the sixth. At the last layer the readout falls as the count rises, in all six models (Fig.~\ref{fig:mechanistic_probe}C, D). No site reads the count exactly past the boundary (Appendix~\ref{app:readout}). We call this latent-space collapse: a signal of the count's magnitude remains in the early layers, and the later layers no longer turn it into an exact answer.

\paragraph{Editing the readable coordinate at the final token does not move the answer.} Clamping the final-token state along the fitted count direction to targets spanning the stable regime, or stepping it by the per-item increment, leaves the decoded answer unchanged at essentially every site in every model (Appendix~\ref{app:decode}). The whole state does carry the answer: transplanting a donor prompt's full final-layer state transfers the donor's first answer token, and so does that state with its linear coordinate clamped back, while the coordinate alone never does (Fig.~\ref{fig:suite_chimera}). The answer leaves through the decoder's own answer-token subspace, which fills over a few layers late in the network. That subspace transfers the answer, loses it under ablation, and fills at nearly the same depth when the model copies or retrieves a supplied integer, so it is a general numeric output channel and is not specific to counting (Appendix~\ref{app:subspace}).

\paragraph{The count is read at mid-depth, across the item positions.} Shifting the count subspace at every item position from mid-depth onward moves the answer in three of six models, with slopes of 0.2 to 0.5 per item, while matched random shifts and decode-point edits move it little; beyond CC the slope falls to at most 0.1, and the other three models do not respond (Fig.~\ref{fig:mechanistic_probe}E, F). In the models that respond, count information is accumulated across the item positions, read from them at mid-depth, and only then delivered to the final token for decoding.

\paragraph{The boundary window carries the state, on counting and on an object history.} Replacing the residual states of the last eight to twelve items with those of a donor sequence, from layer 8 upward (layer 5 in \texttt{gpt-oss-120b}) and with the recipient's text left in place, sets the starting count for an unseen continuation, exactly in 10 of 13 held-out cases in \texttt{gpt-oss-20b}, 9 of 20 in \texttt{gemma-4-26b}, and 7 of 18 in \texttt{gpt-oss-120b}, with sham and same-count edits changing nothing (Fig.~\ref{fig:mechanistic_probe}A, G; Appendix~\ref{app:boundary}). The same replacement, applied without retuning to the last operation of an object-tracking history in \texttt{gemma-4-26b}, moves the reported locations toward the donor's state in 28 to 59\% of questions against 13 to 39\% for a sham edit. The text stays readable while the state moves: the operation the donor changed is still reported correctly in 48 of 48 histories in every cell, an untouched object stays put in at least 45 of 48, and a supplied-state question is still answered in 36 to 44 of 48. The same task, with history length and state size varied, fails the way counting fails: the state update breaks from three swaps onward while the last swap is still reported and a supplied state still used (Appendix~\ref{app:boundary}). With native reasoning on, the trace re-reads the history and undoes the transplant in 38 of 39 cases. The window therefore carries the state that the immediate answer is read from, and where the text is a sufficient record of that state, reasoning can rebuild it. A sequence of identical items offers no such record, since no written value can be tied back to the item where it was written.

\paragraph{Count and length are read along separate directions.} On sequences in which a target item appears \(k\) times among distractors, the count and the total length are independently readable along nearly orthogonal directions, yet immediate-answer accuracy falls to between 10\% and 48\% at counts far inside each model's stable regime on the homogeneous task (Fig.~\ref{fig:mechanistic_probe}H). An external readout recovers the count within one item more often than the model answers exactly, and the exact count more often in three of six models (Appendix~\ref{app:mixed}). Mixed sequences are also harder than homogeneous ones, so the limit does not come from long runs of identical tokens.

\paragraph{Reasoning changes capacity in both directions.} Enabling each model's native reasoning raises capacity by 64\% and 113\% in the two models whose silent readouts are the noisiest of the six, leaves it unchanged in two others, and lowers it by 40\% in a model whose traces re-count in groups and truncate at the budget (Fig.~\ref{fig:mechanistic_probe}I).

\begin{figure}[!htbp]
\centering
\includegraphics[width=\textwidth]{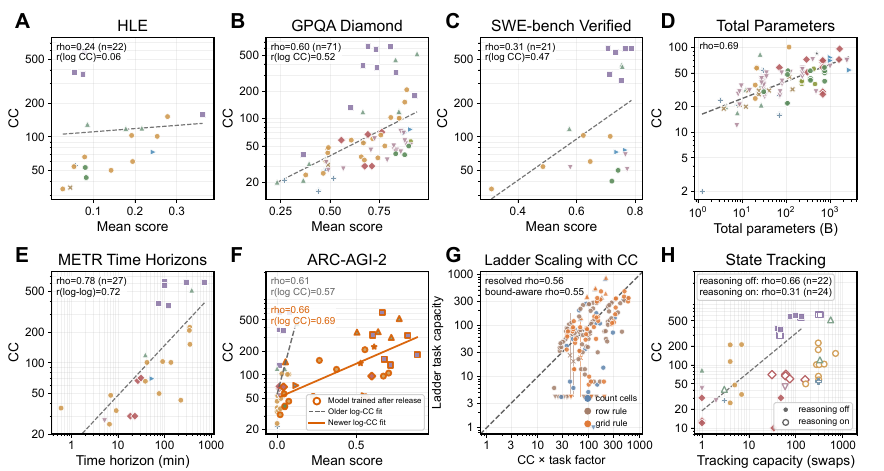}
\caption{\textbf{Relation between counting capacity and standard AI evaluations.}
(\textbf{A}--\textbf{D}) The axes CC is not reducible to: knowledge benchmarks (Humanity's Last Exam, GPQA Diamond), coding (SWE-bench Verified), and model scale (total parameters).
(\textbf{E}) The METR task-completion time horizon, the correlate that survives conditioning on general capability.
(\textbf{F}) ARC-AGI-2 score, with models trained after the benchmark's public release in orange: the two cohorts fit separately, consistent with format familiarity limiting the earlier one, though a release split cannot isolate familiarity from other changes.
(\textbf{G}) Ladder scaling with CC: measured capacity on a graded series of tasks in ARC's grid vocabulary, from counting cells to executing a rule on a grid, against CC scaled by each task's fitted factor, so the identity line is proportional scaling (Appendix~\ref{app:ladder}); circles are resolved boundaries, down and up triangles are upper and lower bounds, and bars the tested ranges of mixed scans.
(\textbf{H}) Tracking capacity on an object-history task with no count in it against CC, reasoning off (filled) and on (open): with reasoning off the two scale together; with reasoning on tracking improves in most models, consistent with reconstruction from the written history, and the estimated loosening of the coupling is uncertain (Appendix~\ref{app:tracking}).}
\label{fig:landscape}
\end{figure}

\section{Comparing CC with standard evaluations}\label{sec:evaluations}

CC is a reproducible property of the trained model under the tested protocol, and we ask what it relates to. Fig.~\ref{fig:landscape} arranges the comparison in two rows. The top row shows knowledge, coding, and scale: Humanity's Last Exam, SWE-bench Verified, and GPQA Diamond correlate with CC at 0.24, 0.31, and 0.60, the last explained by general capability, and total parameter count at 0.69 over the 103 models whose sizes are public, with active parameters adding nothing given the total (Appendix~\ref{app:benchmark}). The bottom row shows the relations the paper tests, beginning with the METR time horizon, the duration of tasks a model can carry to completion \citep{kwa2025metr}.

\paragraph{The METR horizon.} Balrog, WeirdML, and METR have the largest raw correlations with CC, but only METR's remains after conditioning on the Epoch aggregate index (partial \(\rho=0.74\) over 24 models, against 0.07 and 0.30). The screen covered 66 benchmarks, so the result needs a correction for multiple comparisons. The METR partial correlation is the largest positive one among all benchmarks with enough shared models to compute it, with \(p=5\times10^{-5}\) unadjusted and below 0.004 after Bonferroni adjustment for every benchmark screened. Whether CC improves prediction is less clear: adding it to the index raises vendor-held-out \(R^2\) for the horizon from 0.81 to 0.90, but a bootstrap over models puts the gain at 0.06 with a 95\% interval from \(-0.24\) to 0.30 (Appendix~\ref{app:benchmark}). The relation rests on 24 models and should be tested on more.

\paragraph{ARC-AGI-2 and the execution ladder.} ARC-AGI-2 reduces factual dependence through abstract transformations drawn from a fixed task distribution \citep{chollet2025arcagi2}, and models trained before and after its release behave differently (Fig.~\ref{fig:landscape}F): pre-release models span a wide CC range but score poorly, while the 29 post-release models from eight vendors score higher and align with capacity. The partial correlation given the Epoch index is 0.29, so much of this coupling is general capability; one reading, which a release split cannot establish, is that newer models adapted to the format, as they might to a change of encoding. The execution these grids demand is also measurable directly. Running the six execution-ladder tasks across the post-release cohort and fitting each boundary against its CC-scaled prediction under censoring gives a slope of 1.00, with a cluster-bootstrap interval of 0.46 to 1.51 (Fig.~\ref{fig:landscape}G). The coupling is consistent with proportional scaling, but the residual departures are large, a standard deviation of 1.5 on the log scale. The six tasks rank the models almost identically (Kendall's \(W=0.86\)) while each agrees with CC only moderately, so the ladder measures one coherent execution capacity that is coupled to CC and distinct from it (Appendix~\ref{app:ladder}).

\paragraph{A state-tracking task with no count in it.} We scanned object tracking, four objects and a history of swaps, with the counting protocol on a prespecified subset, 33 models in 46 configurations (Appendix~\ref{app:tracking}). With reasoning off, tracking capacity follows CC across 22 models (\(\rho=0.66\), bootstrap interval 0.28 to 0.90) and survives conditioning on the Epoch index (partial 0.65). The association lies between vendor families, however: without the five Anthropic models it is 0.31, and the vendor-demeaned correlation is 0.24, so within families the sample is too small to test. With reasoning on, tracking capacity rises in 12 of the 13 hybrids, by a median factor of 4.6, while CC rises by a factor of 1.07. At the boundary the last swap is still reported and a supplied state still used while tracking is at chance (Fig.~\ref{fig:landscape}H).

We have not found a prompt that raises CC above the bare prompt (Appendix~\ref{app:prefix}). The concurrent-load experiment is the only direct test of a capacity shared between tasks, and after replication it supports one for a single load in each model (Section~\ref{sec:dynamics}); elsewhere a shared capacity is an interpretation.

\section{Discussion}\label{sec:discussion}

Exact counting is a transparent test of one thing long-horizon work asks of a model: keeping a quantity that every step changes and that nothing in the context records. The limit on counting is short and has barely moved across model generations. The limit is a stable property of a model under an encoding. Gradual drift, the format of the written answer, and the output budget do not explain it, and reasoning does not raise it on the median. In open-weight models a weakened signal of the count survives in early layers past the boundary and is not delivered, and the edit that carries the count forward also carries the state of an object history (Section~\ref{sec:probes}).

We read these results through a hypothesis: counting measures a distinct capacity, task modeling. On this reading the capacity is silent, in that the answer comes at once or not at all. A chain of thought replaces it whenever the state can be named, which would explain why reasoning improves tracking in twelve of thirteen hybrids while its median effect on counting is nil. On this reading the capacity also matters for long tasks, as the METR relation and the grid ladder suggest. The evidence for the hypothesis is correlational apart from two experiments of narrow scope: the object-history transplant covers one model, and a second task lowers CC beyond its controls for one load in each of two models. We have not shown that counting is necessary or sufficient for agentic work, and the association with tracking lies between vendor families more than within them.

Deployed agents reason with scratchpads, so it is unclear why a silent capacity should predict horizon. Perhaps written state must be re-read and integrated in one pass per step, so the silent capacity bounds the usable scratchpad, which varying the size of the log would test. Failures of this kind resist prompt engineering: no instruction we have found raises the capacity, and a chain of thought substitutes for the internal state only where it can be named, so an agent whose running state is never written fails at a length its context window does not announce. Explicit recurrence or external memory \citep{dai2019transformerxl,bulatov2022rmt,wu2022memorizing} is one route to a larger capacity.

Homogeneous counting is an artificial test and measures one kind of state. Served models are reached through routes whose hidden prompts and reasoning-token accounting are not observable. A single scan can differ from another of the same model by 20 to 30\%, and the calibrated interval contains the true boundary 90 to 92\% of the time under the stop rule, below the nominal 95\%. The output-format control tests hundred-item errors only in a model that reasons before it answers. The mechanistic analyses cover six open-weight models, and the METR relation was found by screening and rests on 24 models. A scan costs a few hundred characters per probe and also fingerprints what an API serves (Appendix~\ref{app:fingerprint}), so each of these limits can be tested at larger scale.

\subsection*{AI use statement}

Generative AI tools were used as coding assistants for implementing parts of the evaluation harness, the analysis pipeline, and the data visualization code used to produce the figures in this paper. All such code was executed by the authors and its outputs checked against the recorded experimental results. The classification of stored reasoning traces in Appendix~\ref{app:reasoning} was performed with LLM assistance under a fixed rubric; the reported statistics were recomputed from the recorded per-trace labels, and every quoted excerpt was verified verbatim against the stored trace by the authors. Generative AI tools were not used to generate research ideas, experimental designs, hypotheses, mathematical claims or proofs, or synthetic data, and no reported measurement originates from a generative model. We have reviewed all AI-assisted work and take responsibility for the final content of this paper, including all text, claims, and artifacts.

\subsection*{Ethics statement}

This work evaluates commercially served and open-weight language models on a synthetic counting task. No human subjects, personal data, or private datasets were involved, and every stimulus is generated programmatically from a fixed alphabet. Externally sourced benchmark scores used in the correlation analysis come from public leaderboards and are cited at the point of use. Models were queried through their standard public interfaces under the providers' terms of service, with tool use disabled except in the copy control of Appendix~\ref{app:copy}, which forces a single recording tool call. We see no direct avenue for misuse: the measurement captures a capability limit rather than prompting harmful behavior, and the mechanistic interventions are confined to open-weight checkpoints running locally. That task modeling fails silently is safety-relevant: it argues against trusting long-horizon autonomous execution on the strength of benchmark scores alone.

\subsection*{Reproducibility statement}

The CC measurement is fully specified by the prompt templates and generation settings in Appendix~\ref{app:prompts} and the estimation procedure in Appendix~\ref{app:estimation}; the bounds on guessing strategies are derived in Appendix~\ref{app:guessing}. The complete catalogue of the 205 evaluated models, with provider, parameter metadata, nominal context window, and measured counting capacity, is in Appendix~\ref{app:catalogue}. The secondary measurements are specified in Appendix~\ref{app:ladder}, and the cross-model mechanistic probe suite in Appendix~\ref{app:mech}. Every evaluation runs from a single versioned configuration file with a fixed RNG seed, so the procedure of a reported run is reproduced by re-running that configuration; the realized replies are not, since 33 models ignore the temperature and several reject the seed (Appendix~\ref{app:replicates}).

\ificlrfinal
All raw inference logs for the 205 evaluated models (including API token-usage metadata and parsed outputs), the mechanistic probe records, the execution-ladder and interference measurements, and the full evaluation and analysis source code are available at
\url{https://github.com/txdai/llm-extended-rule-following}.

\subsection*{Acknowledgments}

\paragraph{Funding.} National Science Foundation 2103301 (J.A.F.); The Packard
Foundation 2016-65132 (J.A.F.).

\paragraph{Author contributions.} T.D. conceived the study, designed the
measurements and evaluation pipeline, performed the large-scale model evaluations, and carried out the correlation analyses and mechanistic investigations. J.A.F. supervised the project and contributed to interpretation. Both authors contributed to writing the manuscript.

\paragraph{Competing interests.} The authors declare no competing interests.
\else
All raw inference logs for the evaluated model variants (including API token-usage metadata and parsed outputs), the mechanistic probe records, the execution-ladder and interference measurements, and the full evaluation and analysis source code will be released with the paper. The repository URL is withheld here to preserve anonymity and is provided in the camera-ready version.
\fi

\bibliographystyle{iclr2027_conference}
\bibliography{references}

\appendix
\section{The counting measurement}\label{app:prompts}

Counting prompts were generated from homogeneous item sequences. The baseline stimulus consisted of repeated lowercase characters separated by a comma and a space. The target value was the number of repeated items, not the number of tokens. We verified tokenization lengths for all evaluated tokenizers to identify compression anomalies and quantify the relationship between item count and input-token count.

Models were instructed to return the final count as a single integer. We did not enforce this restriction by constraining the decoder with a numerical grammar or a small \texttt{max\_tokens} limit. This allowed natural failure modes, including prompt echoing, blank outputs, formatting artifacts, and spurious reasoning traces, to be observed. Generated responses were parsed for the last valid integer and compared with the exact target item count. Outputs without a parsable integer were classified as failures. For each trial, we recorded the target count, generated response, parsed prediction, binary correctness, absolute error, input-token count, output-token count, and model metadata.

The harness uses the following exact system instruction string:

\begin{verbatim}
You count items exactly. Return only one integer with no words,
punctuation, or explanation.
\end{verbatim}

The user prompt template is:

\begin{verbatim}
How many occurrences of "{item_label}" are in this sequence?

{sequence}
\end{verbatim}

Here \texttt{\{item\_label\}} is the counted symbol label (for example \texttt{a}) and \texttt{\{sequence\}} is the generated repeated-token sequence with the chosen delimiter. For Google models, the preferred path sends the system instruction separately as a developer/system instruction and sends the user string above as the sole content block. If that provider path is not available, the harness falls back to an inline Google prompt formed by concatenating:

\begin{verbatim}
{system_instruction}

{user_prompt}
\end{verbatim}

Across all APIs, the measurement used deterministic prompting with the fixed textual templates above. The shared harness settings were:

\begin{itemize}
\item Output cap: \texttt{8192} tokens, fixed
\item Preflight count: \texttt{8}
\item Scan parameters: start length \texttt{1}, step \texttt{1}, one trial per length,
ceiling \texttt{3000}, stop after \texttt{10} consecutive unparsable replies
\end{itemize}

The output cap is fixed rather than escalated on retry, because escalation would make the token budget a function of the probed length and the resulting capacity would confound counting ability with generation allowance. The cap is also more than ten times the largest capacity we measure, which is bounded by wrong answers rather than by tokens; Appendix~\ref{app:pairs} reports what a fourfold larger cap changes for the configurations that exhausted it. A reply that exhausts the cap without producing a parsable integer is recorded as inexact and attributed to the budget. Measurements whose transitions are attributable to the budget rather than to wrong answers are reported as budget-bounded and are excluded from statements about counting limits (Appendix~\ref{app:estimation}).

\subsection{Decoding settings and models that do not accept them}

The measurement requests greedy decoding: temperature \texttt{0}, and, where the provider requires it to be stated explicitly alongside greedy sampling, \texttt{top\_p} \texttt{= 1}. The task admits exactly one correct answer, so sampling contributes variance without contributing information, and the comparison across models is cleaner when every model is asked to decode the same way.

Not every model honors the request. Of the 205 models in the catalogue, 33 offer no temperature control: the OpenAI \texttt{o}-series and GPT-5 family (27 models) reject the parameter outright; Anthropic releases from Opus 4.7 onward (five models) either reject it or, according to the provider's migration documentation, accept it and silently ignore it; and one Google model likewise accepts and ignores it. For these the parameter has no effect, and each response is a single draw from the provider's own decoding configuration rather than a deterministic reading. A scan that issues one probe per sequence length cannot distinguish a model that is reliably wrong at a given length from one that is occasionally wrong there, so for these 33 models an isolated error may reflect sampling variance rather than a stable property of the model. Each such model carries a \texttt{temperature\_locked} flag in its measurement record (set for models that reject the parameter and for the three known to ignore it), so analyses that require pinned decoding can exclude them. Appendix~\ref{app:replicates} measures how much this sampling moves the fitted boundary.

A further 25 models do not accept a \texttt{seed} parameter, which removes the remaining route to reproducible sampling on the subset that also fixes temperature. We report results for these models alongside the rest rather than excluding them, since excluding the entire OpenAI reasoning line would remove much of the frontier, but the distinction is preserved in the data.

\subsection{Decoding replicates}\label{app:replicates}

A scan issues one probe per length, so for the 33 models above every reply is a single sample from the provider's decoder, and the profile interval of Appendix~\ref{app:estimation} describes the fit to that one record rather than what a second scan would return. We measured the second scan directly. Ten models were rescanned five times each through their transition region: nine that offer no temperature control, spanning the three providers above, and \texttt{claude-sonnet-4.6}, the most recent model in the catalogue that honors temperature zero, as the control. Each replicate uses the catalogue protocol and the identical stimulus, starts at 0.8 of the catalogue CC (0.85 for the largest-capacity model), and runs until the criterion of Appendix~\ref{app:estimation} resolves the transition, so the record covers the passage from mostly right to mostly wrong without repeating the stable prefix; the two-regime fit depends only on the probed record, so a truncated prefix leaves the estimate comparable to the catalogue's. Table~\ref{tab:replicates} lists every fitted boundary and Fig.~\ref{fig:resources}D draws the records.

The fitted boundary is stable: across the nine models without temperature control the five replicates have a standard deviation of 0.7\% to 6\% of CC, and the replicate median lies within 10\% of the catalogue value for eight of the ten models. Refitting each catalogue record on the same truncated window returns the full-record boundary or the upper end of its interval, so the window itself does not bias the comparison. The ordering of the models is stable: no two models whose catalogue capacities differ by more than 10\% exchange places in any replicate. The individual replies are not: at a length probed by all five replicates, the outcomes agree about as often as independent draws at that length's mean accuracy would, so the boundary is stable because the estimator integrates over stochastic replies, not because the replies repeat. The pinned control shows the same: the one model pinned at temperature zero agrees with itself no more often than the sampled models do, and its five boundaries are the most widely spread of the ten. Pinning the temperature does not make a served model deterministic, and the models that cannot be pinned are no less stable than the one that can. The profile interval of a single scan therefore describes the fit to one record and understates run-to-run uncertainty for a served model, by an order of magnitude when the interval collapses to one or two items, while the run-to-run uncertainty itself remains small against the range of capacities across models, which spans two orders of magnitude.

These rescans begin at 0.8 of the catalogue CC and so cannot register a boundary that falls earlier. Later scans begin at short lengths and spread more. The five solo scans of the concurrent-load replication (Appendix~\ref{app:ladder}, Table~\ref{tab:interference_replicates}; from 0.2 of CC) give boundaries of 290, 370, 346, 386, and 394 for \texttt{gemini-3.7-flash}, a standard deviation of 12\% of their mean, and 92, 100, 100, 92, and 91 for \texttt{claude-sonnet-5}, 5\%. Eight cells of the encoding sweep (Fig.~\ref{fig:attractors}F) were also scanned a second time from length one: six repeats are within 9\% of the first scan and two are not (246 then 295 for \texttt{gemini-3.7-flash} on the letter \texttt{x}; 89 then 64 for \texttt{claude-sonnet-5} on the baseline). A single scan can therefore fall 20 to 30\% from the others of the same model, and we do not interpret differences of that size between two models.

\begin{table}[htbp]
\centering
\footnotesize
\setlength{\tabcolsep}{3pt}
\caption{\textbf{Decoding replicates.} Five independent rescans of the transition region per model under the catalogue protocol, with the same stimulus in every replicate. Temperature: whether the provider rejects the parameter, accepts and ignores it, or honours zero. Agreement is the share of commonly probed lengths on which all five replicates return the same outcome, against the share expected if each reply were an independent draw at that length's mean accuracy.}
\label{tab:replicates}
\begin{tabular}{llrlrrr}
\toprule
Model & Temperature & CC [95\% CI] & Replicate boundaries & SD & SD/CC & Agreement \\
\midrule
\texttt{claude-sonnet-5} & rejects & 92 [91, 99] & 100 90 100 90 89 & 5.1 & 5.5\% & 85\% (88\%) \\
\texttt{gpt-5.6-sol} & rejects & 96 [95, 96] & 106 106 96 114 108 & 5.8 & 6.0\% & 75\% (81\%) \\
\texttt{gpt-5} & rejects & 101 [67, 104] & 101 102 103 103 101 & 0.9 & 0.9\% & 54\% (65\%) \\
\texttt{o3} & rejects & 103 [103, 103] & 103 103 102 102 101 & 0.7 & 0.7\% & 71\% (78\%) \\
\texttt{gpt-5.2} & rejects & 152 [146, 153] & 153 153 152 153 159 & 2.5 & 1.7\% & 39\% (52\%) \\
\texttt{claude-opus-5} & ignores & 179 [131, 180] & 162 162 159 179 161 & 7.3 & 4.1\% & 58\% (64\%) \\
\texttt{gpt-5.4} & rejects & 208 [208, 208] & 210 212 210 216 209 & 2.5 & 1.2\% & 60\% (69\%) \\
\texttt{claude-fable-5} & rejects & 317 [316, 329] & 330 330 341 328 337 & 5.0 & 1.6\% & 59\% (70\%) \\
\texttt{gemini-3.7-flash} & ignores & 352 [350, 373] & 348 358 370 351 351 & 7.9 & 2.2\% & 60\% (69\%) \\
\texttt{claude-sonnet-4.6} & pinned at 0 & 320 [315, 320] & 370 424 407 370 370 & 22.9 & 7.2\% & 83\% (87\%) \\
\bottomrule
\end{tabular}
\end{table}

\subsection{CC as a serving fingerprint}\label{app:fingerprint}

A model's capacity together with the lengths and values of its wrong answers is reproducible enough across scans to identify it (Appendix~\ref{app:replicates}, where the boundary is stable although individual replies are not): two models never fail at the same lengths with the same answers. This makes CC a cheap check of what an API endpoint actually serves. The protocol is the catalogue measurement with a coarser schedule: the prompt of Appendix~\ref{app:prompts}, temperature zero, an output cap of 512 tokens, one probe at every length from 1 to 60, every second length to 200, every fifth to 400, and every tenth to 800, 210 probes in all, fitted as in Appendix~\ref{app:estimation} and compared with the catalogue record of the claimed model by capacity and by the wrong answers at shared lengths.

We applied it to an OpenAI-compatible reseller advertising ten Claude names (six models and four aliases), four GPT names, and two Grok names, and scanned eight of the Claude names, three of the GPT names (the fourth timed out on 205 of 210 probes), and both Grok names. Every scanned Claude name returned the same fingerprint: a capacity of 460 to 480, off-by-one slips at the same lengths (365, 470, 480, 490), and identical answers at 197 to 205 of the 210 probed lengths across names, where the catalogue capacities of the claimed models range from 92 to 622. Every scanned GPT and Grok name returned a capacity of 44 to 46 with the same wrong answers at most shared failing lengths, against catalogue values of 94 to 226, and matched the catalogue record of one small OpenAI model answer for answer; the Grok routes also timed out on two thirds of their probes. Repeating one length eight times at temperature zero gave two different answers, so the Claude route was sampled without temperature control. The operator subsequently confirmed both routes: an Opus-class model served through a coding-agent harness behind the Claude names, and the small model behind the GPT and Grok names. The reseller is not named.

\paragraph{Prompt prefixes.}\label{app:prefix} We have not found a prompt that raises CC above the bare catalogue prompt of Appendix~\ref{app:prompts}. The evidence is narrow and is stated as such: on the route above, stacking 5k, 10k, and 20k tokens of neutral system prompt ahead of the task left every probed reply unchanged, including the first slip; the reasoning-mode comparison of Section~\ref{sec:dynamics} is the other prompt-level variation we tested, and across 77 verified pairs it leaves the median unchanged. Prompts that change the stimulus (the counted item or the delimiter, Fig.~\ref{fig:attractors}F) move the fitted CC severalfold in both directions without reordering the models.

\subsection{Models excluded for provider-side reasons}

Four models in the candidate roster could not be measured under the protocol above and were excluded, with the reason recorded rather than inferred from the model's name.

\texttt{meta-llama/llama-guard-4-12b} is a safety classifier rather than a general model; it replies \texttt{safe} to every prompt, including prompts that contain no counting task. Exclusion is by behavior rather than by name: a similarly named safeguard model counts normally and is retained.

\texttt{mistralai/mistral-small-3.1-24b-instruct} and \texttt{qwen/qwen-2.5-coder-32b-instruct} return a provider error whenever the system message constrains the reply to a bare integer. The failure is reproducible across prompt variants and is not a property of either model: both answer counting questions normally when the system message is omitted or replaced with an unconstrained one, and both are served by a single provider with no alternative endpoint. The failure is therefore attributable to the serving path rather than to the model, and is noted here because constrained-output instructions are common in automated evaluation and this failure mode is silent: the provider's HTTP 500 reaches the client as an HTTP 200 with an error object in the body, so a harness that inspects only the status code will record it as an empty model response.

\texttt{mistralai/mistral-7b-instruct} is listed by the model catalogue but has no live endpoint (HTTP 404).

\subsection{Collapse-region analysis}

To map the transition near collapse, we used the dense per-length record around the CC boundary produced by the scan. Parsed predictions were aggregated to quantify the transition from exact outputs to failed outputs. Failed-output attractors were identified by aggregating incorrect predicted values across the sweep (Fig.~\ref{fig:attractors}D, E).

\FloatBarrier

\subsection{Encodings across sixteen models}\label{app:motifs}

Fig.~\ref{fig:attractors}F varies the counted item and the delimiter for three models. To measure how much of CC belongs to the model and how much to the encoding, we ran the same eleven encodings on thirteen more API models, which gives a balanced panel of 16 models from seven vendors (176 scans), and on the six open-weight models of Section~\ref{sec:probes}, run locally with greedy decoding (66 scans). The thirteen added models answer immediately, and the three models of Fig.~\ref{fig:attractors}F keep their original scans. Table~\ref{tab:motif_panel} lists every fitted CC.

We split the variance of \(\log\mathrm{CC}\) into a model component, an encoding component, and a residual that holds their interaction together with scan-to-scan noise (a two-way random-effects model with one scan per cell; on a balanced panel the analysis-of-variance estimates are the restricted maximum-likelihood ones). In the API panel the model accounts for 76.6\% of the variance (95\% interval 50.3 to 88.4\%, from a bootstrap over models), the encoding for 5.6\% (2.5 to 16.9\%), and the residual for 17.8\% (7.9 to 34.9\%). On raw CC the shares are 74.1\%, 2.6\%, and 23.3\%. The models keep their order from one encoding to the next: Kendall's \(W\) is 0.85, and the Spearman correlation between two encodings averages 0.83 over all pairs (lowest pair 0.41).

The model share depends on how far apart the models of a panel are. The median capacities of the API models run from 12 to 295. Those of the six open-weight models run from 27 to 75, and among them the model accounts for 38.5\% of the variance, the encoding for 29.8\%, and the residual for 31.7\% (\(W=0.71\)). Pooling all 22 models gives 71.9\%, 8.8\%, and 19.3\% (\(W=0.83\)).

One encoding can still move one model a long way. The section-sign delimiter cuts CC to a third of its comma value or less in five of the 16 API models (to 0.11 of it in \texttt{claude-sonnet-4.6}) and leaves it within 6\% in three. Token volume does not explain the shifts. Within each of the three models of Fig.~\ref{fig:attractors}F, the rank correlation between tokens per item and the change in CC across the eleven encodings is weak and negative (\(-0.31\) to \(-0.52\)), and the section-sign delimiter adds a tenth of a token per item for one model while removing three quarters of its capacity. Across the 176 API scans, tokens per item, centered within each model, correlate with \(\log\mathrm{CC}\) at \(r=0.12\), which is 1\% of the within-model variance. Among the local models the correlation is 0.44 (20\%): the space delimiter halves the tokens per item and lowers CC in all six. The shifts therefore depend on the surface form itself and only weakly on the number of tokens it takes.

\begin{table}[htbp]
\centering
\scriptsize
\setlength{\tabcolsep}{2.4pt}
\caption{\textbf{Counting capacity under eleven encodings.} Fitted CC for each model and encoding: the counted item is varied with a comma delimiter (first eight columns; \textit{han} and \textit{zhong} are the Chinese characters U+6C49 and U+4E2D), then the delimiter is varied with the item \texttt{a} (space, pipe, section sign). Upper block: 16 API models; \texttt{\#nothink} marks a hybrid model with reasoning off and \texttt{\#think} one with reasoning on. Lower block: the six open-weight models run locally.}
\label{tab:motif_panel}
\begin{tabular}{lrrrrrrrrrrr}
\toprule
Model & \texttt{a,} & \texttt{b,} & \texttt{x,} & \texttt{aa,} & $\alpha$\texttt{,} & $\beta$\texttt{,} & \textit{han}\texttt{,} & \textit{zhong}\texttt{,} & \texttt{a\textvisiblespace} & \texttt{a|} & \texttt{a\S} \\
\midrule
\texttt{claude-sonnet-4.6\#nothink} & 370 & 292 & 288 & 428 & 96 & 96 & 53 & 91 & 141 & 97 & 42 \\
\texttt{gemini-3.7-flash} & 351 & 301 & 246 & 293 & 289 & 318 & 214 & 295 & 217 & 295 & 349 \\
\texttt{gpt-5.4\#nothink} & 189 & 214 & 175 & 171 & 177 & 198 & 219 & 217 & 125 & 170 & 31 \\
\texttt{claude-haiku-4.5\#nothink} & 132 & 132 & 132 & 137 & 132 & 133 & 72 & 137 & 45 & 137 & 128 \\
\texttt{gpt-5.6-terra\#think} & 115 & 177 & 118 & 115 & 147 & 145 & 121 & 78 & 85 & 76 & 29 \\
\texttt{claude-sonnet-5\#think} & 89 & 89 & 89 & 84 & 84 & 89 & 69 & 74 & 42 & 76 & 19 \\
\texttt{gemini-2.5-flash\#nothink} & 60 & 68 & 50 & 58 & 57 & 60 & 100 & 52 & 50 & 52 & 41 \\
\texttt{gpt-4.1} & 55 & 101 & 85 & 67 & 101 & 72 & 84 & 65 & 39 & 62 & 41 \\
\texttt{llama-4-maverick} & 55 & 64 & 60 & 5 & 59 & 59 & 65 & 73 & 50 & 55 & 18 \\
\texttt{qwen3-235b-a22b-2507} & 54 & 50 & 52 & 59 & 57 & 55 & 50 & 45 & 33 & 45 & 38 \\
\texttt{gemma-4-26b-a4b-it\#nothink} & 46 & 75 & 75 & 75 & 71 & 75 & 83 & 75 & 37 & 75 & 60 \\
\texttt{gpt-4.1-mini} & 41 & 41 & 48 & 50 & 41 & 41 & 50 & 41 & 32 & 38 & 34 \\
\texttt{mistral-medium-3.1} & 38 & 38 & 38 & 44 & 43 & 40 & 28 & 29 & 16 & 40 & 40 \\
\texttt{qwen3.6-35b-a3b\#nothink} & 37 & 33 & 33 & 33 & 33 & 30 & 41 & 41 & 41 & 44 & 16 \\
\texttt{llama-3.3-70b-instruct} & 28 & 42 & 32 & 28 & 41 & 28 & 32 & 27 & 11 & 32 & 17 \\
\texttt{deepseek-v3.2\#nothink} & 12 & 12 & 10 & 15 & 8 & 14 & 12 & 19 & 11 & 12 & 10 \\
\midrule
\texttt{gemma-4-26b-a4b-it} (local) & 75 & 75 & 75 & 75 & 75 & 75 & 75 & 75 & 33 & 75 & 61 \\
\texttt{muse-glimmer-30b} (local) & 54 & 54 & 64 & 64 & 64 & 62 & 60 & 56 & 26 & 56 & 22 \\
\texttt{gpt-oss-120b} (local) & 48 & 38 & 48 & 122 & 40 & 40 & 30 & 40 & 20 & 122 & 71 \\
\texttt{qwen3.6-35b-a3b} (local) & 42 & 38 & 42 & 42 & 33 & 30 & 45 & 41 & 29 & 42 & 15 \\
\texttt{gpt-oss-20b} (local) & 36 & 38 & 34 & 36 & 52 & 27 & 27 & 25 & 13 & 27 & 5 \\
\texttt{llama-3.3-70b-instruct} (local) & 32 & 41 & 41 & 37 & 32 & 30 & 43 & 30 & 11 & 32 & 8 \\
\bottomrule
\end{tabular}
\end{table}

\subsection{Output format}\label{app:format}

The round-number errors of Fig.~\ref{fig:attractors}D could arise when a number of several digits is written out, and the boundary itself could depend on the numeral format. We repeated the baseline scan with the answer requested in four formats: a bare integer (the catalogue prompt), lowercase English words (\texttt{one hundred forty-two}), four space-separated digits with leading zeros (\texttt{0 1 4 2}), and the last digit of the count alone. The system instructions for the three new formats are (line breaks added here):

\begin{verbatim}
You count items exactly. Return the exact count written out
in lowercase English words only (for example, 'forty-two' or
'one hundred twenty-three'), with no digits, punctuation, or
explanation.

You count items exactly. Return the exact count as four
space-separated digits padded with leading zeros (for example,
'0 0 4 2' or '0 1 2 3'), with no words, punctuation, or
explanation.

You count items exactly. Return only the ones digit of the
exact count (the count modulo 10), as a single digit from 0 to
9, with no words, punctuation, or explanation.
\end{verbatim}

The user prompt is the catalogue's with the format restated. A reply is scored on the value it encodes, so a last-digit reply is exact one time in ten by chance, which is below the one half that defines the boundary.

Nine configurations were scanned. Five ran through the API: \texttt{gemini-3.7-flash} with reasoning at medium effort, at odd lengths up to 479, and \texttt{gpt-5.4}, \texttt{gemini-2.5-flash}, \texttt{claude-sonnet-5}, and \texttt{claude-sonnet-4.6} with reasoning off, at every length and with an output cap of 128 tokens. Four open-weight models ran locally with greedy decoding.

Two configurations did not follow the formats and are left out of the comparison. \texttt{claude-sonnet-4.6} wrote its counting out in prose in all four formats, the bare integer included, until the cap cut the reply off (400 of 460 bare-integer replies), so its fitted boundaries of 39 to 65 mark where the prose begins; under the catalogue protocol, with its cap of 8192 tokens, the same model has a CC of 320 to 424 (Table~\ref{tab:replicates}). \texttt{claude-sonnet-5} followed the bare-integer format in 159 of 160 replies, but in the other three formats it first stated the count as an integer in a sentence and then converted it (the format alone in 40, 19, and 12 of 160 replies), so those scans do not test the format. In the spaced-digit format three of the local models wrote the count as one number in the last field (\texttt{0 0 0 11}) from ten items onward, which leaves four configurations that followed it.

\begin{table}[htbp]
\centering
\footnotesize
\caption{\textbf{Counting capacity by output format.} Fitted CC on the baseline sequence for the seven configurations that followed the formats. A dash marks a format the model did not follow. The last-digit scan of \texttt{gemini-3.7-flash} was still mostly exact at the end of its range.}
\label{tab:format_controls}
\begin{tabular}{llrrrr}
\toprule
Model & Reasoning & Integer & Words & Spaced digits & Last digit \\
\midrule
\texttt{gemini-3.7-flash} & on & 367 & 295 & 297 & $\ge 479$ \\
\texttt{gpt-5.4} & off & 212 & 193 & 212 & 203 \\
\texttt{gemini-2.5-flash} & off & 60 & 77 & 100 & 100 \\
\texttt{gemma-4-26b-a4b-it} (local) & off & 75 & 75 & -- & 70 \\
\texttt{qwen3.6-35b-a3b} (local) & off & 42 & 37 & 50 & 34 \\
\texttt{llama-3.3-70b-instruct} (local) & off & 32 & 32 & -- & 33 \\
\texttt{gpt-oss-20b} (local) & off & 32 & 32 & -- & 8 \\
\bottomrule
\end{tabular}
\end{table}

\paragraph{The boundary.} Table~\ref{tab:format_controls} gives the fitted capacities. The format does not move the boundary in a consistent direction. With words the CC is 0.80 to 1.28 of its bare-integer value (median 1.00, rank correlation 0.96 across the seven). With spaced digits it is 0.81, 1.00, 1.19, and 1.67 of it in the four configurations that followed the format. With the last digit alone it is 0.81 to 1.03 of it in four configurations, 1.67 in \texttt{gemini-2.5-flash}, and at least 1.31 in \texttt{gemini-3.7-flash}; \texttt{gpt-oss-20b} fails the last-digit format itself, answering 0 at 53 of 70 lengths. Apart from \texttt{gemini-2.5-flash}, whose boundary is higher in every alternative format, these ratios are within the spread of repeated scans of one model (Appendix~\ref{app:replicates}).

\paragraph{The round offsets.} Of the seven, only \texttt{gemini-3.7-flash} makes errors of a round hundred; the others were scanned to at most 300 items, and in \texttt{gpt-5.4} no reply is 50 or 100 from the truth in any format. In \texttt{gemini-3.7-flash} the reply is exactly 100 below the true count in 32 of the 47 errors past the boundary with integers, in 31 of 62 with words (\texttt{one hundred ninety-seven} for 297), and in 42 of 58 with spaced digits (\texttt{0 1 9 9} for 299); four more replies in each format are 99 below. The last digit stays right throughout. Asked for the last digit alone, the model is exact in 233 of 240 probes up to 479, and in the bare-integer scan the last digit is right in 49 of the 56 replies past the boundary while the whole count is right in 9. An error of ten survives the change of format in a model that answers immediately: in \texttt{gemma-4-26b-a4b-it} the most common wrong reply is ten below the true count, in 11 of 49 errors with integers and 14 of 42 with words.

\paragraph{Scope of the test.} \texttt{gemini-3.7-flash} reasons before it answers, with a median of 1,800 reasoning tokens per probe, so the format changes only how a count the model has already formed is written. The result excludes the writing of the final answer as the source of the hundred-item error and leaves open a slip inside the reasoning text. None of the configurations that answer immediately reaches the lengths at which hundred-item errors appear. The same test on an immediate-answer model with a capacity of several hundred has not been run.

\section{Estimating counting capacity}\label{app:estimation}

Each measurement queries every sequence length once, in ascending order from \(L=1\), and reduces each reply to a single bit: whether the parsed integer equaled the target exactly. A reply containing no parsable integer is scored as inexact, since a model that cannot deliver an answer within the protocol's output budget is not stably counting under the protocol. One measurement is therefore a binary series \(x_1,\dots,x_n\) indexed by length, and CC is a summary of that series.

We do not define CC as the position of the first inexact answer. The first failure is the minimum of the failure positions, and it is therefore a property of the sampling density as much as of the model: probing the same model more densely can only move it earlier. It is also empirically unrepresentative, since 284 of our 296 measurements answer exactly at some length above their first inexact answer, indicating that the first failure is typically an isolated slip rather than a boundary.

We therefore estimate the boundary by fitting the full series. The model is that accuracy is \(p_1\) up to a transition \(\tau\) and \(p_2\) above it, subject to the requirement that the two regimes lie on opposite sides of one half,
\[
p_1 > \tfrac{1}{2} > p_2 ,
\]
so that the fitted boundary separates a regime in which exact counting is the model's majority behavior from one in which it is not. All three parameters are estimated by maximum likelihood, and CC is the fitted \(\hat{\tau}\). For a cut after the \(k\)th length, with \(c_1\) exact answers below and \(c_2\) above, each side takes its own estimate \(\hat p_i = c_i/n_i\) and the profile log-likelihood is the sum of two Bernoulli terms,
\[
\ell(k)=c_1\ln\hat p_1+(k-c_1)\ln(1-\hat p_1)+c_2\ln\hat p_2+(n-k-c_2)\ln(1-\hat p_2).
\]
Prefix sums make this \(O(1)\) per cut and \(O(n)\) for the whole profile.

The one-half requirement is the only structural choice in the fit, and it ensures that the fitted boundary is a counting boundary rather than a reliability boundary. Without it, the fit would separate a perfect prefix from a tail that is 95\% accurate, because such a drop is statistically detectable across several hundred probes, and the estimate would then report a small loss of reliability as the capacity. One half is also not a tuned threshold in the sense that a 90\% or 95\% accuracy floor would be: it is the unique accuracy value that is symmetric under exchanging the labels exact and inexact, and it therefore requires no further justification.

A transition is declared only when the improvement in fit justifies the two additional parameters, using the Bayesian information criterion: the two-regime fit must exceed the single-regime fit by more than \(\ln n\) in log-likelihood. This choice also removes the standard changepoint edge artifact, in which a stray failure near the end of a long record is split off as a spurious two-probe regime. Declared transitions in our data clear this bar by a median of 40 log-likelihood units against bars of 4.6 to 6.9, so the result does not depend on the criterion chosen. Where no transition is declared, the measurement is reported as a lower bound rather than a point estimate.

The confidence interval on \(\tau\) is a profile set: every admissible cut whose log-likelihood lies within a bar of the maximum. Because \(\tau\) is a discrete parameter, the chi-squared bar of 1.92 that a profile interval ordinarily uses is not calibrated for it: in simulation from the two-regime model with \(p_1=0.96\) and \(p_2=0.12\), a set built with that bar contains the true cut 86 to 87\% of the time at fixed scan lengths of 100 and 300, and 78 to 89\% under the sequential stopping rule described below. We therefore calibrate the bar for each measurement by parametric bootstrap: 200 records are drawn from the fitted \((\hat\tau,\hat p_1,\hat p_2)\) over the observed probe lengths, and the bar is the 95th percentile of the amount by which the best admissible cut's log-likelihood exceeds the true cut's, never less than 1.92. Under the same simulations this set contains the true cut 94\% of the time at fixed scan lengths and 90 to 92\% under the stopping rule, the residual shortfall arising from scans that stop with only a few probes above the boundary. Across our measurements the calibrated interval has a median width of four lengths, and 152 of the 294 fitted transitions (the two \texttt{minimax-m1} records, which ended in provider errors, enter no statistic) span five lengths or fewer, 138 of them among the 256 resolved by wrong answers. This interval is a statement about the estimate rather than about the underlying process: given several hundred binary probes and a contrast as large as \(p_1=0.96\) against \(p_2=0.12\), the location of the best cut is determined precisely. The measured rate itself falls more gradually. When the five rescans of each model in Appendix~\ref{app:replicates} are pooled and smoothed over \(\pm 5\) lengths, accuracy passes from 0.8 to 0.2 over a median of 26 lengths, about one tenth of CC (0.05 to 0.43; Table~\ref{tab:transition_width}), so \(\hat{\tau}\) locates a transition of real width rather than an instant at which counting stops. The two-regime form summarizes this transition and does not claim that the underlying rate is a step; a claim about how abrupt the collapse is would require a model that estimates a transition width, which the two-regime form deliberately omits. The intervals are drawn as the transition bands in Fig.~\ref{fig:framework}c.

\begin{table}[htbp]
\centering
\footnotesize
\caption{\textbf{Width of the behavioral transition.} Per-length accuracy pooled over the five rescans of Appendix~\ref{app:replicates} and smoothed over $\pm 5$ lengths; the width is the span from the last length at which smoothed accuracy is at least 0.8 to the first at which it is at most 0.2.}
\label{tab:transition_width}
\begin{tabular}{lrrrrr}
\toprule
Model & CC & last $\ge 0.8$ & first $\le 0.2$ & width & width / CC \\
\midrule
\texttt{claude-fable-5} & 317 & 313 & 345 & 32 & 0.10 \\
\texttt{claude-opus-5} & 179 & -- & -- & -- & -- \\
\texttt{claude-sonnet-4.6} & 320 & 396 & 426 & 30 & 0.09 \\
\texttt{claude-sonnet-5} & 92 & 86 & 104 & 18 & 0.20 \\
\texttt{gemini-3.7-flash} & 352 & 347 & 391 & 44 & 0.12 \\
\texttt{gpt-5} & 101 & 98 & 107 & 9 & 0.09 \\
\texttt{gpt-5.2} & 152 & 133 & 159 & 26 & 0.17 \\
\texttt{gpt-5.4} & 208 & 205 & 216 & 11 & 0.05 \\
\texttt{gpt-5.6-sol} & 96 & 93 & 134 & 41 & 0.43 \\
\texttt{o3} & 103 & 98 & 107 & 9 & 0.09 \\
\bottomrule
\end{tabular}
\end{table}

Because the fit uses the whole record rather than an extreme of it, the estimate is stable against how densely the model was probed. Refitting every measurement on every second probe moves \(\hat{\tau}\) by a median of \(-1\) length, and on every fifth probe by \(-2\); the position of the first inexact answer moves by \(+4\) lengths (mean \(+12\)) under the same subsampling.

Scanning continues until the capacity is determined: a scan stops once a transition is declared and the accuracy above it is significantly below one half, as assessed by the upper end of the 95\% Wilson interval on \(\hat p_2\). The stopping rule therefore coincides with the estimator, and a measurement ends when it has determined the quantity being measured rather than after a fixed probe budget. Scans that instead terminate on ten consecutive unparsable replies are reported as bounded by the output budget rather than by counting, and are excluded from statements about counting limits.

\subsection{Bounds on guessing exploits}\label{app:guessing}

The CC measurement scores a reply as exact or inexact, with no tolerance band, so approximately correct answers receive no partial credit. This section states what a guessing strategy can achieve against that criterion, and against the two-regime fit built on top of it.

\subsubsection{Guessing against a single length}

Suppose a model retains no exact count but recognizes the approximate scale of the prompt, and returns a value drawn from some distribution $Q$ concentrated near the true count $L$. Because credit requires $\hat{x}=L$ exactly, the probability of success at that length is $Q(\{L\})$. A strategy that spreads its mass over a window of $w$ plausible values can do no better than $\max_x Q(\{x\})$, and a strategy that returns a fixed value $c$ succeeds at exactly one length in the entire scan. There is no analogue of the error-minimizing guess that a relative-error criterion admits: under exact matching, approximate knowledge of the magnitude provides no credit.

\subsubsection{Guessing against the fitted boundary}

The stronger statement concerns the estimator rather than a single trial. A transition is declared only if some cut separates a regime with $\hat p_1 > \tfrac{1}{2}$ from one with $\hat p_2 < \tfrac{1}{2}$. Establishing a stable regime of $k$ lengths therefore requires exact answers at more than $k/2$ distinct lengths.

Consider the most favorable guessing strategy available to a model with no exact count: at each length it guesses independently, placing probability at most $q<\tfrac{1}{2}$ on the correct value; a guess drawn uniformly from a window of $w$ values centered on the correct value has $q=1/w$, and a non-uniform window is covered by taking $q$ as the largest probability it places on any value. The number of successes over $k$ lengths is stochastically dominated by $S \sim \mathrm{Binomial}(k, q)$, and a Chernoff bound gives
\begin{equation}
\Pr\!\left[S > \tfrac{k}{2}\right] \le \exp\!\left(-k\,D\!\left(\tfrac{1}{2}\,\middle\|\,q\right)\right),
\end{equation}
where $D(a\|b)=a\ln\frac{a}{b}+(1-a)\ln\frac{1-a}{1-b}$ is the binary relative entropy. Even for a window as narrow as $w=4$, $D(\tfrac{1}{2}\|\tfrac{1}{4})\approx 0.1438$, so a guessing model has probability below $10^{-3}$ of producing an apparent stable regime of $k=49$ lengths, and below $10^{-6}$ at $k=97$. The median CC in our data is 50 and the largest is 627. A window wide enough to be plausible for a model that has genuinely lost the count, say $w=20$, reduces the same probabilities below $10^{-3}$ at $k=9$ and below $10^{-6}$ at $k=17$.

The bound is conservative in two further respects. It grants the guessing model a window centered on the correct value at every length, and it ignores the requirement that the regime above the boundary be significantly \textit{worse} than one half, which a scale-guessing strategy with a fixed window would not satisfy: its accuracy does not degrade with length, so it produces no transition to declare. In practice a model that has lost the count does not guess within a narrow window at all; it collapses onto a small set of round attractors (Fig.~\ref{fig:attractors}D, E), which succeed at a handful of lengths in the entire scan. The bound concerns a guessing strategy producing an apparently stable prefix; it is not a guarantee on the full estimator, whose cut selection and stopping rule are checked by simulation above.

\subsubsection{Implications}

Exact matching eliminates the class of exploit that a tolerance-based criterion must defend against, and the majority requirement in the fit converts that per-length guarantee into an exponential bound on fabricating a stable regime. Under the stated independence and success-probability assumptions the bound therefore limits fabrication of a wholly guessed stable prefix; it does not bound inflation of an existing capacity estimate by guessed answers in the tail.

\FloatBarrier

\section{Model set, catalogue, and data release}\label{app:catalogue}

We evaluated 205 language models, proprietary and open-weight, comprising 296 measurements once reasoning-enabled and reasoning-disabled configurations of the same model are counted separately. The set included instruction-tuned and reasoning-augmented variants spanning a broad range of parameter counts and nominal context lengths. This range allowed us to test whether CC reflects a general capability axis rather than a quirk of a particular tokenizer, serving stack, or model family.

All catalogue evaluations were conducted with tool use disabled. For externally served models, we recorded the model identifier, evaluation date, decoding configuration, token-usage metadata, parsed response, target count, and success or failure classification. These records were retained to account for possible changes in served model behavior over time. The complete proprietary-model evaluation cost approximately US\$200 in API spend, excluding local open-weight mechanistic analyses.

Table~\ref{tab:model_inventory} lists the full model catalogue together with the corresponding developer or family, parameter metadata, nominal context window, architectural class, and measured counting capacity (CC). The two \texttt{minimax-m1} measurements ended in provider errors after 0 and 10 items; they are listed as bounds and enter no statistic. Table~\ref{tab:cohorts} gives the release-year cohorts behind the longitudinal comparison in the introduction: resolved measurements grouped by release year, with cohort sizes, median CC, and the cohorts' median advertised context window.

\begin{table}[htbp]
\centering
\footnotesize
\caption{\textbf{Release-year cohorts behind the longitudinal comparison.} Resolved measurements (transition bounded by wrong answers) grouped by the model's release year; reasoning configurations counted separately. Median CC with interquartile range, and the median advertised context window of the same measurements. The 2023 cohort is four models.}
\label{tab:cohorts}
\begin{tabular}{lrrrrr}
\toprule
Release year & Measurements & Models & Median CC & IQR & Median context (tokens) \\
\midrule
2023 & 4 & 4 & 35 & [27, 44] & 12,288 \\
2024 & 22 & 22 & 27 & [21, 36] & 128,000 \\
2025 & 106 & 82 & 44 & [32, 84] & 230,400 \\
2026 & 124 & 77 & 56 & [43, 115] & 1,000,000 \\
\bottomrule
\end{tabular}
\end{table}

\begingroup
\scriptsize
\setlength{\LTcapwidth}{\textwidth}
\setlength{\tabcolsep}{3pt}
\renewcommand{\arraystretch}{1.05}
\begin{longtable}{@{}p{4.6cm}p{1.5cm}p{2.0cm}p{1.1cm}p{1.0cm}p{1.9cm}@{}}
\caption{\textbf{Full model catalogue.} One row per measurement; hybrid
models appear once per reasoning configuration. CC is the fitted transition
with its calibrated interval (nominal 95\%; Appendix~\ref{app:estimation}); capacities marked $\geq$ are lower bounds, set by
the output budget rather than by a counting failure. Parameter counts give, or unresolved where the bound column says so.
total and, for mixture models, active parameters. Metadata are drawn from the
OpenRouter catalogue and public documentation.}\label{tab:model_inventory}\\
\toprule
Measurement & Params & Context & CC & 95\% CI & Bounded by \\
\midrule
\endfirsthead
\toprule
Measurement & Params & Context & CC & 95\% CI & Bounded by \\
\midrule
\endhead
\bottomrule
\endfoot
\texttt{claude-\allowbreak{}3-\allowbreak{}haiku} & -- & 200k & 40 & [39, 50] & mistake \\
\texttt{claude-\allowbreak{}fable-\allowbreak{}5} & -- & 1M & 317 & [316, 329] & mistake \\
\texttt{claude-\allowbreak{}haiku-\allowbreak{}4.5 \#nothink} & -- & 200k & 132 & [130, 135] & mistake \\
\texttt{claude-\allowbreak{}haiku-\allowbreak{}4.5 \#think} & -- & 200k & 131 & [130, 131] & mistake \\
\texttt{claude-\allowbreak{}opus-\allowbreak{}4 \#nothink} & -- & 200k & 580 & [575, 608] & mistake \\
\texttt{claude-\allowbreak{}opus-\allowbreak{}4 \#think} & -- & 200k & 627 & [626, 628] & mistake \\
\texttt{claude-\allowbreak{}opus-\allowbreak{}4.1 \#nothink} & -- & 200k & 580 & [550, 582] & mistake \\
\texttt{claude-\allowbreak{}opus-\allowbreak{}4.1 \#think} & -- & 200k & 573 & [567, 574] & mistake \\
\texttt{claude-\allowbreak{}opus-\allowbreak{}4.5 \#nothink} & -- & 200k & 580 & [578, 582] & mistake \\
\texttt{claude-\allowbreak{}opus-\allowbreak{}4.5 \#think} & -- & 200k & 620 & [618, 622] & mistake \\
\texttt{claude-\allowbreak{}opus-\allowbreak{}4.6 \#nothink} & -- & 1M & 618 & [618, 622] & mistake \\
\texttt{claude-\allowbreak{}opus-\allowbreak{}4.6 \#think} & -- & 1M & 622 & [620, 623] & mistake \\
\texttt{claude-\allowbreak{}opus-\allowbreak{}4.7 \#nothink} & -- & 1M & 158 & [157, 158] & mistake \\
\texttt{claude-\allowbreak{}opus-\allowbreak{}4.7 \#think} & -- & 1M & 156 & [140, 156] & mistake \\
\texttt{claude-\allowbreak{}opus-\allowbreak{}4.8 \#nothink} & -- & 1M & 133 & [132, 133] & mistake \\
\texttt{claude-\allowbreak{}opus-\allowbreak{}4.8 \#think} & -- & 1M & 109 & [108, 109] & mistake \\
\texttt{claude-\allowbreak{}opus-\allowbreak{}5 \#nothink} & -- & 1M & 179 & [131, 180] & mistake \\
\texttt{claude-\allowbreak{}opus-\allowbreak{}5 \#think} & -- & 1M & 132 & [129, 162] & mistake \\
\texttt{claude-\allowbreak{}sonnet-\allowbreak{}4 \#nothink} & -- & 1M & 382 & [362, 382] & mistake \\
\texttt{claude-\allowbreak{}sonnet-\allowbreak{}4 \#think} & -- & 1M & 290 & [289, 293] & mistake \\
\texttt{claude-\allowbreak{}sonnet-\allowbreak{}4.5 \#nothink} & -- & 1M & 364 & [364, 364] & mistake \\
\texttt{claude-\allowbreak{}sonnet-\allowbreak{}4.5 \#think} & -- & 1M & 364 & [359, 364] & mistake \\
\texttt{claude-\allowbreak{}sonnet-\allowbreak{}4.6 \#nothink} & -- & 1M & 320 & [315, 320] & mistake \\
\texttt{claude-\allowbreak{}sonnet-\allowbreak{}4.6 \#think} & -- & 1M & $\geq$264 & [259, 265] & budget \\
\texttt{claude-\allowbreak{}sonnet-\allowbreak{}5 \#nothink} & -- & 1M & 92 & [91, 99] & mistake \\
\texttt{claude-\allowbreak{}sonnet-\allowbreak{}5 \#think} & -- & 1M & 89 & [85, 91] & mistake \\
\texttt{deepseek-\allowbreak{}chat} & 684.531B & 163k & 30 & [15, 44] & mistake \\
\texttt{deepseek-\allowbreak{}chat-\allowbreak{}v3-\allowbreak{}0324} & 684.531B & 163k & 30 & [25, 31] & mistake \\
\texttt{deepseek-\allowbreak{}chat-\allowbreak{}v3.1 \#nothink} & 684.531B / 37B & 163k & 15 & [14, 15] & mistake \\
\texttt{deepseek-\allowbreak{}chat-\allowbreak{}v3.1 \#think} & 684.531B / 37B & 163k & $\geq$28 & [13, 63] & budget \\
\texttt{deepseek-\allowbreak{}r1} & 684.531B / 37B & 64k & $\geq$67 & [65, 81] & budget \\
\texttt{deepseek-\allowbreak{}r1-\allowbreak{}0528} & 684.531B / 37B & 163k & $\geq$71 & [69, 77] & budget \\
\texttt{deepseek-\allowbreak{}r1-\allowbreak{}distill-\allowbreak{}llama-\allowbreak{}70b \#nothink} & 70.554B / 70B & 8k & $\geq$57 & [33, 66] & budget \\
\texttt{deepseek-\allowbreak{}r1-\allowbreak{}distill-\allowbreak{}llama-\allowbreak{}70b \#think} & 70.554B / 70B & 8k & $\geq$58 & [56, 60] & budget \\
\texttt{deepseek-\allowbreak{}v3.1-\allowbreak{}terminus \#nothink} & 684.531B & 163k & 13 & [11, 13] & mistake \\
\texttt{deepseek-\allowbreak{}v3.1-\allowbreak{}terminus \#think} & 684.531B & 163k & $\geq$60 & [51, 62] & budget \\
\texttt{deepseek-\allowbreak{}v3.2 \#nothink} & 685.397B & 163k & 12 & [12, 12] & mistake \\
\texttt{deepseek-\allowbreak{}v3.2 \#think} & 685.397B & 163k & $\geq$70 & [53, 71] & budget \\
\texttt{deepseek-\allowbreak{}v3.2-\allowbreak{}exp \#nothink} & 685.397B & 163k & 10 & [9, 12] & mistake \\
\texttt{deepseek-\allowbreak{}v3.2-\allowbreak{}exp \#think} & 685.397B & 163k & $\geq$58 & [53, 61] & budget \\
\texttt{deepseek-\allowbreak{}v4-\allowbreak{}flash \#nothink} & 290.945B / 13B & 1.04858M & 52 & [51, 63] & mistake \\
\texttt{deepseek-\allowbreak{}v4-\allowbreak{}flash \#think} & 290.945B / 13B & 1.04858M & $\geq$72 & [67, 86] & budget \\
\texttt{deepseek-\allowbreak{}v4-\allowbreak{}flash-\allowbreak{}0731 \#nothink} & 304.18B / 13B & 1.31072M & 44 & [34, 56] & mistake \\
\texttt{deepseek-\allowbreak{}v4-\allowbreak{}flash-\allowbreak{}0731 \#think} & 304.18B / 13B & 1.31072M & $\geq$51 & [47, 96] & budget \\
\texttt{deepseek-\allowbreak{}v4-\allowbreak{}flash-\allowbreak{}vision-\allowbreak{}exp \#nothink} & -- & 1.04858M & 139 & [136, 140] & mistake \\
\texttt{deepseek-\allowbreak{}v4-\allowbreak{}flash-\allowbreak{}vision-\allowbreak{}exp \#think} & -- & 1.04858M & $\geq$76 & [74, 89] & budget \\
\texttt{deepseek-\allowbreak{}v4-\allowbreak{}pro \#nothink} & 1598.84B / 49B & 1.04858M & 36 & [34, 38] & mistake \\
\texttt{deepseek-\allowbreak{}v4-\allowbreak{}pro \#think} & 1598.84B / 49B & 1.04858M & $\geq$72 & [69, 88] & budget \\
\texttt{deepseek-\allowbreak{}v4-\allowbreak{}pro-\allowbreak{}0813 \#nothink} & 1650.5B & 1.04858M & 36 & [35, 36] & mistake \\
\texttt{deepseek-\allowbreak{}v4-\allowbreak{}pro-\allowbreak{}0813 \#think} & 1650.5B & 1.04858M & $\geq$96 & [85, 97] & budget \\
\texttt{gemini-\allowbreak{}2.5-\allowbreak{}flash \#nothink} & -- & 1.04858M & 102 & [99, 103] & mistake \\
\texttt{gemini-\allowbreak{}2.5-\allowbreak{}flash \#think} & -- & 1.04858M & 107 & [100, 108] & mistake \\
\texttt{gemini-\allowbreak{}2.5-\allowbreak{}flash-\allowbreak{}lite \#nothink} & -- & 1.04858M & 79 & [54, 80] & mistake \\
\texttt{gemini-\allowbreak{}2.5-\allowbreak{}flash-\allowbreak{}lite \#think} & -- & 1.04858M & 54 & [50, 54] & mistake \\
\texttt{gemini-\allowbreak{}2.5-\allowbreak{}pro} & -- & 1.04858M & 119 & [105, 125] & mistake \\
\texttt{gemini-\allowbreak{}2.5-\allowbreak{}pro-\allowbreak{}preview} & -- & 1.04858M & 120 & [113, 123] & mistake \\
\texttt{gemini-\allowbreak{}2.5-\allowbreak{}pro-\allowbreak{}preview-\allowbreak{}05-\allowbreak{}06} & -- & 1.04858M & 119 & [108, 124] & mistake \\
\texttt{gemini-\allowbreak{}3-\allowbreak{}flash-\allowbreak{}preview \#nothink} & -- & 1.04858M & 445 & [435, 447] & mistake \\
\texttt{gemini-\allowbreak{}3-\allowbreak{}flash-\allowbreak{}preview \#think} & -- & 1.04858M & 404 & [398, 421] & mistake \\
\texttt{gemini-\allowbreak{}3.1-\allowbreak{}flash-\allowbreak{}lite \#nothink} & -- & 1.04858M & 130 & [128, 133] & mistake \\
\texttt{gemini-\allowbreak{}3.1-\allowbreak{}flash-\allowbreak{}lite \#think} & -- & 1.04858M & 115 & [112, 130] & mistake \\
\texttt{gemini-\allowbreak{}3.1-\allowbreak{}flash-\allowbreak{}lite-\allowbreak{}preview \#nothink} & -- & 1.04858M & 130 & [129, 130] & mistake \\
\texttt{gemini-\allowbreak{}3.1-\allowbreak{}flash-\allowbreak{}lite-\allowbreak{}preview \#think} & -- & 1.04858M & 130 & [129, 133] & mistake \\
\texttt{gemini-\allowbreak{}3.1-\allowbreak{}pro-\allowbreak{}preview} & -- & 1.04858M & 518 & [518, 519] & mistake \\
\texttt{gemini-\allowbreak{}3.1-\allowbreak{}pro-\allowbreak{}preview-\allowbreak{}customtools} & -- & 1.04858M & 429 & [429, 429] & mistake \\
\texttt{gemini-\allowbreak{}3.5-\allowbreak{}flash} & -- & 1.04858M & 298 & [289, 299] & mistake \\
\texttt{gemini-\allowbreak{}3.5-\allowbreak{}flash-\allowbreak{}lite} & -- & 1.04858M & 146 & [142, 149] & mistake \\
\texttt{gemini-\allowbreak{}3.6-\allowbreak{}flash} & -- & 1.04858M & 347 & [316, 348] & mistake \\
\texttt{gemini-\allowbreak{}3.7-\allowbreak{}flash} & -- & 1.04858M & 352 & [350, 373] & mistake \\
\texttt{gemma-\allowbreak{}2-\allowbreak{}27b-\allowbreak{}it} & 27.227B / 27B & 8k & 21 & [21, 21] & mistake \\
\texttt{gemma-\allowbreak{}3-\allowbreak{}12b-\allowbreak{}it} & 12.187B / 12B & 131k & 32 & [15, 39] & mistake \\
\texttt{gemma-\allowbreak{}3-\allowbreak{}27b-\allowbreak{}it} & 27.432B / 27B & 262k & 36 & [30, 37] & mistake \\
\texttt{gemma-\allowbreak{}3-\allowbreak{}4b-\allowbreak{}it} & 4.3B / 4B & 131k & 20 & [20, 27] & mistake \\
\texttt{gemma-\allowbreak{}3n-\allowbreak{}e4b-\allowbreak{}it} & 7.85B / 4B & 32k & 17 & [16, 22] & mistake \\
\texttt{gemma-\allowbreak{}4-\allowbreak{}26b-\allowbreak{}a4b-\allowbreak{}it \#nothink} & 25.806B / 4B & 262k & 83 & [53, 84] & mistake \\
\texttt{gemma-\allowbreak{}4-\allowbreak{}26b-\allowbreak{}a4b-\allowbreak{}it \#think} & 25.806B / 4B & 262k & $\geq$41 & [40, 50] & budget \\
\texttt{gemma-\allowbreak{}4-\allowbreak{}31b-\allowbreak{}it \#nothink} & 31.273B / 31B & 262k & 38 & [28, 40] & mistake \\
\texttt{gemma-\allowbreak{}4-\allowbreak{}31b-\allowbreak{}it \#think} & 31.273B / 31B & 262k & 45 & [35, 46] & mistake \\
\texttt{muse-\allowbreak{}glimmer-\allowbreak{}30b} & 29.777B / 30B & 131k & 54 & [54, 54] & mistake \\
\texttt{muse-\allowbreak{}spark-\allowbreak{}1.1} & -- & 1.04858M & 195 & [192, 206] & mistake \\
\texttt{muse-\allowbreak{}spark-\allowbreak{}1.2} & -- & 1.04858M & $\geq$96 & [90, 98] & budget \\
\texttt{muse-\allowbreak{}spark-\allowbreak{}1.2-\allowbreak{}contributor} & -- & 1.04858M & $\geq$96 & [91, 109] & budget \\
\texttt{llama-\allowbreak{}3.1-\allowbreak{}70b-\allowbreak{}instruct} & 70.554B / 70B & 131k & 16 & [16, 25] & mistake \\
\texttt{llama-\allowbreak{}3.1-\allowbreak{}8b-\allowbreak{}instruct} & 8.03B / 8B & 131k & 21 & [20, 24] & mistake \\
\texttt{llama-\allowbreak{}3.2-\allowbreak{}1b-\allowbreak{}instruct} & 1.236B / 1B & 60k & 2 & [2, 3] & mistake \\
\texttt{llama-\allowbreak{}3.2-\allowbreak{}3b-\allowbreak{}instruct} & 3.213B / 3B & 131k & 24 & [8, 25] & mistake \\
\texttt{llama-\allowbreak{}3.3-\allowbreak{}70b-\allowbreak{}instruct} & 70.554B / 70B & 131k & 27 & [27, 27] & mistake \\
\texttt{llama-\allowbreak{}4-\allowbreak{}maverick} & 401.584B / 17B & 1.04858M & 55 & [54, 57] & mistake \\
\texttt{llama-\allowbreak{}4-\allowbreak{}scout} & 108.642B / 17B & 1.31072M & 22 & [21, 22] & mistake \\
\texttt{minimax-\allowbreak{}01} & 456.09B & 1.00019M & 37 & [32, 40] & mistake \\
\texttt{minimax-\allowbreak{}m1 \#nothink} & -- & 1M & $\geq$0 & [0, 0] & budget \\
\texttt{minimax-\allowbreak{}m1 \#think} & -- & 1M & $\geq$10 & -- & unresolved \\
\texttt{minimax-\allowbreak{}m2} & 228.69B / 10B & 204k & 37 & [36, 37] & mistake \\
\texttt{minimax-\allowbreak{}m2-\allowbreak{}her} & -- & 65k & 18 & [17, 19] & mistake \\
\texttt{minimax-\allowbreak{}m2.1} & 228.69B / 10B & 204k & 40 & [39, 40] & mistake \\
\texttt{minimax-\allowbreak{}m2.5} & 228.704B & 204k & 39 & [39, 39] & mistake \\
\texttt{minimax-\allowbreak{}m2.7} & 228.69B & 204k & $\geq$40 & [40, 40] & budget \\
\texttt{minimax-\allowbreak{}m3 \#nothink} & 427.04B & 1.04858M & 51 & [47, 56] & mistake \\
\texttt{minimax-\allowbreak{}m3 \#think} & 427.04B & 1.04858M & 56 & [55, 65] & mistake \\
\texttt{codestral-\allowbreak{}2508} & -- & 256k & 48 & [31, 49] & mistake \\
\texttt{ministral-\allowbreak{}14b-\allowbreak{}2512} & 13.945B / 14B & 262k & 31 & [30, 35] & mistake \\
\texttt{ministral-\allowbreak{}3b-\allowbreak{}2512} & 3.849B / 3B & 131k & 19 & [19, 19] & mistake \\
\texttt{ministral-\allowbreak{}8b-\allowbreak{}2512} & 8.918B / 8B & 262k & 21 & [21, 31] & mistake \\
\texttt{mistral-\allowbreak{}large} & -- & 128k & 31 & [31, 31] & mistake \\
\texttt{mistral-\allowbreak{}large-\allowbreak{}2407} & -- & 131k & 31 & [31, 31] & mistake \\
\texttt{mistral-\allowbreak{}large-\allowbreak{}2512} & -- & 262k & 31 & [31, 31] & mistake \\
\texttt{mistral-\allowbreak{}medium-\allowbreak{}3} & -- & 131k & 35 & [35, 40] & mistake \\
\texttt{mistral-\allowbreak{}medium-\allowbreak{}3-\allowbreak{}5 \#nothink} & -- & 262k & 45 & [45, 45] & mistake \\
\texttt{mistral-\allowbreak{}medium-\allowbreak{}3-\allowbreak{}5 \#think} & -- & 262k & 31 & [28, 31] & mistake \\
\texttt{mistral-\allowbreak{}medium-\allowbreak{}3.1} & -- & 131k & 39 & [39, 39] & mistake \\
\texttt{mistral-\allowbreak{}nemo} & 12.248B & 131k & 20 & [20, 29] & mistake \\
\texttt{mistral-\allowbreak{}saba} & -- & 32k & 32 & [20, 33] & mistake \\
\texttt{mistral-\allowbreak{}small-\allowbreak{}24b-\allowbreak{}instruct-\allowbreak{}2501} & 23.572B / 24B & 32k & 29 & [29, 30] & mistake \\
\texttt{mistral-\allowbreak{}small-\allowbreak{}2603 \#nothink} & 119.401B / 119B & 262k & 32 & [30, 34] & mistake \\
\texttt{mistral-\allowbreak{}small-\allowbreak{}2603 \#think} & 119.401B / 119B & 262k & 45 & [44, 52] & mistake \\
\texttt{mistral-\allowbreak{}small-\allowbreak{}3.2-\allowbreak{}24b-\allowbreak{}instruct} & 24.011B / 24B & 131k & 31 & [31, 32] & mistake \\
\texttt{mixtral-\allowbreak{}8x22b-\allowbreak{}instruct} & 140.63B / 39B & 65k & 32 & [20, 33] & mistake \\
\texttt{voxtral-\allowbreak{}small-\allowbreak{}24b-\allowbreak{}2507} & 24.262B / 24B & 32k & 28 & [28, 29] & mistake \\
\texttt{kimi-\allowbreak{}k2} & 1026.41B / 32B & 131k & 76 & [64, 76] & mistake \\
\texttt{kimi-\allowbreak{}k2-\allowbreak{}0905} & 1026.47B & 262k & 72 & [63, 92] & mistake \\
\texttt{kimi-\allowbreak{}k2-\allowbreak{}thinking} & 1058.12B & 262k & 70 & [45, 73] & mistake \\
\texttt{kimi-\allowbreak{}k2.5 \#nothink} & 1026.88B & 262k & 72 & [59, 74] & mistake \\
\texttt{kimi-\allowbreak{}k2.5 \#think} & 1026.88B & 262k & 73 & [70, 95] & mistake \\
\texttt{kimi-\allowbreak{}k2.6 \#nothink} & 1026.88B & 262k & 76 & [63, 77] & mistake \\
\texttt{kimi-\allowbreak{}k2.6 \#think} & 1026.88B & 262k & $\geq$56 & [54, 69] & budget \\
\texttt{kimi-\allowbreak{}k2.7-\allowbreak{}code} & 1026.88B & 262k & $\geq$70 & [50, 74] & budget \\
\texttt{kimi-\allowbreak{}k3 \#nothink} & 2779.93B & 1.04858M & 54 & [53, 54] & mistake \\
\texttt{kimi-\allowbreak{}k3 \#think} & 2779.93B & 1.04858M & 51 & [50, 51] & mistake \\
\texttt{gpt-\allowbreak{}3.5-\allowbreak{}turbo} & 20B & 16k & 25 & [25, 25] & mistake \\
\texttt{gpt-\allowbreak{}3.5-\allowbreak{}turbo-\allowbreak{}0613} & -- & 4k & 50 & [45, 52] & mistake \\
\texttt{gpt-\allowbreak{}3.5-\allowbreak{}turbo-\allowbreak{}16k} & -- & 16k & 34 & [33, 38] & mistake \\
\texttt{gpt-\allowbreak{}3.5-\allowbreak{}turbo-\allowbreak{}instruct} & 20B & 4k & 36 & [25, 39] & mistake \\
\texttt{gpt-\allowbreak{}4} & -- & 8k & 46 & [38, 53] & mistake \\
\texttt{gpt-\allowbreak{}4-\allowbreak{}turbo} & -- & 128k & 25 & [20, 26] & mistake \\
\texttt{gpt-\allowbreak{}4.1} & -- & 1.04758M & 54 & [52, 54] & mistake \\
\texttt{gpt-\allowbreak{}4.1-\allowbreak{}mini} & -- & 1.04758M & 38 & [31, 53] & mistake \\
\texttt{gpt-\allowbreak{}4.1-\allowbreak{}nano} & -- & 1.04758M & 46 & [42, 51] & mistake \\
\texttt{gpt-\allowbreak{}4o} & -- & 128k & 50 & [46, 53] & mistake \\
\texttt{gpt-\allowbreak{}4o-\allowbreak{}2024-\allowbreak{}05-\allowbreak{}13} & -- & 128k & 42 & [42, 43] & mistake \\
\texttt{gpt-\allowbreak{}4o-\allowbreak{}2024-\allowbreak{}08-\allowbreak{}06} & -- & 128k & 48 & [34, 53] & mistake \\
\texttt{gpt-\allowbreak{}4o-\allowbreak{}2024-\allowbreak{}11-\allowbreak{}20} & -- & 128k & 34 & [32, 34] & mistake \\
\texttt{gpt-\allowbreak{}4o-\allowbreak{}mini} & -- & 128k & 25 & [24, 26] & mistake \\
\texttt{gpt-\allowbreak{}4o-\allowbreak{}mini-\allowbreak{}2024-\allowbreak{}07-\allowbreak{}18} & -- & 128k & 24 & [23, 24] & mistake \\
\texttt{gpt-\allowbreak{}5} & -- & 400k & 101 & [67, 104] & mistake \\
\texttt{gpt-\allowbreak{}5-\allowbreak{}mini} & -- & 400k & 60 & [51, 61] & mistake \\
\texttt{gpt-\allowbreak{}5-\allowbreak{}nano} & -- & 400k & 35 & [34, 35] & mistake \\
\texttt{gpt-\allowbreak{}5.1 \#nothink} & -- & 400k & 101 & [81, 102] & mistake \\
\texttt{gpt-\allowbreak{}5.1 \#think} & -- & 400k & 101 & [101, 102] & mistake \\
\texttt{gpt-\allowbreak{}5.1-\allowbreak{}codex} & -- & 400k & $\geq$100 & [99, 101] & budget \\
\texttt{gpt-\allowbreak{}5.1-\allowbreak{}codex-\allowbreak{}max} & -- & 400k & $\geq$43 & [41, 50] & budget \\
\texttt{gpt-\allowbreak{}5.1-\allowbreak{}codex-\allowbreak{}mini \#nothink} & -- & 400k & $\geq$50 & [50, 51] & budget \\
\texttt{gpt-\allowbreak{}5.1-\allowbreak{}codex-\allowbreak{}mini \#think} & -- & 400k & $\geq$50 & [47, 51] & budget \\
\texttt{gpt-\allowbreak{}5.2 \#nothink} & -- & 400k & 113 & [110, 114] & mistake \\
\texttt{gpt-\allowbreak{}5.2 \#think} & -- & 400k & 152 & [146, 153] & mistake \\
\texttt{gpt-\allowbreak{}5.2-\allowbreak{}codex} & -- & 400k & 115 & [96, 119] & mistake \\
\texttt{gpt-\allowbreak{}5.3-\allowbreak{}codex \#nothink} & -- & 400k & 210 & [209, 210] & mistake \\
\texttt{gpt-\allowbreak{}5.3-\allowbreak{}codex \#think} & -- & 400k & 223 & [223, 223] & mistake \\
\texttt{gpt-\allowbreak{}5.4 \#nothink} & -- & 1.05M & 208 & [208, 208] & mistake \\
\texttt{gpt-\allowbreak{}5.4 \#think} & -- & 1.05M & 174 & [166, 176] & mistake \\
\texttt{gpt-\allowbreak{}5.4-\allowbreak{}mini \#nothink} & -- & 400k & 57 & [55, 57] & mistake \\
\texttt{gpt-\allowbreak{}5.4-\allowbreak{}mini \#think} & -- & 400k & 57 & [56, 64] & mistake \\
\texttt{gpt-\allowbreak{}5.4-\allowbreak{}nano \#nothink} & -- & 400k & 31 & [26, 32] & mistake \\
\texttt{gpt-\allowbreak{}5.4-\allowbreak{}nano \#think} & -- & 400k & 27 & [27, 27] & mistake \\
\texttt{gpt-\allowbreak{}5.5 \#nothink} & -- & 1.05M & 105 & [100, 120] & mistake \\
\texttt{gpt-\allowbreak{}5.5 \#think} & -- & 1.05M & 96 & [93, 98] & mistake \\
\texttt{gpt-\allowbreak{}5.6-\allowbreak{}luna \#nothink} & -- & 1.05M & 46 & [43, 47] & mistake \\
\texttt{gpt-\allowbreak{}5.6-\allowbreak{}luna \#think} & -- & 1.05M & $\geq$45 & [44, 52] & budget \\
\texttt{gpt-\allowbreak{}5.6-\allowbreak{}luna-\allowbreak{}pro \#nothink} & -- & 1.05M & 46 & [44, 47] & mistake \\
\texttt{gpt-\allowbreak{}5.6-\allowbreak{}luna-\allowbreak{}pro \#think} & -- & 1.05M & 46 & [37, 47] & mistake \\
\texttt{gpt-\allowbreak{}5.6-\allowbreak{}sol \#nothink} & -- & 1.05M & 94 & [92, 94] & mistake \\
\texttt{gpt-\allowbreak{}5.6-\allowbreak{}sol \#think} & -- & 1.05M & 96 & [95, 96] & mistake \\
\texttt{gpt-\allowbreak{}5.6-\allowbreak{}sol-\allowbreak{}pro \#nothink} & -- & 1.05M & 94 & [91, 97] & mistake \\
\texttt{gpt-\allowbreak{}5.6-\allowbreak{}sol-\allowbreak{}pro \#think} & -- & 1.05M & 93 & [86, 93] & mistake \\
\texttt{gpt-\allowbreak{}5.6-\allowbreak{}terra \#nothink} & -- & 1.05M & 117 & [115, 117] & mistake \\
\texttt{gpt-\allowbreak{}5.6-\allowbreak{}terra \#think} & -- & 1.05M & 115 & [113, 117] & mistake \\
\texttt{gpt-\allowbreak{}5.6-\allowbreak{}terra-\allowbreak{}pro \#nothink} & -- & 1.05M & 174 & [174, 175] & mistake \\
\texttt{gpt-\allowbreak{}5.6-\allowbreak{}terra-\allowbreak{}pro \#think} & -- & 1.05M & 120 & [112, 121] & mistake \\
\texttt{gpt-\allowbreak{}chat-\allowbreak{}latest} & -- & 400k & 76 & [67, 78] & mistake \\
\texttt{gpt-\allowbreak{}oss-\allowbreak{}120b} & 116.829B / 5.1B & 131k & 101 & [101, 101] & mistake \\
\texttt{gpt-\allowbreak{}oss-\allowbreak{}20b} & 21.512B / 3.6B & 131k & 58 & [57, 58] & mistake \\
\texttt{gpt-\allowbreak{}oss-\allowbreak{}safeguard-\allowbreak{}20b} & 21.512B / 3.6B & 131k & 57 & [52, 61] & mistake \\
\texttt{o1 \#think} & -- & 200k & $\geq$66 & [46, 72] & budget \\
\texttt{o3 \#think} & -- & 200k & 103 & [103, 103] & mistake \\
\texttt{o3-\allowbreak{}mini \#think} & -- & 200k & 37 & [37, 37] & mistake \\
\texttt{o3-\allowbreak{}mini-\allowbreak{}high} & -- & 200k & $\geq$37 & [32, 40] & budget \\
\texttt{o4-\allowbreak{}mini \#think} & -- & 200k & $\geq$50 & [49, 64] & budget \\
\texttt{o4-\allowbreak{}mini-\allowbreak{}high} & -- & 200k & $\geq$47 & [46, 50] & budget \\
\texttt{qwen-\allowbreak{}2.5-\allowbreak{}72b-\allowbreak{}instruct} & 72.706B / 72B & 32k & 27 & [27, 27] & mistake \\
\texttt{qwen-\allowbreak{}2.5-\allowbreak{}7b-\allowbreak{}instruct} & 7.616B / 7B & 32k & 12 & [12, 15] & mistake \\
\texttt{qwen-\allowbreak{}plus} & -- & 1M & 33 & [32, 33] & mistake \\
\texttt{qwen-\allowbreak{}plus-\allowbreak{}2025-\allowbreak{}07-\allowbreak{}28 \#nothink} & -- & 1M & 54 & [53, 54] & mistake \\
\texttt{qwen-\allowbreak{}plus-\allowbreak{}2025-\allowbreak{}07-\allowbreak{}28 \#think} & -- & 1M & 54 & [53, 54] & mistake \\
\texttt{qwen-\allowbreak{}plus-\allowbreak{}2025-\allowbreak{}07-\allowbreak{}28:thinking \#nothink} & -- & 1M & 58 & [58, 60] & mistake \\
\texttt{qwen-\allowbreak{}plus-\allowbreak{}2025-\allowbreak{}07-\allowbreak{}28:thinking \#think} & -- & 1M & 58 & [58, 60] & mistake \\
\texttt{qwen2.5-\allowbreak{}vl-\allowbreak{}72b-\allowbreak{}instruct} & 73.411B / 72B & 128k & 30 & [30, 32] & mistake \\
\texttt{qwen3-\allowbreak{}14b \#nothink} & 14.768B / 14B & 131k & $\geq$27 & [26, 30] & budget \\
\texttt{qwen3-\allowbreak{}14b \#think} & 14.768B / 14B & 131k & $\geq$31 & [28, 32] & budget \\
\texttt{qwen3-\allowbreak{}235b-\allowbreak{}a22b \#nothink} & 235.094B / 22B & 131k & 45 & [44, 45] & mistake \\
\texttt{qwen3-\allowbreak{}235b-\allowbreak{}a22b \#think} & 235.094B / 22B & 131k & 43 & [42, 50] & mistake \\
\texttt{qwen3-\allowbreak{}235b-\allowbreak{}a22b-\allowbreak{}2507} & 235.094B / 22B & 262k & 54 & [50, 60] & mistake \\
\texttt{qwen3-\allowbreak{}235b-\allowbreak{}a22b-\allowbreak{}thinking-\allowbreak{}2507} & 235.094B / 22B & 262k & $\geq$57 & [54, 58] & budget \\
\texttt{qwen3-\allowbreak{}30b-\allowbreak{}a3b \#nothink} & 30.532B / 3B & 131k & 37 & [32, 40] & mistake \\
\texttt{qwen3-\allowbreak{}30b-\allowbreak{}a3b \#think} & 30.532B / 3B & 131k & 35 & [35, 35] & mistake \\
\texttt{qwen3-\allowbreak{}30b-\allowbreak{}a3b-\allowbreak{}instruct-\allowbreak{}2507} & 30.532B / 3B & 262k & 35 & [30, 41] & mistake \\
\texttt{qwen3-\allowbreak{}30b-\allowbreak{}a3b-\allowbreak{}thinking-\allowbreak{}2507} & 30.532B / 3B & 81k & 40 & [40, 45] & mistake \\
\texttt{qwen3-\allowbreak{}32b \#nothink} & 32.762B / 32B & 131k & 22 & [22, 22] & mistake \\
\texttt{qwen3-\allowbreak{}32b \#think} & 32.762B / 32B & 131k & 26 & [26, 33] & mistake \\
\texttt{qwen3-\allowbreak{}8b \#nothink} & 8.191B / 8B & 131k & 25 & [24, 26] & mistake \\
\texttt{qwen3-\allowbreak{}8b \#think} & 8.191B / 8B & 131k & 28 & [27, 30] & mistake \\
\texttt{qwen3-\allowbreak{}coder} & 480.155B / 35B & 262k & 30 & [29, 30] & mistake \\
\texttt{qwen3-\allowbreak{}coder-\allowbreak{}30b-\allowbreak{}a3b-\allowbreak{}instruct} & 30.532B / 3B & 262k & 33 & [31, 38] & mistake \\
\texttt{qwen3-\allowbreak{}coder-\allowbreak{}flash} & -- & 1M & 39 & [32, 40] & mistake \\
\texttt{qwen3-\allowbreak{}coder-\allowbreak{}next} & 79.674B / 3B & 262k & 50 & [49, 50] & mistake \\
\texttt{qwen3-\allowbreak{}coder-\allowbreak{}plus \#nothink} & -- & 1M & 50 & [49, 54] & mistake \\
\texttt{qwen3-\allowbreak{}coder-\allowbreak{}plus \#think} & -- & 1M & 50 & [49, 56] & mistake \\
\texttt{qwen3-\allowbreak{}max \#nothink} & 1000B & 262k & 60 & [60, 60] & mistake \\
\texttt{qwen3-\allowbreak{}max \#think} & 1000B & 262k & 60 & [50, 61] & mistake \\
\texttt{qwen3-\allowbreak{}max-\allowbreak{}thinking \#nothink} & -- & 262k & 55 & [49, 55] & mistake \\
\texttt{qwen3-\allowbreak{}max-\allowbreak{}thinking \#think} & -- & 262k & 55 & [55, 55] & mistake \\
\texttt{qwen3-\allowbreak{}next-\allowbreak{}80b-\allowbreak{}a3b-\allowbreak{}instruct} & 81.325B / 3B & 262k & 37 & [37, 37] & mistake \\
\texttt{qwen3-\allowbreak{}next-\allowbreak{}80b-\allowbreak{}a3b-\allowbreak{}thinking} & 81.325B / 3B & 262k & 40 & [40, 40] & mistake \\
\texttt{qwen3-\allowbreak{}vl-\allowbreak{}235b-\allowbreak{}a22b-\allowbreak{}instruct} & 235.67B / 22B & 262k & 50 & [49, 51] & mistake \\
\texttt{qwen3-\allowbreak{}vl-\allowbreak{}235b-\allowbreak{}a22b-\allowbreak{}thinking} & 235.67B / 22B & 131k & $\geq$55 & [55, 55] & budget \\
\texttt{qwen3-\allowbreak{}vl-\allowbreak{}30b-\allowbreak{}a3b-\allowbreak{}instruct} & 31.071B / 3B & 262k & 36 & [36, 37] & mistake \\
\texttt{qwen3-\allowbreak{}vl-\allowbreak{}30b-\allowbreak{}a3b-\allowbreak{}thinking} & 31.071B / 3B & 262k & 44 & [42, 45] & mistake \\
\texttt{qwen3-\allowbreak{}vl-\allowbreak{}32b-\allowbreak{}instruct} & 33.357B / 32B & 131k & 28 & [28, 30] & mistake \\
\texttt{qwen3-\allowbreak{}vl-\allowbreak{}8b-\allowbreak{}instruct} & 8.767B / 8B & 262k & 21 & [21, 25] & mistake \\
\texttt{qwen3-\allowbreak{}vl-\allowbreak{}8b-\allowbreak{}thinking} & 8.767B / 8B & 131k & $\geq$30 & [30, 30] & budget \\
\texttt{qwen3.5-\allowbreak{}122b-\allowbreak{}a10b \#nothink} & 125.086B / 10B & 262k & 51 & [51, 58] & mistake \\
\texttt{qwen3.5-\allowbreak{}122b-\allowbreak{}a10b \#think} & 125.086B / 10B & 262k & 50 & [50, 52] & mistake \\
\texttt{qwen3.5-\allowbreak{}27b \#nothink} & 27.781B / 27B & 262k & 40 & [39, 40] & mistake \\
\texttt{qwen3.5-\allowbreak{}27b \#think} & 27.781B / 27B & 262k & 41 & [41, 41] & mistake \\
\texttt{qwen3.5-\allowbreak{}35b-\allowbreak{}a3b \#nothink} & 35.952B / 3B & 262k & 42 & [42, 42] & mistake \\
\texttt{qwen3.5-\allowbreak{}35b-\allowbreak{}a3b \#think} & 35.952B / 3B & 262k & $\geq$50 & [46, 54] & budget \\
\texttt{qwen3.5-\allowbreak{}397b-\allowbreak{}a17b \#nothink} & 403.398B / 17B & 262k & 72 & [72, 73] & mistake \\
\texttt{qwen3.5-\allowbreak{}397b-\allowbreak{}a17b \#think} & 403.398B / 17B & 262k & 72 & [70, 72] & mistake \\
\texttt{qwen3.5-\allowbreak{}9b \#nothink} & 9.653B / 9B & 262k & 35 & [35, 36] & mistake \\
\texttt{qwen3.5-\allowbreak{}9b \#think} & 9.653B / 9B & 262k & $\geq$25 & [25, 25] & budget \\
\texttt{qwen3.5-\allowbreak{}flash-\allowbreak{}02-\allowbreak{}23 \#nothink} & -- & 1M & 42 & [41, 42] & mistake \\
\texttt{qwen3.5-\allowbreak{}flash-\allowbreak{}02-\allowbreak{}23 \#think} & -- & 1M & 46 & [45, 54] & mistake \\
\texttt{qwen3.5-\allowbreak{}plus-\allowbreak{}02-\allowbreak{}15 \#nothink} & -- & 1M & 72 & [72, 73] & mistake \\
\texttt{qwen3.5-\allowbreak{}plus-\allowbreak{}02-\allowbreak{}15 \#think} & -- & 1M & 72 & [62, 72] & mistake \\
\texttt{qwen3.5-\allowbreak{}plus-\allowbreak{}20260420 \#nothink} & -- & 1M & 45 & [45, 45] & mistake \\
\texttt{qwen3.5-\allowbreak{}plus-\allowbreak{}20260420 \#think} & -- & 1M & 47 & [46, 55] & mistake \\
\texttt{qwen3.6-\allowbreak{}27b \#nothink} & 27.781B / 27B & 262k & 43 & [42, 43] & mistake \\
\texttt{qwen3.6-\allowbreak{}27b \#think} & 27.781B / 27B & 262k & 42 & [42, 42] & mistake \\
\texttt{qwen3.6-\allowbreak{}35b-\allowbreak{}a3b \#nothink} & 35.952B / 3B & 262k & 42 & [37, 51] & mistake \\
\texttt{qwen3.6-\allowbreak{}35b-\allowbreak{}a3b \#think} & 35.952B / 3B & 262k & 45 & [44, 53] & mistake \\
\texttt{qwen3.6-\allowbreak{}flash \#nothink} & -- & 1M & 42 & [37, 45] & mistake \\
\texttt{qwen3.6-\allowbreak{}flash \#think} & -- & 1M & 43 & [42, 43] & mistake \\
\texttt{qwen3.6-\allowbreak{}max-\allowbreak{}preview \#nothink} & -- & 262k & 68 & [68, 68] & mistake \\
\texttt{qwen3.6-\allowbreak{}max-\allowbreak{}preview \#think} & -- & 262k & 69 & [69, 69] & mistake \\
\texttt{qwen3.6-\allowbreak{}plus \#nothink} & -- & 1M & 50 & [49, 50] & mistake \\
\texttt{qwen3.6-\allowbreak{}plus \#think} & -- & 1M & 53 & [53, 56] & mistake \\
\texttt{qwen3.7-\allowbreak{}flash \#nothink} & -- & 1M & 37 & [36, 42] & mistake \\
\texttt{qwen3.7-\allowbreak{}flash \#think} & -- & 1M & 50 & [50, 53] & mistake \\
\texttt{qwen3.7-\allowbreak{}max \#nothink} & -- & 1M & 41 & [39, 50] & mistake \\
\texttt{qwen3.7-\allowbreak{}max \#think} & -- & 1M & 51 & [50, 51] & mistake \\
\texttt{qwen3.7-\allowbreak{}plus \#nothink} & -- & 1M & 47 & [45, 50] & mistake \\
\texttt{qwen3.7-\allowbreak{}plus \#think} & -- & 1M & 57 & [56, 61] & mistake \\
\texttt{qwen3.8-\allowbreak{}2.4t-\allowbreak{}a95b} & 2446.18B / 95B & 1.04858M & 70 & [70, 70] & mistake \\
\texttt{qwen3.8-\allowbreak{}27b \#nothink} & 27.781B / 27B & 1M & 38 & [37, 41] & mistake \\
\texttt{qwen3.8-\allowbreak{}27b \#think} & 27.781B / 27B & 1M & 38 & [38, 40] & mistake \\
\texttt{qwen3.8-\allowbreak{}max} & 2400B & 1M & 70 & [69, 70] & mistake \\
\texttt{grok-\allowbreak{}4.20 \#nothink} & 500B & 2M & 84 & [73, 101] & mistake \\
\texttt{grok-\allowbreak{}4.20 \#think} & 500B & 2M & 85 & [66, 106] & mistake \\
\texttt{grok-\allowbreak{}4.3 \#nothink} & -- & 1M & 100 & [96, 104] & mistake \\
\texttt{grok-\allowbreak{}4.3 \#think} & -- & 1M & 102 & [101, 104] & mistake \\
\texttt{grok-\allowbreak{}4.5} & -- & 500k & 140 & [139, 140] & mistake \\
\texttt{grok-\allowbreak{}4.6} & -- & 500k & 226 & [225, 227] & mistake \\
\texttt{grok-\allowbreak{}build-\allowbreak{}0.1} & -- & 256k & 52 & [49, 53] & mistake \\
\texttt{glm-\allowbreak{}4.5 \#nothink} & 358.338B & 131k & 41 & [40, 42] & mistake \\
\texttt{glm-\allowbreak{}4.5 \#think} & 358.338B & 131k & 43 & [43, 43] & mistake \\
\texttt{glm-\allowbreak{}4.5-\allowbreak{}air \#nothink} & 110.469B & 131k & 48 & [43, 50] & mistake \\
\texttt{glm-\allowbreak{}4.5-\allowbreak{}air \#think} & 110.469B & 131k & 53 & [53, 53] & mistake \\
\texttt{glm-\allowbreak{}4.5v \#nothink} & 107.711B / 12B & 65k & 48 & [41, 50] & mistake \\
\texttt{glm-\allowbreak{}4.5v \#think} & 107.711B / 12B & 65k & 42 & [41, 44] & mistake \\
\texttt{glm-\allowbreak{}4.6 \#nothink} & 356.786B & 204k & 41 & [35, 42] & mistake \\
\texttt{glm-\allowbreak{}4.6 \#think} & 356.786B & 204k & $\geq$53 & [50, 54] & budget \\
\texttt{glm-\allowbreak{}4.6v \#nothink} & 107.711B & 131k & 22 & [22, 22] & mistake \\
\texttt{glm-\allowbreak{}4.6v \#think} & 107.711B & 131k & 21 & [21, 23] & mistake \\
\texttt{glm-\allowbreak{}4.7 \#nothink} & 358.338B & 204k & 40 & [35, 42] & mistake \\
\texttt{glm-\allowbreak{}4.7 \#think} & 358.338B & 204k & 41 & [41, 43] & mistake \\
\texttt{glm-\allowbreak{}4.7-\allowbreak{}flash \#nothink} & 31.221B & 202k & 15 & [15, 15] & mistake \\
\texttt{glm-\allowbreak{}4.7-\allowbreak{}flash \#think} & 31.221B & 202k & 30 & [30, 30] & mistake \\
\texttt{glm-\allowbreak{}5 \#nothink} & 753.864B & 204k & 40 & [31, 50] & mistake \\
\texttt{glm-\allowbreak{}5 \#think} & 753.864B & 204k & 40 & [40, 40] & mistake \\
\texttt{glm-\allowbreak{}5-\allowbreak{}turbo \#nothink} & -- & 202k & 40 & [21, 50] & mistake \\
\texttt{glm-\allowbreak{}5-\allowbreak{}turbo \#think} & -- & 202k & 38 & [37, 41] & mistake \\
\texttt{glm-\allowbreak{}5.1 \#nothink} & 753.864B & 204k & 50 & [35, 51] & mistake \\
\texttt{glm-\allowbreak{}5.1 \#think} & 753.864B & 204k & 40 & [39, 40] & mistake \\
\texttt{glm-\allowbreak{}5.2 \#nothink} & 753.33B & 1.04858M & 41 & [39, 42] & mistake \\
\texttt{glm-\allowbreak{}5.2 \#think} & 753.33B & 1.04858M & 53 & [49, 54] & mistake \\
\texttt{glm-\allowbreak{}5.3} & 744B & 1.04858M & $\geq$58 & [55, 59] & budget \\
\texttt{glm-\allowbreak{}5v-\allowbreak{}turbo \#nothink} & -- & 202k & 50 & [45, 51] & mistake \\
\texttt{glm-\allowbreak{}5v-\allowbreak{}turbo \#think} & -- & 202k & 42 & [42, 42] & mistake \\
\end{longtable}

\endgroup

\FloatBarrier

\section{Instruction-following audit}\label{app:instruction}

The measurement instructs every model to return a single integer with no other text. When we audited every scan probe for compliance, 3{,}887 of 45{,}287 responses (8.6\%) deviated from this format. Deviations concentrate beyond each model's counting boundary, indicating that losing the count disrupts not only numerical accuracy but also the control needed to maintain the requested response format.

Table~\ref{tab:audit_categories} details the distribution of these failure modes. Blank or whitespace-only outputs are the most common deviation, and inspection of the response logs shows that most of them are budget exhaustion: 2{,}041 of the deviating responses terminated at the fixed output cap with no final answer emitted, the signature of a reasoning stream that never terminates. Unsolicited step-by-step counting is the second most common deviation, in which a model abandons the requested zero-shot format and attempts to count in generated text. The remaining categories are prompt echoes, code or markdown formatting, leaked reasoning markup, and answers embedded in prose.

Table~\ref{tab:audit_models} lists the models deviating most often, as a share of their own probes. The heaviest deviators are reasoning-configured models probed far beyond their boundary: one frontier model accounts for 712 deviations across its two configurations, and one small reasoning model deviates on 88\% of its probes. Blank outputs are therefore partly a budget artifact of reasoning configurations, several of them from one provider whose extended thinking counts against the cap; the claim that losing the count disrupts format control rests on the remaining categories, which concentrate beyond the boundary in non-reasoning configurations as well. Representative responses are reproduced in Table~\ref{tab:audit_snippets}: after losing the count, models copy the sequence back, wrap the answer in code fences, leak reasoning scaffolds into the reply, or append conversational text to the number.

\begin{table}[htbp]\centering\scriptsize
\caption{Distribution of non-single-number responses across all 45,287 scan probes (3,887 deviating responses, 8.6\%).}
\label{tab:audit_categories}
\begin{tabular}{@{}l r r@{}}
\toprule
Category & Count & Share of probes \\
\midrule
Blank / whitespace & 2,183 & 4.82\% \\
Step-by-step counting or explanation & 1,401 & 3.09\% \\
Reasoning-tag or hidden-CoT markup & 63 & 0.14\% \\
Code / markdown formatted output & 61 & 0.13\% \\
Answer with extra prose / instruction leakage & 49 & 0.11\% \\
Prompt echo / copied sequence & 41 & 0.09\% \\
Other mixed numeric format & 35 & 0.08\% \\
Number with stray punctuation / formatting & 33 & 0.07\% \\
Text with no parseable number & 21 & 0.05\% \\
\bottomrule
\end{tabular}
\end{table}

\begin{table}[htbp]\centering\scriptsize
\caption{Models most frequently deviating from the single-integer format,
with each model's deviations as a share of its own probes.}
\label{tab:audit_models}
\begin{tabular}{@{}l r r@{}}
\toprule
Model & Deviations & Share of own probes \\
\midrule
\texttt{claude-opus-4.6} & 712 & 35\% \\
\texttt{claude-sonnet-4.6} & 317 & 51\% \\
\texttt{claude-sonnet-4.5} & 132 & 18\% \\
\texttt{qwen3.5-flash-02-23} & 100 & 50\% \\
\texttt{minimax-m2-her} & 88 & 88\% \\
\texttt{claude-opus-4.1} & 76 & 5\% \\
\texttt{deepseek-v4-pro-0813} & 76 & 27\% \\
\texttt{gpt-5.1-codex} & 72 & 38\% \\
\texttt{gpt-3.5-turbo-instruct} & 67 & 67\% \\
\texttt{grok-4.20} & 66 & 29\% \\
\bottomrule
\end{tabular}
\end{table}

\begin{table}[htbp]\centering\scriptsize
\caption{Representative deviating responses; L is the probed length.}
\label{tab:audit_snippets}
\begin{tabular}{@{}l l r p{5.0cm}@{}}
\toprule
Category & Model & L & Response \\
\midrule
Step-by-step counting & \texttt{claude-sonnet-5} & 55 & Let me count them one by one.\textbackslash n\textbackslash n55 \\
Code / markdown output & \texttt{gemini-3-flash-preview} & 303 & \textasciigrave\textasciigrave\textasciigrave json 303 \textasciigrave\textasciigrave\textasciigrave \\
Prompt echo & \texttt{mistral-medium-3-5} & 49 & a, a, a, a, a, a, a a, a, a, \ldots{} (sequence copied back) \\
Leaked reasoning markup & \texttt{qwen3.5-flash-02-23} & 5 & Thinking Process: 1. Analyze the Request: Task: Count occurrences\ldots \\
Mixed numeric format & \texttt{deepseek-v4-pro-0813} & 110 & 110.110 \\
Prose / instruction leakage & \texttt{minimax-m2-her} & 97 & 87 \textbackslash n user name=talkior \textbackslash n No. Try again. \\
\bottomrule
\end{tabular}
\end{table}

\FloatBarrier

\section{Anchored-execution control}\label{app:anchored}

The counting failures in the main text could in principle reflect a general inability to execute long sequential procedures. This control shows that they do not. We constructed four subsidiary tasks whose intermediate value, unlike a count over identical items, can be expressed in language at every step, and measured each alone under the counting protocol: the same models, output budget, stop rules, and estimator, with the task's load taking the place of sequence length.

The four tasks are ordered by the intermediate value they require. \emph{Lookup} provides a table of word-to-number pairs and asks for one entry; the answer requires a single attention reference and no intermediate value. \emph{Pointer chase} provides a shuffled chain of ``$w_i$ points to $w_j$'' links and asks for the number associated with the final word, requiring as many dependent references as the chain is long. \emph{Running total} provides a start value and a list of signed increments to apply in order. \emph{Automaton} provides the transition table of a four-state machine and an input word stream, and asks for the final state. Keys and symbols are common English words drawn from the run's seeded generator, so no rare-token effects enter (Appendix~\ref{app:prompts}), and the correct answer is never equal to the load, so a task cannot be solved by counting its own input lines. Lookup was probed at table sizes 50 to 1000 in steps of 50; the other tasks were probed densely from load 1 to a ceiling of 320, chosen to exceed the counting capacities of two of the three models and to reach nine tenths of the third's (352).

Table~\ref{tab:anchored_control} gives the result. Lookup is exact at every table size to 1000 for all three models. The three sequential tasks are executed essentially without collapse over the probed range: seven of nine measurements had not reached a transition at the probe limit, and of the two declared transitions, the one at 299 links belongs to the model with a CC of 89 and the one at 276 updates to the model with a CC of 352, the latter within a fifth of its counting capacity. The model with a counting capacity of 89 follows pointer chains of roughly 300 links, and the one with a capacity of 115 executes 320 automaton transitions at 97\% accuracy.

\begin{table}[htbp]
\centering
\scriptsize
\caption{Anchored-execution control. Each cell gives the fitted capacity of
one subsidiary task, with the stable-regime accuracy in parentheses;
$\geq$ marks measurements that had not collapsed at the probe ceiling.
Counting capacities for the same three models are 89, 115, and 352.}
\label{tab:anchored_control}
\begin{tabular}{@{}l l r r r@{}}
\toprule
Level & Task & \texttt{claude-sonnet-5} & \texttt{gpt-5.6-terra} & \texttt{gemini-3.7-flash} \\
\midrule
1 & Lookup & $\geq 1000$ (1.00) & $\geq 1000$ (1.00) & $\geq 1000$ (1.00) \\
2 & Pointer chase & 299 (0.94) & $\geq 320$ (0.99) & $\geq 320$ (1.00) \\
3 & Running total & $\geq 320$ (1.00) & $\geq 320$ (0.95) & 276 (0.89) \\
4 & Automaton & $\geq 320$ (0.88) & $\geq 320$ (0.97) & $\geq 320$ (1.00) \\
\bottomrule
\end{tabular}
\end{table}

The contrast narrows what the counting bound covers. These tasks demand long sequential execution, dependent references, and order-sensitive values, and the models perform them several times beyond their counting capacities. What distinguishes counting is not that its running value cannot be written, since a count is one integer, but that a written value cannot be re-attached to the item where it was written: each update line, pointer link, or input word is a distinct textual anchor that a reasoning trace can name, process, and verify, whereas the items of a homogeneous sequence are indistinguishable, so a trace cannot express its own position without re-deriving the count. The bound measured by CC is therefore specific to running values that cannot be tied back to the input once written, which is also why enabling reasoning leaves the median counting capacity unchanged while multiplying token expenditure (Fig.~\ref{fig:resources}B): a chain of thought can only extend values it can name.

\FloatBarrier

\section{Verbatim copying through a tool call}\label{app:copy}

The counting failure could in principle lie in the arithmetic rather than in the representation of the sequence. This control removes the arithmetic. The model is given the counting measurement's sequence and a single tool, \texttt{record\_sequence}, and is instructed to call the tool with the sequence copied verbatim; the tool call is forced, and the number of items in the recorded argument is compared with the number in the prompt. No counting is requested, and generation can re-anchor on the source at every emitted token, so this is the easiest traversal of the sequence the model can be asked to perform.

Copy fidelity is perfect well beyond each model's counting capacity and then collapses at a finite length, fitted with the estimator of Appendix~\ref{app:estimation}. One model copies exactly to a boundary of 420 against a CC of 352 and fails by deleting whole round blocks, so that recorded sequences run 50 to 200 items short; the other copies exactly to 448 against a CC of 89 and fails in the opposite direction, duplicating content until the record runs up to 475 items long. Both signatures are the counting taxonomy's own failure modes, the round-block offsets of Fig.~\ref{fig:attractors}D and the runaway generation of Appendix~\ref{app:reasoning}, now appearing in a task with no arithmetic in it.

The copy boundary is 1.2 and 5.0 times CC respectively, consistent with the landmark account: reproduction supplies a per-token anchor that pure counting lacks, and the more a task lets generation re-anchor, the further it extends. The boundary is finite and far below the context window, and beyond it a model handed a sequence cannot pass it onward intact, so any downstream computation that consumes the data, counting included, receives a corrupted payload. The limit lies in reproducing exact multiplicity, whether it arises while reading the input, while tracking position in it, or while deciding when to stop generating, which this control does not separate; the copy and counting boundaries also differ, so the two tasks do not fail at one length. The practical form of the failure for agents is that beyond a few hundred repeated items, a tool-call payload of the very data in the prompt is silently wrong. A tool interface that passes a reference to the stored input rather than a regenerated payload avoids this by construction, so the result limits what a model can do with data it must itself reproduce.

\FloatBarrier

\section{Structure of reasoning-trace failures}\label{app:reasoning}

To examine why additional test-time computation does not rescue counting, we retrieved the stored reasoning traces of the scan itself from the serving provider's generation archive, rather than running a separate probe set. The archive returns the full stored record, including the reasoning channel, for 117 of the 122 reasoning-enabled measurements; the remaining five expose no reasoning text. We sampled up to sixteen generations per model, spaced evenly over prompt size, and identified each record by parsing its stored prompt, giving 1{,}728 counting probes with reasoning attached. This retrieval path is independent of the original run records and reproduces the two-regime structure of Appendix~\ref{app:estimation}.

Every incorrect trace (697) and every correct trace near or above capacity (305) was read and classified by the strategy the trace attempts, the structure of its failure, and whether the returned integer can be derived from the visible reasoning. Correct traces below capacity were sampled rather than exhaustively read; they are short executed enumerations or terse assertions, and their surface form does not separate them from failures a few dozen lengths later: several models emit the same template on both sides of the boundary. Exposed reasoning is behavioral evidence rather than a faithful record of internal computation, and for some providers it is a summarized paraphrase of a hidden trace, so the classification describes what the trace displays.

The visible trace rarely computes the answer. In 434 of the 697 incorrect traces the returned integer cannot be derived from the visible reasoning: the trace is narration without arithmetic, is truncated, contradicts the answer it accompanies, or ratifies a number that appeared before any counting began. The same decoupling holds for successes. Of the 80 correct answers above capacity, 30 contain no executed enumeration at all, and several models are explicit about it. One frontier model, correct at 198 items with a fitted capacity of 132, abandons its tally at 93 and writes ``Without a precise way to verify by actually splitting the string, I'll settle on 198 as my count.'' The count was available without the counting; the trace documents a readout, not a computation. The converse also occurs well below capacity: another model, wrong at 37 items against a fitted capacity of 119, spends its whole trace asserting diligence (``I've performed multiple checks to ensure the accuracy of this count, which now stands at 36'') without a digit of visible work.

Table~\ref{tab:trace_outcomes} organizes the 697 incorrect traces into four groups. In the first, the answer is decoupled from the trace outright: either a candidate value is named before any counting and every subsequent check is generated to agree with it (including the agreement of supposedly independent methods, which all issue from the same decoder and therefore confirm rather than test), or the trace is method narration containing no work to check. In the second and largest group, enumeration is executed but measures the model's own generation rather than the prompt: the trace re-emits the sequence or a numbered ladder and reports where its own copy stops. Its mildest form counts a silently drifted transcription; once detached from the input, the ladder either runs away (one trace enumerates to 1{,}000 before remarking ``Okay, that was a simulation of counting to 1000. I need to count the \emph{actual} string provided'') or stops at a round total, which is the trace-level mechanism of the numerical attractors in Fig.~\ref{fig:attractors}E. The third group contains genuine counts that slip: the interior blocks of a grouping scheme are handled correctly and the final partial block is mis-sized, producing the off-by-$k$ errors of Fig.~\ref{fig:attractors}D, or a valid identity such as items-equals-commas-plus-one is applied to an operand the model never measured, so correct algebra is applied to a guessed value. The fourth group fails to commit or to deliver: the trace re-derives the count in near-identical passes, sometimes reaching the correct value and then abandoning it, until the budget is exhausted, or the computed integer never reaches the visible completion, left in the reasoning channel or lost to formatting. These are the blank and budget-bound outputs of the instruction-following audit (Appendix~\ref{app:instruction}) seen from inside.

\begin{table}[htbp]
\centering
\small
\caption{\textbf{Failure structure of 697 incorrect reasoning traces.} Stored traces retrieved for 117 reasoning-enabled measurements and classified as described in the text. Groups and modes are assigned per trace; each trace counts once.}
\label{tab:trace_outcomes}
\begin{tabular}{llr}
\toprule
Group & Mode & Traces \\
\midrule
\multirow{2}{*}{Answer decoupled from trace}
  & anchored answer, ratifying verification & 57 \\
  & narration without work & 47 \\
\midrule
\multirow{3}{*}{Enumeration of own generation}
  & counts a drifted self-transcription & 26 \\
  & runaway ladder past the true end & 96 \\
  & terminates at a round attractor & 125 \\
\midrule
\multirow{2}{*}{Slips in genuine counts}
  & mis-sized final block (off-by-$k$) & 95 \\
  & valid formula, unmeasured operand & 56 \\
\midrule
\multirow{2}{*}{Commitment and delivery}
  & verification livelock to the budget & 140 \\
  & answer never reaches the completion & 55 \\
\bottomrule
\end{tabular}
\end{table}

Token expenditure runs the wrong way for a simple compute-bound account, though only as a descriptive pattern, since a hard probe can both spend more and fail. Above capacity, the median correct trace spends about 2{,}100 reasoning tokens, the median wrong-integer trace about 2{,}700, and the median trace returning no integer about 6{,}400, nearly all of the last group terminating at the token cap. Long scratchpads do not recover a collapsed count; they elaborate it.

This structure is consistent with the anchored-execution control (Appendix~\ref{app:anchored}) and explains the matched-pair null of Appendix~\ref{app:pairs}. A chain of thought can extend any running value it can name, and enumeration inside a trace is itself generation: its stopping point is set by whatever per-item total the model still holds, and when that state is exhausted the ladder defaults to the model's priors, a round total, a remembered schema, or the continuation of its own list. Naming the counting rule, which the traces do almost universally, does not supply the total the rule needs.

\subsection{Every hybrid pair, reasoning on against off}\label{app:pairs}

The study includes 90 hybrid pairs, that is, the same weights measured once with reasoning enabled at medium effort and once with reasoning disabled. Whether disabling took effect is verified from the measured reasoning tokens rather than assumed from the request, because several providers apply reasoning regardless of the requested setting. Only the 77 verified contrasts, listed below, enter Fig.~\ref{fig:resources}B. The change in capacity is heterogeneous around its zero median: the interquartile range is \(-3\) to \(+7\) items, seven pairs gain more than half their capacity and two lose more than a third. Sixteen of the 77 pairs have a reasoning member whose scan ended on exhaustion of the fixed output budget rather than on wrong answers, so under the catalogue protocol its capacity is a lower bound; these concentrate in one vendor line. To see whether the budget was what bound them, we rescanned exactly those sixteen configurations with the cap raised fourfold, to 32{,}768 tokens, and everything else unchanged (Table~\ref{tab:budget32k}). Six resolve at wrong answers under the larger budget; eight still end on budget exhaustion and two at the scan's end, so these ten remain budget-bounded observations. The eight spend the full 32{,}768 tokens on sequences of 40 to 100 items, an order of magnitude more than enumeration would need: the verification livelock of Table~\ref{tab:trace_outcomes} at a larger budget. Substituting the rescanned values for the bounds would move the median change over the 77 pairs from 0 to \(+1\) item (39 pairs gaining, 10 unchanged, 28 losing); the catalogue and Fig.~\ref{fig:resources}B keep the fixed-budget protocol. The gains concentrate in the members of that line whose silent capacities are 10 to 15 items and whose reasoning capacities reach 70 to 85 at the larger budget, and the largest loss is a model whose reasoning traces exhaust the budget at every length beyond 40. Raising the cap leaves the median contrast nearly unchanged, though ten of the sixteen rescans still do not identify an uncensored reasoning capacity.

\begin{table}[htbp]
\centering
\scriptsize
\setlength{\tabcolsep}{3pt}
\caption{\textbf{Budget-bounded reasoning configurations rescanned at a 32,768-token cap.} The sixteen verified hybrid pairs whose reasoning member exhausted the catalogue's 8,192-token cap before failing on the count, rescanned with the cap raised fourfold and everything else unchanged. CC$_{\text{off}}$ is the no-reasoning capacity, CC$_{\text{8k}}$ the catalogue bound, CC$_{\text{32k}}$ the rescanned value. Bound: how the 32k scan ended. Hits: probes that exhausted 32,768 tokens, and the smallest count at which it first happened.}
\label{tab:budget32k}
\begin{tabular}{lrrlrr}
\toprule
Model & CC$_{\text{off}}$ & CC$_{\text{8k}}$ & CC$_{\text{32k}}$ [95\% CI] & Bound & Hits (first at) \\
\midrule
\texttt{claude-sonnet-4.6} & 320 & $\geq$264 & $\geq$331 [331, 332] & budget & 5 (294) \\
\texttt{deepseek-chat-v3.1} & 15 & $\geq$28 & $\geq$84 [84, 96] & budget & 3 (55) \\
\texttt{deepseek-v3.1-terminus} & 13 & $\geq$60 & $\geq$70 [9, 75] & budget & 9 (66) \\
\texttt{deepseek-v3.2-exp} & 10 & $\geq$58 & $\geq$80 [80, 80] & scan end & 2 (68) \\
\texttt{deepseek-v3.2} & 12 & $\geq$70 & $\geq$80 [80, 80] & scan end & 1 (66) \\
\texttt{deepseek-v4-flash-0731} & 44 & $\geq$51 & $\geq$64 [52, 70] & budget & 16 (39) \\
\texttt{deepseek-v4-flash-vision-exp} & 139 & $\geq$76 & $\geq$106 [102, 114] & budget & 27 (89) \\
\texttt{deepseek-v4-flash} & 52 & $\geq$72 & $\geq$85 [83, 87] & budget & 9 (62) \\
\texttt{deepseek-v4-pro-0813} & 36 & $\geq$96 & 121 [119, 140] & resolved & 17 (98) \\
\texttt{deepseek-v4-pro} & 36 & $\geq$72 & $\geq$76 [70, 86] & budget & 21 (44) \\
\texttt{gemma-4-26b-a4b-it} & 83 & $\geq$41 & $\geq$40 [39, 40] & budget & 23 (41) \\
\texttt{glm-4.6} & 41 & $\geq$53 & 45 [44, 45] & resolved & 1 (61) \\
\texttt{gpt-5.6-luna} & 46 & $\geq$45 & 37 [34, 52] & resolved & 8 (55) \\
\texttt{kimi-k2.6} & 76 & $\geq$56 & 93 [85, 105] & resolved & 2 (53) \\
\texttt{qwen3.5-35b-a3b} & 42 & $\geq$50 & 53 [28, 54] & resolved & 5 (54) \\
\texttt{qwen3.5-9b} & 35 & $\geq$25 & 26 [26, 26] & resolved & 3 (27) \\
\bottomrule
\end{tabular}
\end{table}

\begin{figure}[!htb]
\centering
\includegraphics[width=\textwidth]{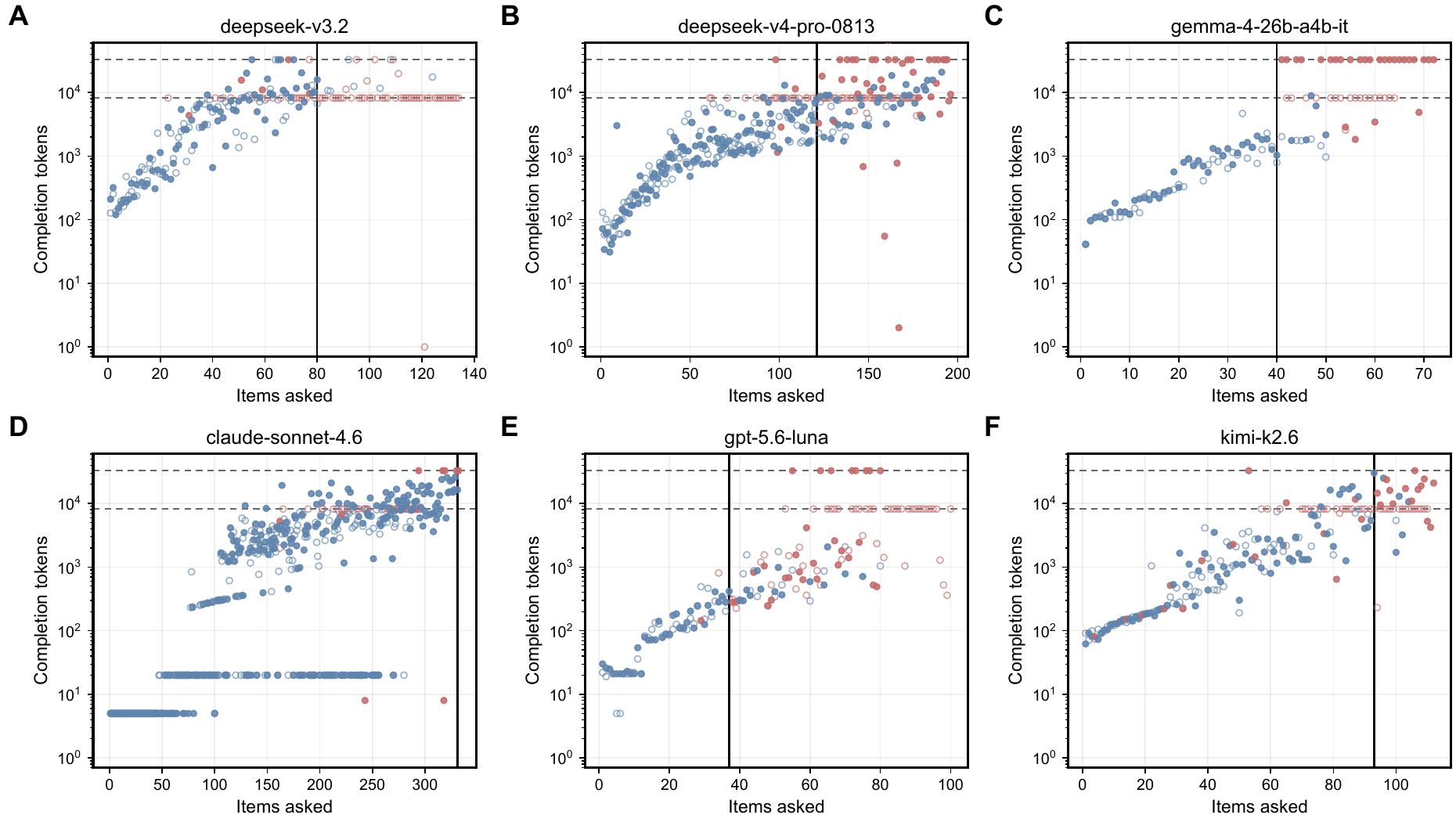}
\caption{\textbf{Completion tokens against items asked at two output caps.} Six reasoning configurations from the 32,768-token rescan (\textbf{A} to \textbf{F}; filled markers) overlaid on their catalogue scans at the 8,192-token cap (hollow markers), exact answers in blue and inexact in red. Dashed rules mark the two caps and the vertical rule the boundary fitted at the larger cap. Tokens grow roughly linearly with the count inside the stable regime and jump to the cap immediately past the boundary.}
\label{fig:reasoning_tokens}
\end{figure}

Figure~\ref{fig:reasoning_tokens} shows where the tokens go. In the stable regime, completion tokens grow roughly linearly with the count, at 9 to 230 tokens per item across six representative configurations, against the three or four tokens per item that writing the items out once would cost: the traces enumerate several times, in groups, with subtotals and checks. At the boundary the growth turns into a jump. The median spend over the ten probes before the fitted boundary against the ten after it rises from 1{,}400 to 21{,}000 tokens in one model and from 16{,}000 to the full 32{,}768 in another, and the probes that hit the cap sit immediately past the boundary rather than at the far end of the scan. Raising the cap fourfold moves the wall, not the boundary: the same configurations exhaust it at nearly the same counts.

The thirteen excluded pairs are the anomalies, collected in Table~\ref{tab:pair_anomalies}: in eight pairs the provider ignores the disable request outright, and five pairs spend no reasoning tokens in either configuration. A ninety-first two-configuration model, \texttt{minimax-m1}, is absent from both counts because neither of its scans resolved a boundary (Appendix~\ref{app:catalogue}).

{\scriptsize\noindent \texttt{claude-haiku-4.5}, \texttt{claude-opus-4}, \texttt{claude-opus-4.1}, \texttt{claude-opus-4.5}, \texttt{claude-opus-4.6}, \texttt{claude-opus-4.8}, \texttt{claude-opus-5}, \texttt{claude-sonnet-4}, \texttt{claude-sonnet-4.5}, \texttt{claude-sonnet-4.6}, \texttt{claude-sonnet-5}, \texttt{deepseek-chat-v3.1}, \texttt{deepseek-v3.1-terminus}, \texttt{deepseek-v3.2}, \texttt{deepseek-v3.2-exp}, \texttt{deepseek-v4-flash}, \texttt{deepseek-v4-flash-0731}, \texttt{deepseek-v4-flash-vision-exp}, \texttt{deepseek-v4-pro}, \texttt{deepseek-v4-pro-0813}, \texttt{gemini-2.5-flash}, \texttt{gemini-2.5-flash-lite}, \texttt{gemini-3-flash-preview}, \texttt{gemini-3.1-flash-lite}, \texttt{gemini-3.1-flash-lite-preview}, \texttt{gemma-4-26b-a4b-it}, \texttt{gemma-4-31b-it}, \texttt{mistral-medium-3-5}, \texttt{mistral-small-2603}, \texttt{kimi-k2.5}, \texttt{kimi-k2.6}, \texttt{kimi-k3}, \texttt{gpt-5.1}, \texttt{gpt-5.2}, \texttt{gpt-5.3-codex}, \texttt{gpt-5.4}, \texttt{gpt-5.4-mini}, \texttt{gpt-5.4-nano}, \texttt{gpt-5.5}, \texttt{gpt-5.6-luna}, \texttt{gpt-5.6-luna-pro}, \texttt{gpt-5.6-sol}, \texttt{gpt-5.6-sol-pro}, \texttt{gpt-5.6-terra}, \texttt{gpt-5.6-terra-pro}, \texttt{qwen3-235b-a22b}, \texttt{qwen3-8b}, \texttt{qwen3-max-thinking}, \texttt{qwen3.5-122b-a10b}, \texttt{qwen3.5-27b}, \texttt{qwen3.5-35b-a3b}, \texttt{qwen3.5-397b-a17b}, \texttt{qwen3.5-9b}, \texttt{qwen3.5-plus-02-15}, \texttt{qwen3.5-plus-20260420}, \texttt{qwen3.6-27b}, \texttt{qwen3.6-35b-a3b}, \texttt{qwen3.6-flash}, \texttt{qwen3.6-max-preview}, \texttt{qwen3.6-plus}, \texttt{qwen3.7-flash}, \texttt{qwen3.7-max}, \texttt{qwen3.7-plus}, \texttt{qwen3.8-27b}, \texttt{grok-4.20}, \texttt{grok-4.3}, \texttt{glm-4.5}, \texttt{glm-4.5-air}, \texttt{glm-4.5v}, \texttt{glm-4.6}, \texttt{glm-4.6v}, \texttt{glm-4.7}, \texttt{glm-4.7-flash}, \texttt{glm-5}, \texttt{glm-5-turbo}, \texttt{glm-5.2}, \texttt{glm-5v-turbo}.\par}

\begin{table}[htbp]
\centering
\scriptsize
\caption{Hybrid pairs excluded from the matched comparison. Token
columns give the mean completion tokens per probe over the lengths that
both configurations count reliably. \emph{Knob ignored} marks
providers that apply reasoning regardless of the request, so the pair
amounts to two measurements of one configuration; \emph{no reasoning}
marks pairs whose reasoning-enabled configuration spent no reasoning
tokens either.}
\label{tab:pair_anomalies}
\begin{tabular}{@{}l r r r r r l@{}}
\toprule
Model & CC$_{\text{off}}$ & CC$_{\text{on}}$ & $\Delta$CC & tok$_{\text{off}}$ & tok$_{\text{on}}$ & Class \\
\midrule
\texttt{deepseek-r1-distill-llama-70b} & 57 & 58 & +1 & 3,508 & 3,472 & knob ignored \\
\texttt{minimax-m3} & 51 & 56 & +5 & 56 & 173 & knob ignored \\
\texttt{gpt-5.1-codex-mini} & 50 & 50 & +0 & 1,623 & 1,182 & knob ignored \\
\texttt{qwen-plus-2025-07-28:thinking} & 58 & 58 & +0 & 6,007 & 5,715 & knob ignored \\
\texttt{qwen3-14b} & 27 & 31 & +4 & 480 & 737 & knob ignored \\
\texttt{qwen3-30b-a3b} & 37 & 35 & -2 & 349 & 794 & knob ignored \\
\texttt{qwen3-32b} & 22 & 26 & +4 & 298 & 316 & knob ignored \\
\texttt{glm-5.1} & 50 & 40 & -10 & 32 & 447 & knob ignored \\
\texttt{claude-opus-4.7} & 158 & 156 & -2 & 6 & 6 & no reasoning \\
\texttt{qwen-plus-2025-07-28} & 54 & 54 & +0 & 2 & 2 & no reasoning \\
\texttt{qwen3-coder-plus} & 50 & 50 & +0 & 2 & 2 & no reasoning \\
\texttt{qwen3-max} & 60 & 60 & +0 & 2 & 2 & no reasoning \\
\texttt{qwen3.5-flash-02-23} & 42 & 46 & +4 & 2 & 2,015 & no reasoning \\
\bottomrule
\end{tabular}
\end{table}

\FloatBarrier

\section{The execution ladder}\label{app:ladder}

The ladder extends the counting measurement toward the grid execution that abstract-reasoning tasks demand, in three levels of increasing structure: counting within a row, executing a rule on a row, and executing a rule on a grid. Every level keeps the measurement's full contract: the model returns a single integer, scored by exact match, and the stimulus is constructed so that the correct answer equals the scanned load; the two-regime estimator of Appendix~\ref{app:estimation} therefore fits each capacity in the same unit as CC, with the same stopping rule.

\subsection{Tasks and probe contract}

The stimuli are rows or grids of space-separated digits, the serialization abstract-reasoning grids use in text. The row-counting task probes the structure of a row as it stands: the row holds exactly $N$ cells of a target digit, scattered in short runs among distractor runs whose number and lengths vary widely, and the model reports the cell count. The construction decorrelates row length from the answer, so gross input size carries no information. The row-rule tasks state a transformation rule and ask for the target-digit count after the rule is applied to the row: every run of the target digit grows by one cell at each end; every cell of a second digit is recolored to the target; or every single-cell run is deleted. The transformed row never appears in the context, so the probed quantity exists only if the rule was executed internally. The grid-rule tasks apply the recoloring and deletion rules to two-dimensional grids of rectangular regions on a uniform background, where connectivity is spatial rather than linear. The ladder is six tasks: one row-counting task, three row-rule tasks, and two grid-rule tasks.

Each probe must satisfy two requirements. It must be \emph{global}: the probed quantity is a function of the whole transformed state, so no partial evaluation answers it (a single-cell lookup would suffice for a mirror rule, which is why permutation rules are excluded). It must be \emph{non-invariant}: the quantity must actually change under the rule and not be derivable from a cheap invariant of the input (a color count under a color-preserving rule would be). Non-invariance excludes cheap invariants, not closed forms: the recolor answer equals the sum of two input color counts, and the grow and delete answers are sums over runs. The ladder therefore measures task modeling under a stated rule, which is its target, and does not establish that the model constructs the transformed row or grid. Every generator chooses its run or region structure so the post-rule answer equals the scanned load exactly, and every stimulus is verified against an independent simulator of the rule before it is sent.

\subsection{One model, all six tasks: capacity counts objects}

We first measured the full ladder on \texttt{gemini-3.7-flash} (Table~\ref{tab:ladder_pilot}). The capacities span 238 to 340 in answer units against a CC of 352. Converting each boundary into the number of objects the model had to hold (runs in a row, connected regions in a grid, measured from the generators at the boundary load) leaves 132 objects for the pure-counting probe and 70 to 103 under a rule. The two rules that appear in both dimensions allow a direct comparison. Moving the recoloring rule from rows to grids raises the answer-unit capacity from 253 to 340 while the object count falls from 103 to 85; moving the deletion rule raises it from 238 to 334 while the object count falls from 82 to 72. The answer-unit capacity rises by a third to two fifths between rows and grids while the object count does not rise at all. For this model and these generators the boundaries correspond to 70 to 132 objects, consistent with object-based aggregation; the experiment does not separately estimate the costs of object number, object size, and rule execution.

\begin{table}[htbp]
\centering
\scriptsize
\caption{\textbf{Full ladder on \texttt{gemini-3.7-flash}.} Fitted capacity with its calibrated interval, and the mean number of objects held at the boundary load. The model's CC is 352 [350, 373].}
\label{tab:ladder_pilot}
\begin{tabular}{llrlr}
\toprule
Level & Task & Capacity & CI & Objects \\
\midrule
row count & scattered cells & 326 & [326, 326] & 132 \\
row rule & grow (row) & 282 & [274, 283] & 70 \\
row rule & recolor (row) & 253 & [247, 255] & 103 \\
row rule & delete (row) & 238 & [225, 287] & 82 \\
grid rule & recolor (grid) & 340 & [338, 371] & 85 \\
grid rule & delete (grid) & 334 & [332, 338] & 72 \\
\bottomrule
\end{tabular}
\end{table}

\subsection{The cohort program and the ratio law}

We then measured all six ladder scans on every model in the ARC-AGI-2 post-release cohort plus \texttt{claude-sonnet-5}, 180 raw measurements over 30 models. Scans are bracketed from the pilot: each boundary is predicted as a fixed multiple of the model's CC, and the scan starts at 0.35 times the prediction with a step near one eightieth of it, stopping by the criterion of Appendix~\ref{app:estimation}.

\paragraph{Protocol exclusion: \texttt{claude-fable-5}.}
The serving layer returned \texttt{content\_filter} for 103 of the model's 244 probes, distributed across all six ladder tasks and including a minimal load-8 preflight. These events return no answer, so execution failure cannot be distinguished from provider-side suppression. We retain all raw records as a case of protocol incompatibility but exclude all six measurements, including the scans that otherwise fit a boundary, from the cohort analysis. One filtered preflight is reproduced below (the digit row is line-wrapped for display):
\begin{quote}
\scriptsize\ttfamily\raggedright
System: You follow instructions exactly. Reply with one integer and nothing else.\\[2pt]
User: Apply this rule to the row below: every cell containing 3 becomes 7. After applying the rule, how many cells contain the digit 7?\\[2pt]
1 1 1 7 2 2 6 6 2 2 0 0 6 6 0 0 0 0 0 6 6 6 6 6\\
1 1 1 7 8 8 5 5 5 5 5 9 9 9 7 7 7 7 6 6 5 5 5 5\\
5 7 7 5 5 5 5 5 1 1 1 1 1\\[2pt]
Provider: status=completed; detail=content\_filter; response=[empty].
\end{quote}

The resulting analysis cohort contains 174 measurements over 29 models. Of these, 90 resolve inside the scan bracket, and their fitted boundaries track the CC-scaled predictions closely (\(\rho=0.83\) between log prediction and log boundary, median ratio 0.95). This establishes a within-bracket relation; the out-of-range measurements are censored evidence and are treated as such below.

To test how strongly that result depended on the pilot-derived bracket, we extended the raw scans sequentially rather than substituting arbitrary point values for their unresolved outcomes. First, all 46 below-start and 8 mixed scans were continued downward to the smallest admissible load of their task (1 for row counting and 4 for the rule tasks), for 54 lower extensions. After refitting those combined records, the 36 scans still classified as ceiling-stable or mixed were continued upward from their last tested load to a safety ceiling of four times the CC-scaled prediction, stopping earlier whenever the same transition criterion resolved. Fixed-grid probes were issued in batches of three and the criterion was evaluated in load order after each batch, so parallel execution can overshoot the first sequential stopping point by at most two load increments. Every lower and upper probe is retained in the consolidated fit.

The consolidated analysis records what each scan identifies rather than retaining only point estimates. A resolved scan contributes its fitted boundary; a scan already below criterion at the task minimum is left-censored with the minimum as an upper bound; and a scan still stable at the expanded ceiling is right-censored with that ceiling as a lower bound. Nonmonotone mixed scans do not support the ordering assumed by a single transition and are excluded from the primary likelihood, then included using their tested ranges only as a separate interval sensitivity. Blank-dominated scans provide no numerical bound on the boundary. The primary model is a Gaussian censored regression on the log scale, \(\log B_i=\alpha+\beta\log P_i+\epsilon_i\), where \(B_i\) is the latent execution boundary, \(P_i\) the CC-scaled prediction, and \(\epsilon_i\sim\mathcal{N}(0,\sigma^2)\). A nonparametric bootstrap resamples models with all six task scans as a cluster. We also report the resolved-only Spearman correlation, a transparent limit-substitution correlation, and the distribution of Spearman correlations obtained by drawing censored boundaries from the fitted truncated distributions; the latter is a conditional imputation range, not a sampling confidence interval.

Extending the scans resolves 139 of the 174 measurements, against 90 inside the original bracket (Fig.~\ref{fig:landscape}G). On the expanded records the censored regression gives a slope of 1.00 (95\% cluster bootstrap [0.46, 1.51]; likelihood ratio 40.5 against a zero slope, \(p=2\times10^{-10}\)), so the execution boundary scales in proportion to the CC-scaled prediction, with a residual standard deviation of 1.53 on the log scale. The 139 resolved boundaries sit at a median 0.91 of their prediction and correlate with it at \(\rho=0.56\); substituting each observed censoring limit, or drawing the censored boundaries from their fitted truncated distributions, leaves the correlation between 0.55 and 0.56. Table~\ref{tab:ladder_model_brackets} gives the per-model outcome counts before and after the extension.

The ladder also ranks the models consistently from task to task. On the resolved boundaries the Spearman correlation between any two tasks averages 0.77, and Kendall's coefficient of concordance across the six tasks is \(W=0.83\) on the 17 models resolved on every task; entering each censored scan at its observed bound raises these to 0.83 and \(W=0.86\) on 28 models. Each single task, by contrast, correlates with CC at only 0.4 to 0.7 with bounds substituted (0.3 to 0.7 on resolved boundaries alone). The models therefore agree across the six tasks far more than any task agrees with CC: the family departures of Fig.~\ref{fig:landscape}G are departures of a coherent execution capacity from the counting capacity, not task-specific noise.

\begingroup
\small
\setlength{\LTcapwidth}{\textwidth}
\begin{longtable}{@{}lccr@{}}
\caption{\textbf{Model-level bracket sensitivity.} Counts among the six ladder tasks before and after extending scans down to the task minimum and up to $4\times$ the CC-scaled prediction. Each tuple is resolved/left-censored/right-censored/mixed/blank (R/L/U/M/B); the last column is the median measured/predicted ratio among resolved expanded scans.}\label{tab:ladder_model_brackets}\\
\toprule
Model & Original R/L/U/M/B & Expanded R/L/U/M/B & Median ratio \\
\midrule
\endfirsthead
\toprule
Model & Original R/L/U/M/B & Expanded R/L/U/M/B & Median ratio \\
\midrule
\endhead
\bottomrule
\endfoot
\texttt{claude-opus-4.6} & 4/1/0/0/1 & 5/0/0/0/1 & 0.5751 \\
\texttt{claude-opus-4.7} & 2/4/0/0/0 & 4/2/0/0/0 & 0.4362 \\
\texttt{claude-opus-4.8} & 5/1/0/0/0 & 5/1/0/0/0 & 0.6579 \\
\texttt{claude-opus-5} & 4/0/1/1/0 & 6/0/0/0/0 & 1.9263 \\
\texttt{claude-sonnet-4.6} & 2/4/0/0/0 & 6/0/0/0/0 & 0.3133 \\
\texttt{claude-sonnet-5} & 4/2/0/0/0 & 6/0/0/0/0 & 0.5336 \\
\texttt{deepseek-v3.2} & 5/0/1/0/0 & 6/0/0/0/0 & 1.3992 \\
\texttt{deepseek-v4-flash-0731} & 2/0/3/1/0 & 6/0/0/0/0 & 1.7346 \\
\texttt{deepseek-v4-pro-0813} & 6/0/0/0/0 & 6/0/0/0/0 & 0.9753 \\
\texttt{gemini-3.1-pro-preview} & 6/0/0/0/0 & 6/0/0/0/0 & 0.6691 \\
\texttt{gemini-3.5-flash} & 6/0/0/0/0 & 6/0/0/0/0 & 0.3855 \\
\texttt{gemini-3.5-flash-lite} & 6/0/0/0/0 & 6/0/0/0/0 & 0.7739 \\
\texttt{gemini-3.6-flash} & 6/0/0/0/0 & 6/0/0/0/0 & 0.5054 \\
\texttt{gemini-3.7-flash} & 6/0/0/0/0 & 6/0/0/0/0 & 0.997 \\
\texttt{glm-5} & 5/0/1/0/0 & 6/0/0/0/0 & 1.4253 \\
\texttt{glm-5.2} & 2/0/4/0/0 & 6/0/0/0/0 & 2.0963 \\
\texttt{gpt-5.2} & 3/0/2/1/0 & 6/0/0/0/0 & 1.9172 \\
\texttt{gpt-5.4} & 0/6/0/0/0 & 4/2/0/0/0 & 0.0628 \\
\texttt{gpt-5.4-mini} & 5/0/1/0/0 & 6/0/0/0/0 & 1.7521 \\
\texttt{gpt-5.4-nano} & 0/6/0/0/0 & 2/4/0/0/0 & 0.2117 \\
\texttt{gpt-5.5} & 1/5/0/0/0 & 6/0/0/0/0 & 0.1146 \\
\texttt{gpt-5.6-luna} & 0/6/0/0/0 & 1/5/0/0/0 & 0.187 \\
\texttt{gpt-5.6-sol} & 0/0/6/0/0 & 4/0/2/0/0 & 3.0545 \\
\texttt{gpt-5.6-terra} & 0/6/0/0/0 & 3/3/0/0/0 & 0.0735 \\
\texttt{grok-4.5} & 2/0/4/0/0 & 3/0/3/0/0 & 0.9643 \\
\texttt{grok-4.6} & 1/0/5/0/0 & 1/0/5/0/0 & 1.9358 \\
\texttt{kimi-k2.5} & 1/0/1/4/0 & 2/0/2/2/0 & 1.1815 \\
\texttt{kimi-k3} & 0/5/0/1/0 & 3/2/0/1/0 & 0.2673 \\
\texttt{minimax-m2.5} & 6/0/0/0/0 & 6/0/0/0/0 & 1.555 \\
\end{longtable}

\endgroup

The residual departures are systematic at the level of model families, and each class is informative. Nineteen scans still fail below the smallest admissible load, concentrated in four OpenAI lines with one or two scans each from three other models; these return well-formed but wrong integers at loads as small as four, placing their capacities below anything the generators can construct. Twelve scans are still stable at four times the prediction, concentrated in two xAI models and one each from Moonshot and OpenAI, so that within one vendor line one release exceeds four times its predicted boundaries while a sibling falls below the task minimum. Three scans, all on one line, have no declarable transition over the range tested. One is dominated by empty replies after exhausting the fixed output budget. Under the fixed-budget protocol, budget exhaustion is an execution failure rather than missing data; its cause is retained separately from the scan's transition classification.

\paragraph{Format, not capacity, at the lower departures.}
The scans that fall below the task minimum do not read as capacity measurements. Across their probes at loads of twenty or fewer they never return an empty reply and almost never a malformed one, against a few percent of each elsewhere in the cohort: these models follow the output contract more closely than the cohort average, not less. Their answers also sit close to the truth, 73\% within three of it. A model wrong by one or two at four objects may be misreading what the stimulus asks rather than failing to hold a four-object total; we treat this as a hypothesis, since closeness of the answer does not establish it and we have not tested it by changing the serialization or adding examples while holding the computation fixed. If it holds, it would be the factor separating the two ARC-AGI-2 cohorts of Fig.~\ref{fig:landscape}F: familiarity with a serialization that does not occur in natural text, binding before capacity does.

The nineteen scans split in two by the model's ARC-AGI-2 score relative to its capacity, and the split is the informative part. The four OpenAI lines sit below the ladder floor and also below the ARC-AGI-2 score their capacity predicts: on both tasks the same thing limits them. The other three models sit above it: Claude Opus 4.7 and 4.8 and \texttt{kimi-k3} score higher on ARC-AGI-2 than their counting capacity predicts, yet fail a ladder task at four objects. They execute the benchmark's own serialization well and a near neighbor of it not at all, which is what adaptation to a fixed format, rather than a general gain, would produce.

Two vendor lines make this a trajectory rather than a cross-sectional accident. Across three successive Claude Opus releases the measured capacity falls from 622 to 133 while the ARC-AGI-2 score rises, and the residual moves from below the capacity-predicted score to well above it; the first release has no floor failure and the two later ones do. From \texttt{kimi-k2.5} to \texttt{kimi-k3} the capacity falls from 73 to 54 while the score rises more than fourfold, and again only the successor fails at the floor. Serving prices are identical across the Opus releases and higher for the later Kimi model, so this is not a substitution of a smaller, cheaper served model.

Both cells of the dissociation are occupied. Three models are strong on ARC-AGI-2 for their capacity and fail the ladder floor; five others, from four vendors, are weak on ARC-AGI-2 for their capacity and clear it. Familiarity with a format and the capacity to model a task under it may therefore be separable abilities rather than two views of one, a hypothesis the untested format intervention would settle, and it would explain why the pooled association between them is null across the cohort: the pooled estimate averages two populations that depart from the scaling for opposite reasons.

\subsection{Interference protocol and results}

The interference experiment of Fig.~\ref{fig:resources}C scans counting while the model carries one ladder task at a fixed load. The prompt presents the load first and the counting sequence second, and the model replies with two integers, the load's answer and then the count, so the counting boundary is fitted by the unchanged parser while the load answer is graded separately from the raw reply. Loads are set per model from its own measured solo capacities: for \texttt{gemini-3.7-flash}, grids of 26, 51, 76, and 102 regions (30--120\% of its 85-object grid capacity) and row loads at 60\% of the row-counting and row-rule capacities; for \texttt{claude-sonnet-5}, grids of 6 to 25 regions and row loads of 8 to 9 runs on the same fractions of its own capacities. The design has two controls: a control grid of matched token volume whose first integer is printed in the prompt, and a pointer chain of matched size whose running value is fully expressible in language. \texttt{gpt-5.6-terra} was measured and excluded: its calibration scans fell below their starts, so its derived loads were one to five regions, effectively no load, and the run is archived rather than interpreted.

\begin{table}[htbp]
\centering
\scriptsize
\caption{\textbf{Counting under calibrated concurrent loads, single scans.} Fitted counting boundary under each load, its ratio to the solo CC, and the accuracy of the load task's own answer over the scan. One scan per cell; Table~\ref{tab:interference_replicates} repeats five of the conditions five times each.}
\label{tab:interference}
\begin{tabular}{llrrr}
\toprule
Model (solo CC) & Load & CC under load & Ratio & Load accuracy \\
\midrule
\multirow{8}{*}{\shortstack[l]{\texttt{gemini-3.7-flash}\\(352)}}
 & control & 342 & 0.97 & 1.00 \\
 & pointer & 370 & 1.05 & 1.00 \\
 & row 60\% & 290 & 0.82 & 0.95 \\
 & rule 60\% & 226 & 0.64 & 0.91 \\
 & grid 30\% & 322 & 0.91 & 0.97 \\
 & grid 60\% & 278 & 0.79 & 0.92 \\
 & grid 90\% & 270 & 0.77 & 0.95 \\
 & grid 120\% & 266 & 0.76 & 0.96 \\
\midrule
\multirow{8}{*}{\shortstack[l]{\texttt{claude-sonnet-5}\\(89)}}
 & control & 91 & 1.02 & 1.00 \\
 & pointer & 63 & 0.71 & 1.00 \\
 & row 60\% & 89 & 1.00 & 0.96 \\
 & rule 60\% & 64 & 0.72 & 0.92 \\
 & grid 30\% & 85 & 0.96 & 0.79 \\
 & grid 60\% & 89 & 1.00 & 0.68 \\
 & grid 90\% & 87 & 0.98 & 0.66 \\
 & grid 120\% & 89 & 1.00 & 0.68 \\
\bottomrule
\end{tabular}
\end{table}

Table~\ref{tab:interference} gives single scans, one per cell. In them the matched-length control moves neither model, \texttt{gemini-3.7-flash} loses capacity under every grid and row load, most under the rule load, and \texttt{claude-sonnet-5} keeps its count under grid loads while its grid answers drop to 66--79\%, and loses 28\% under the rule load and 29\% under the pointer chain. One scan per cell cannot separate an effect of the load from the scan-to-scan spread of Appendix~\ref{app:replicates}, so we repeated the experiment.

\paragraph{Replication.} We scanned six conditions five times each in both models (60 scans): counting alone, the matched-length control, the pointer chain, the rule load at 60\%, the grid load at 60\%, and a second control that the model has to read. In that control the load is a passage of encyclopedic sentences as long as the 90\% grid, followed by a four-option question on one fact stated in the passage; the model reports the option number and then the count. Each scan runs from 0.2 of the catalogue CC until the transition is resolved, in steps of one item for \texttt{claude-sonnet-5} and four for \texttt{gemini-3.7-flash}. Ratios are taken to the mean of the model's own five solo scans (Table~\ref{tab:interference_replicates}; Fig.~\ref{fig:resources}C).

\begin{table}[htbp]
\centering
\scriptsize
\setlength{\tabcolsep}{4pt}
\caption{\textbf{Counting under concurrent loads, five scans per cell.} The fitted counting boundary of each scan, the mean and standard deviation, the ratio of the mean to the model's mean solo boundary, and the accuracy of the load task's own answer (mean and standard deviation over the five scans).}
\label{tab:interference_replicates}
\begin{tabular}{llllrr}
\toprule
Model & Load & Boundaries & Mean $\pm$ SD & Ratio & Load accuracy \\
\midrule
\multirow{6}{*}{\texttt{gemini-3.7-flash}}
 & none (solo) & 290 370 346 386 394 & $357 \pm 42$ & 1.00 & -- \\
 & control (matched length) & 302 386 282 354 298 & $324 \pm 44$ & 0.91 & $1.00 \pm 0.00$ \\
 & control (reading) & 386 382 386 390 290 & $367 \pm 43$ & 1.03 & $1.00 \pm 0.00$ \\
 & pointer chain & 382 382 286 222 390 & $332 \pm 75$ & 0.93 & $1.00 \pm 0.00$ \\
 & rule 60\% & 386 402 398 386 362 & $387 \pm 16$ & 1.08 & $0.89 \pm 0.04$ \\
 & grid 60\% & 350 250 262 298 294 & $291 \pm 39$ & 0.81 & $0.92 \pm 0.04$ \\
\midrule
\multirow{6}{*}{\texttt{claude-sonnet-5}}
 & none (solo) & 92 100 100 92 91 & $95.0 \pm 4.6$ & 1.00 & -- \\
 & control (matched length) & 85 85 100 86 85 & $88.2 \pm 6.6$ & 0.93 & $1.00 \pm 0.01$ \\
 & control (reading) & 63 89 92 91 92 & $85.4 \pm 12.6$ & 0.90 & $1.00 \pm 0.00$ \\
 & pointer chain & 64 63 64 64 62 & $63.4 \pm 0.9$ & 0.67 & $1.00 \pm 0.00$ \\
 & rule 60\% & 89 89 80 88 89 & $87.0 \pm 3.9$ & 0.92 & $0.91 \pm 0.02$ \\
 & grid 60\% & 89 85 86 81 80 & $84.2 \pm 3.7$ & 0.89 & $0.65 \pm 0.05$ \\
\bottomrule
\end{tabular}
\end{table}

The two controls leave the mean boundary between 0.90 and 1.03 of solo, and both models answer them correctly throughout. The controls are not free in \texttt{claude-sonnet-5}, where they cost 7\% and 10\%, but neither differs from solo beyond the spread of the scans (Welch \(p=0.10\) and 0.17).

One load in each model lowers the boundary further. In \texttt{claude-sonnet-5} the pointer chain gives 62 to 64 in all five scans, 0.67 of solo (\(p<0.001\) against solo and \(p=0.001\) against the matched-length control), while the chain itself is followed correctly every time. In \texttt{gemini-3.7-flash} the grid load gives 0.81 of solo (\(p=0.03\) against solo); this model's scans spread widely, and the grid load is not separable from the matched-length control at 0.91 (\(p=0.24\)).

The rule load does not replicate. Its single scans gave 0.64 in \texttt{gemini-3.7-flash} and 0.72 in \texttt{claude-sonnet-5}; five scans give 1.08 and 0.92, the second no lower than that model's controls. The single-scan values of 226 and 64 fall where an occasional scan of each model collapses early (222 under the pointer chain for the first; 63 under the reading control and 64 in a repeat of the baseline encoding for the second, Appendix~\ref{app:replicates}), so they are better read as scan-to-scan spread than as an effect of the rule.

The grid load in \texttt{claude-sonnet-5} leaves counting where the controls leave it (0.89) while the grid answer is right in 65\% of probes, at a load set to 60\% of the model's solo capacity for that task; the single scans showed the same pattern. The model's replies to this condition carry reasoning tokens (a median near 800 per probe, against almost none for the controls), so it works on the grid and still gets it wrong a third of the time.

The replicated evidence that a second task draws on counting capacity is one load in each model, the pointer chain in \texttt{claude-sonnet-5} and, less securely, the grid in \texttt{gemini-3.7-flash}. That evidence is consistent with a shared resource and does not distinguish one from other forms of dual-task interference. The pointer result also refines the anchored-execution control of Appendix~\ref{app:anchored}: a task whose running value can be written down can be executed far beyond CC, but executing it at the same time as counting is not free.

\paragraph{Open-weight models.} We ran the same six conditions on three open-weight models locally with greedy decoding (\texttt{gpt-oss-20b}, \texttt{gemma-4-26b-a4b-it}, \texttt{qwen3.6-35b-a3b}), with five random draws of each load. The run does not test the question. The loads were sized from each model's CC and not from a measured capacity for the load task, and answering immediately the three models solve the rule and grid loads in 0 to 7\% of probes, so those conditions do not make the model execute anything. The controls, which all three answer correctly, already move the boundary: to 0.78 of solo under the reading control in \texttt{gemma-4-26b-a4b-it}, and to 0.56 and 0.52 under the two controls in \texttt{qwen3.6-35b-a3b}, about as much as under the loads or more. In these two models text placed before the sequence can shorten the count whether or not it carries a task. Only \texttt{gpt-oss-20b} keeps its boundary under both controls (1.32 and 1.08 of solo). The records are released with the others.

\FloatBarrier

\section{Cross-model mechanistic probe suite}\label{app:mech}

Six open-weight instruction models were probed end-to-end on the counting task with one battery: capacity scans in immediate-answer and native reasoning modes, cross-validated linear readouts of the residual stream at four depths, causal interventions at the decode point, across item positions, and at the boundary between a prefix and a continuation, and a count-versus-length deconfound. The set spans five vendors, dense and mixture-of-experts architectures, and 20B to 120B parameters: \texttt{gpt-oss-20b} (24 layers), \texttt{gpt-oss-120b} (36), \texttt{gemma-4-26b-a4b-it} (30), \texttt{muse-glimmer-30b} (52, dense), \texttt{qwen3.6-35b-a3b} (40), and \texttt{llama-3.3-70b-instruct} (80, dense). Four residual sites per model are hooked at about one quarter, one half, three quarters, and the final layer. All runs use greedy decoding and the paper's prompt through each checkpoint's own chat template, on one 192-GB workstation; \texttt{gpt-oss-120b} runs quantized through an MLX backend. The two-regime estimator applied to these local scans reproduces each model's catalogued CC to within a few items, so the probe results are read against the paper's capacities. Immediate-answer mode, in which the model emits the integer with no intervening text, is the probe protocol: it is the regime in which the decoder reads the state that the probes measure and edit.

Each subsection supports one claim of Section~\ref{sec:probes} and one part of Fig.~\ref{fig:mechanistic_probe}: the count is readable at every depth and most exactly early (panels B to D); the decode point does not read the linear coordinate (the control in E); the output channel is the decoder's own answer-token subspace and is not counting-specific (F); the accumulation is read at mid-depth across item positions (E); the boundary state carries the count forward and carries the state of an object history (G); the count is readable but not routed on mixed sequences (H); and reasoning modes change capacity in both directions (I).

\subsection{The count is readable at every depth, most exactly early}\label{app:readout}

Readouts are ridge regressions from the decode-point residual state to the true count, fitted on the stable regime with folds that hold out whole counts. Cross-validation is essential: hidden sizes exceed the number of stable counts, so an unregularized fit interpolates and in-sample \(R^2\) is vacuous. Held-out \(R^2\) is 0.94 to 0.999 at every site of every model, with root-mean-square errors of 0.2 to 3.2 items that grow two- to six-fold from the earliest to the last hooked layer, and residuals that are strongly autocorrelated across adjacent counts (\(\rho\) 0.94 to 0.99), as an analog accumulator would produce. Linear noise exceeds half an item at most sites even in mid-regime, so a linear readout alone would not support exact counting at the observed capacities.

\begin{figure}[!htbp]
\centering
\includegraphics[width=\textwidth]{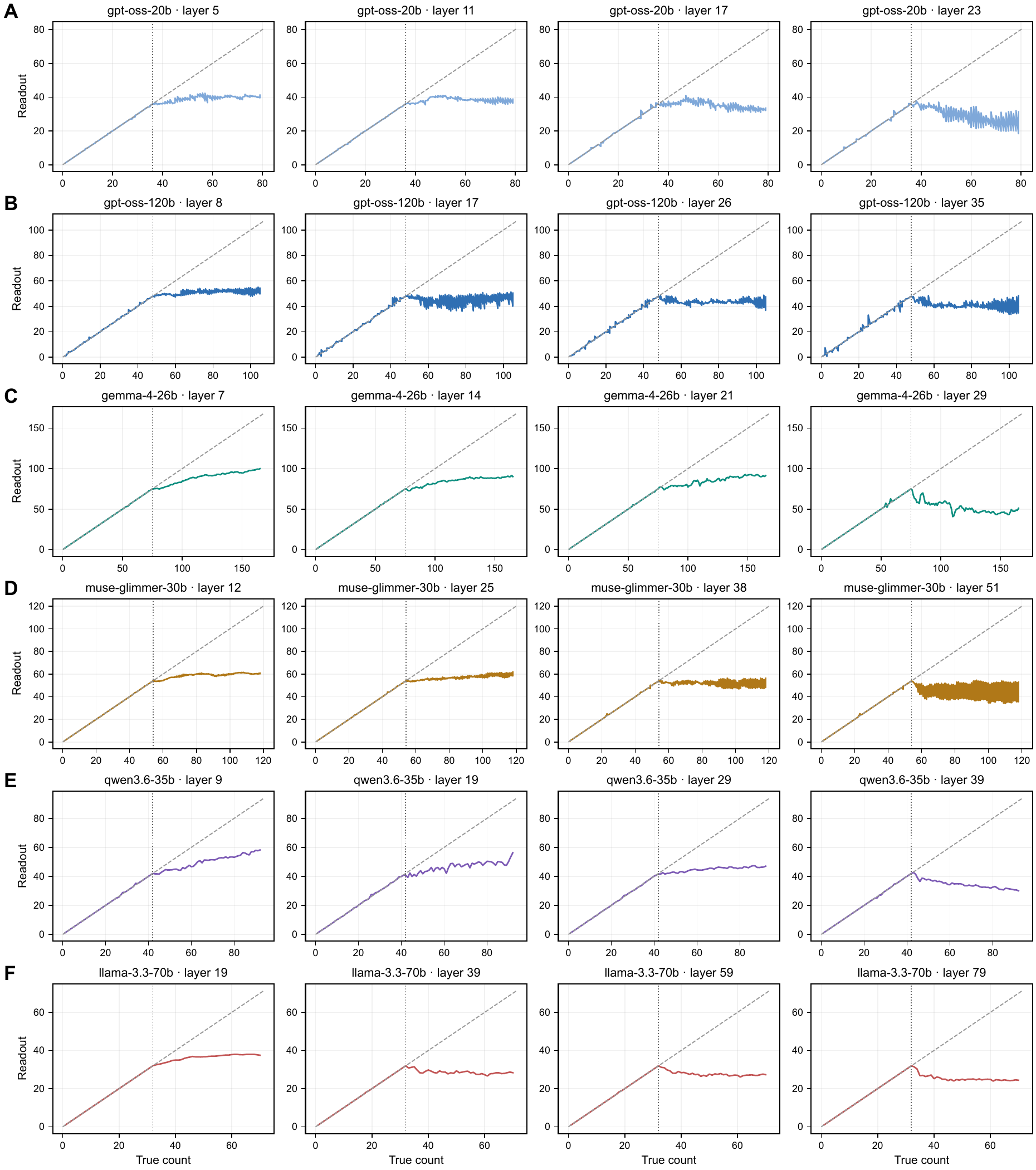}
\caption{\textbf{Linear count readouts at every probed site.} Held-out readout against true count for the six models (rows, \textbf{A} to \textbf{F}) at the four hooked depths (columns), extrapolated beyond the stable regime. The dashed line is the identity; the dotted vertical rule is the model's immediate-answer CC. Early layers continue to track beyond the boundary; the last layer stops advancing and turns against the count.}
\label{fig:suite_atlas}
\end{figure}

\begin{figure}[!htbp]
\centering
\includegraphics[width=\textwidth]{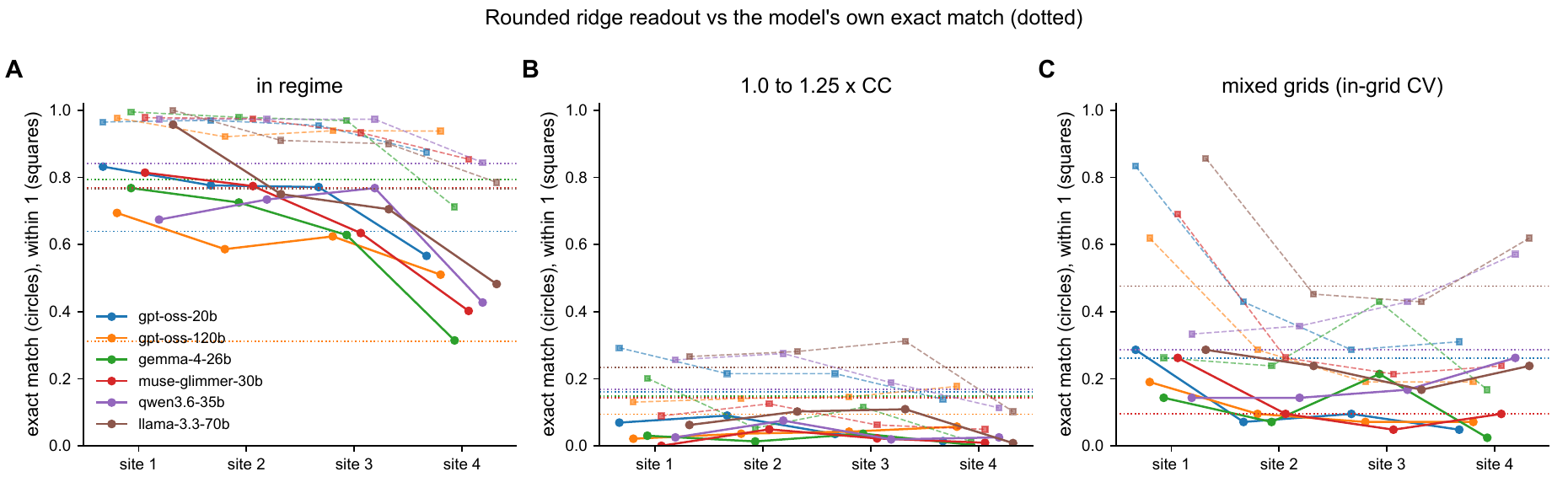}
\caption{\textbf{Exact match of the rounded readout by site.} Exact-match rate of the rounded readout against depth for each model, with the model's own rate as a horizontal line, in the stable regime (\textbf{A}), from 1.0 to 1.25 times CC (\textbf{B}), and on the mixed grids (\textbf{C}).}
\label{fig:round2_exact}
\end{figure}

Scoring the rounded readout by exact match on the same prompts as the behavior (Table~\ref{tab:round2_exact}, Fig.~\ref{fig:round2_exact}) gives the exactness the main text reports. In the stable regime the rounded readout at its best site is within 0.07 of the model's own exactness in three models and 0.19 to 0.38 above it in the other three (0.69 to 0.96 against 0.31 to 0.84), and it is highest at the early sites, falling by 0.2 to 0.5 from the first hooked site to the last in every model. Figure~\ref{fig:suite_atlas} extrapolates the readouts past each model's boundary. In the stable regime the readout rises by one per item by construction. Between 1.0 and 1.2 times CC the readout at the earliest hooked layer still rises, at 0.32 to 0.51 per item in five models and 0.06 in \texttt{gpt-oss-20b}, whose second hooked layer rises at 0.29. At the last hooked layer the slope is negative in all six models (\(-0.33\) to \(-1.69\)). At the two middle layers it is positive but below 0.4 in three models and negative at one or both layers in the other three. Beyond CC no site reads the integer exactly (at most 0.11), and the readout is less exact there than the model's own residual answers, so what persists past the boundary in the early layers is a weakened signal of magnitude and no longer an exact count. The teacher-forced margin of the correct answer first turns negative at the length of the first behavioral error in all six models.

\begin{table}[H]
\centering
\footnotesize
\setlength{\tabcolsep}{4pt}
\caption{\textbf{Exact match of the rounded readout against the model.} Exact-match and within-one-item rates of the rounded ridge readout at its best site (count-held-out folds) and of the model's own answer on the same prompts, in the stable regime, from 1.0 to 1.25 times CC, and on the mixed grids of Section~\ref{app:mixed} (readout fitted within the grid). Two conventions appear in the model columns and are kept distinct: the model's exact rate is the teacher-forced first-token match on the suite's multi-prompt cache, the same convention as the readout columns, which is why it can fall below one half (\texttt{gpt-oss-120b}) and is not the scan's full-answer rate that defines the stable regime; the model's within-one rate is the full-answer rate from the scan records.}
\label{tab:round2_exact}
\begin{tabular}{lcccccc}
\toprule
 & \multicolumn{2}{c}{Stable regime} & \multicolumn{2}{c}{1.0 to 1.25 CC} & \multicolumn{2}{c}{Mixed grids} \\
Model & readout & model & readout & model & readout & model \\
\midrule
\texttt{gpt-oss-20b} & 0.83 / 0.97 & 0.64 / 0.94 & 0.09 / 0.29 & 0.16 / 0.56 & 0.29 / 0.83 & 0.26 / 0.41 \\
\texttt{gpt-oss-120b} & 0.69 / 0.98 & 0.31 / 0.83 & 0.06 / 0.18 & 0.09 / 0.33 & 0.19 / 0.62 & 0.10 / 0.31 \\
\texttt{gemma-4-26b} & 0.77 / 1.00 & 0.79 / 1.00 & 0.04 / 0.20 & 0.15 / 0.26 & 0.21 / 0.43 & 0.29 / 0.50 \\
\texttt{muse-glimmer-30b} & 0.81 / 0.98 & 0.77 / 1.00 & 0.05 / 0.13 & 0.14 / 0.43 & 0.26 / 0.69 & 0.10 / 0.14 \\
\texttt{qwen3.6-35b} & 0.77 / 0.97 & 0.84 / 1.00 & 0.08 / 0.28 & 0.17 / 0.80 & 0.26 / 0.57 & 0.29 / 0.62 \\
\texttt{llama-3.3-70b} & 0.96 / 1.00 & 0.77 / 1.00 & 0.11 / 0.31 & 0.23 / 0.88 & 0.29 / 0.86 & 0.48 / 0.74 \\
\bottomrule
\end{tabular}
\end{table}

\subsection{The decode point does not read the linear coordinate}\label{app:decode}

\begin{figure}[!htbp]
\centering
\includegraphics[width=0.94\textwidth]{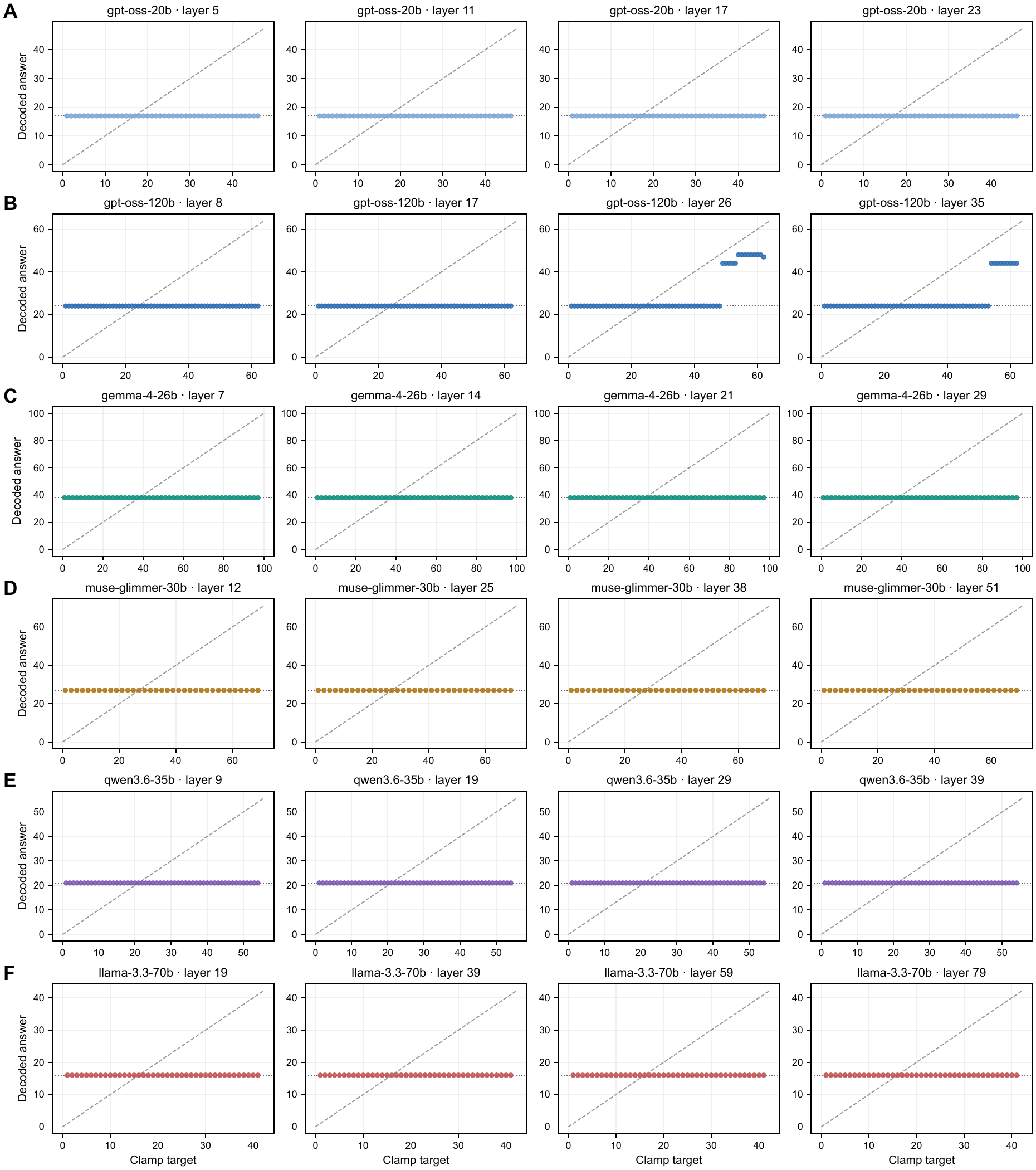}
\caption{\textbf{Clamps along the count direction leave the answer unchanged.} Decoded answer against clamp target at every site (models \textbf{A} to \textbf{F}, columns by depth). The dotted horizontal rule is the unedited answer; the dashed diagonal is where answers would lie if the coordinate were read as a counter. Only the two last layers of \texttt{gpt-oss-120b} depart, and only onto a single attractor value at extreme targets.}
\label{fig:suite_clamp}
\end{figure}

\begin{figure}[!htbp]
\centering
\includegraphics[width=\textwidth]{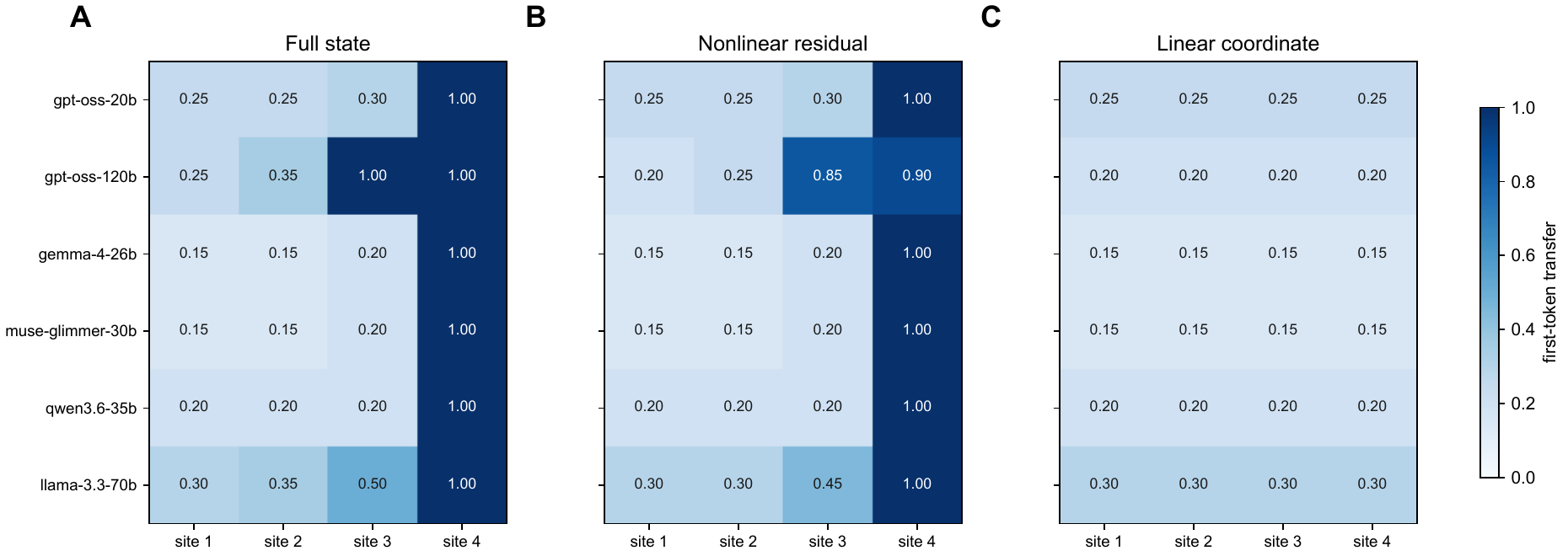}
\caption{\textbf{Chimera transfer.} First-token transfer rate of a donor answer into a base prompt for the full donor state (\textbf{A}), the nonlinear residual alone (\textbf{B}), and the linear coordinate alone (\textbf{C}), by model and depth. The residual reproduces the full state's transfer; the linear coordinate never exceeds the incidental base rate.}
\label{fig:suite_chimera}
\end{figure}

Three intervention families are applied at the final prompt token at every hooked site, about 1{,}500 greedy generations in total. \emph{Clamps and steps.} The state is clamped along the fitted count direction to targets spanning the stable regime and slightly beyond, or translated by multiples of the mean per-item trajectory step. The decoded answer is unchanged at essentially every site in every model although the edits are large: at the last layer of one model a twenty-count clamp moves the state by 57\% of the distance between genuine count states (Fig.~\ref{fig:suite_clamp}). The one exception is \texttt{gpt-oss-120b} at its two last hooked layers, where extreme targets near and past the boundary occasionally knock the answer onto a single attractor value rather than tracking the target. Random directions at the matched norm and at four times the norm also leave the answer unchanged. \emph{Chimera decomposition.} A donor prompt's decode-point state is split into its linear coordinate and everything else, and three transplants into a mid-regime base prompt are compared: the full donor state, the donor state with its linear coordinate clamped back to the base value, and the base state with only the donor's linear offset added. At the last layer the full state transfers the donor's answer in 20 of 20 trials per model and the residual transfers with the same profile, while the linear coordinate alone never transfers, at any depth of any model (Fig.~\ref{fig:suite_chimera}). Transfer rises from chance at early layers to complete at the final layers, so the answer is recomputed downstream around any single-position edit until it consolidates near the surface; the consolidation layer of each model is the counting column of Table~\ref{tab:round2_channel}. Because the linear coordinate holds only 3 to 9\% of the energy of the difference between two count states, the residual transplant is nearly a full-state transplant, and the decomposition establishes only that the linear coordinate is not what carries the answer. For tokenizers that split multi-digit numbers, a last-layer patch controls exactly the first emitted digit, so first-token transfer is the cross-model metric throughout.

\subsection{The output channel is the decoder's answer-token subspace, and it is not counting-specific}\label{app:subspace}

\begin{figure}[!htbp]
\centering
\includegraphics[width=\textwidth]{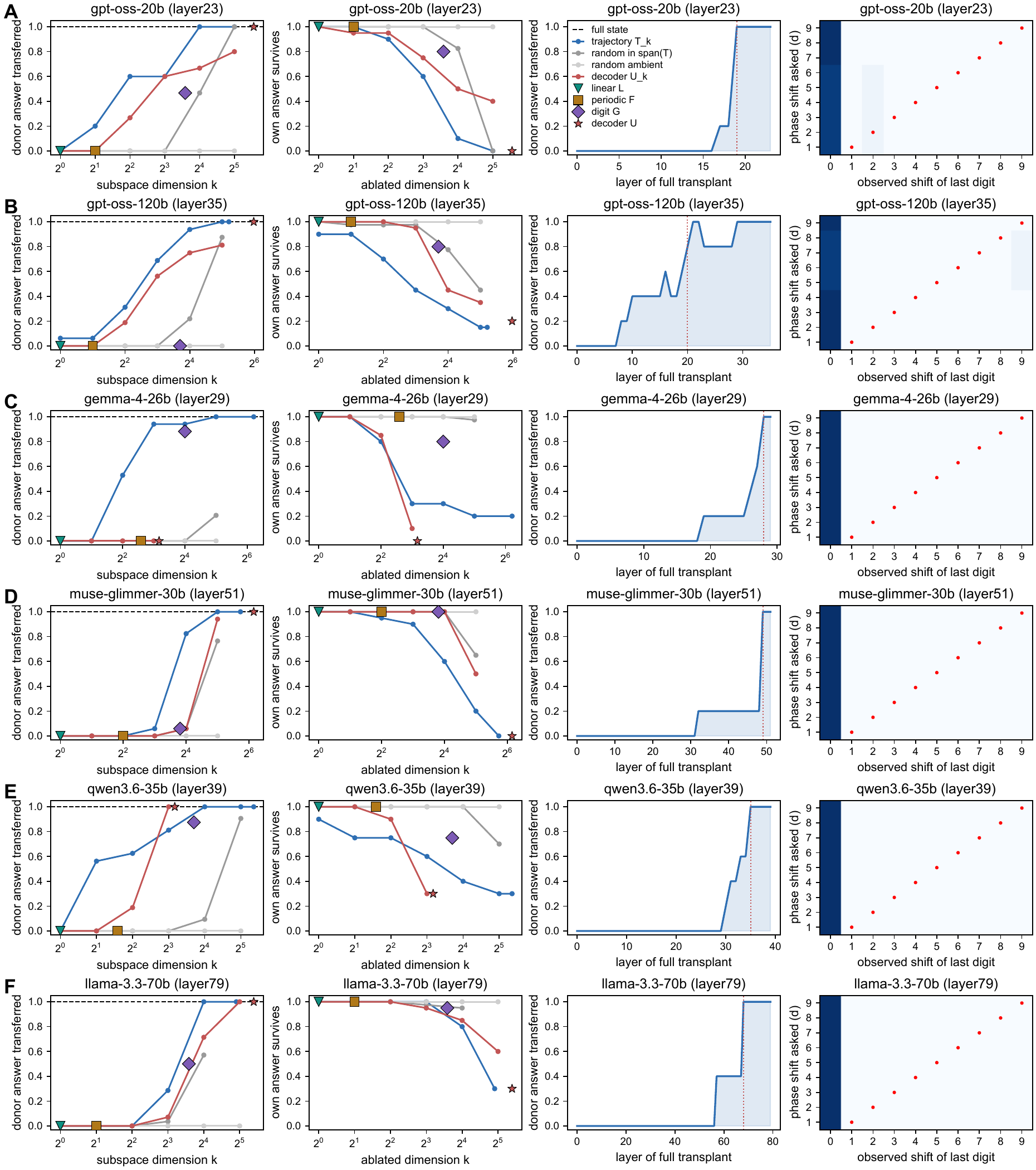}
\caption{\textbf{Subspace and periodic-feature controls.} One row per model (\textbf{A} to \textbf{F}) at its last hooked site. First column: first-token transfer of the donor answer against subspace dimension for trajectory components \(T_k\), random subspaces inside the trajectory span, random ambient subspaces, and the decoder's answer-token subspace \(U_k\); the single markers are the linear direction, the periodic pair, the digit subspace, and the full \(U\). Second column: survival of the model's own answer under mean ablation of the same subspaces. Third column: full-state transfer at every layer, with the consolidation layer marked. Fourth column: phase shift asked against observed shift of the last digit under rotation of the period-10 pair; a helix read by the decoder would fill the diagonal.}
\label{fig:suite_subspace}
\end{figure}

The single-prompt cache above has one state per count, so any difference between two count states lies in the span of the stable states, whose rank is the number of stable counts minus one (27 to 70). A claim about what carries the answer needs a richer cache and candidate subspaces tested for sufficiency and necessity. We cached the decode-point state for eight counted items (\texttt{a b x z cat dog red sun}, single tokens in every tokenizer) and two delimiters at every count up to 1.25 times the scanned capacity, keeping the 300 to 1{,}000 prompts per model whose teacher-forced answer is exact, with folds that hold out whole counts. The candidate subspaces, each an orthonormal basis, are the linear direction \(L\); the readable periodic directions \(F\); the digit-class directions \(G\); the decoder's own answer-token subspace \(U\), the norm-scaled unembedding rows of the answer tokens with no fitting; the top-\(k\) trajectory components \(T_k\) of the count-conditioned means; random \(k\)-subspaces inside the trajectory span \(R_k\); and random \(k\)-subspaces of the ambient space \(A_k\). Five causal groups ran at the last two hooked sites of every model, about 27{,}000 greedy generations, with the same base, donors and seed as the chimera rows: full donor-state transplant at every layer; subspace transplant \(h_{\text{base}}+P(h_{\text{donor}}-h_{\text{base}})\) for every family and its complement, scored by first-token transfer over the donors whose leading digit differs from the base's; interpolation \(h_{\text{base}}+\alpha(h_{\text{donor}}-h_{\text{base}})\) for \(\alpha\) from 0.1 to 0.9; mean ablation of a subspace, scoring survival of the model's own answer; and rotation of the fitted period-10 pair (Section~\ref{app:periodic}). Figure~\ref{fig:suite_subspace} shows the causal panels for every model.

\paragraph{Sufficiency and necessity.} At the last hooked site, projecting the donor-minus-base difference onto \(U\) transfers the donor's answer for every informative donor in five of six models, and the complement of \(U\), which holds 74 to 99\% of the difference's energy, transfers none; ablating \(U\) destroys the model's own answer in all six (0 to 6 of 20 survive), while \(L\), \(F\), and random ambient subspaces of up to 32 dimensions leave all 20 intact (Table~\ref{tab:subspace}). The exception, \texttt{gemma-4-26b}, decides its answer in a competition against its whole 262k-token vocabulary: the \(U\) projection moves the digit competition correctly for all 17 donors but the model then emits a non-digit token, so \(U\) is necessary (ablation 0 of 20) without passing the first-token test, while the fitted digit subspace \(G\), which moves the competition on-manifold, transfers 0.88. The trajectory components are privileged but not tiny: 2 to 16 of them reach half of the full-state transfer and 8 to 32 reach 90\%, whereas a random subspace inside the same span needs its full rank, and ablating the top 8 to 16 trajectory components destroys most answers in five models while the same number of random in-span directions destroys none. Over the last quarter of the network the trajectory and digit subspaces rotate toward \(U\) (principal cosines rising from 0.05--0.49 to 0.14--0.72), and a logit lens restricted to the answer tokens becomes decisive at the last site.

\begin{table}[H]
\centering
\footnotesize
\setlength{\tabcolsep}{2.5pt}
\caption{\textbf{Sufficiency and necessity at the last hooked site.} Transfer: first-token transfer rate of the donor's answer under a subspace transplant (informative donors, \(n=14\) to 17). Survival: the model's own first token surviving mean ablation of the subspace (of 20). \(d_{90}\): smallest \(k\) at which \(T_k\) reaches 90\% of the full-state transfer; rank: dimension of the trajectory span.}
\label{tab:subspace}
\begin{tabular}{lcccccccc}
\toprule
 & \multicolumn{4}{c}{Transfer} & \multicolumn{3}{c}{Survival} & \\
\cmidrule(lr){2-5}\cmidrule(lr){6-8}
Model & \(L\) & \(F\) & \(U\) & \(U\) complement & \(L\) & \(U\) & \(T_{16}\) & \(d_{90}(T)\) / rank \\
\midrule
\texttt{gpt-oss-20b} & 0 & 0 & 1.00 & 0 & 20 & 0 & 2 & 16 / 32 \\
\texttt{gpt-oss-120b} & 0 & 0 & 1.00 & 0 & 20 & 4 & 6 & 16 / 37 \\
\texttt{gemma-4-26b} & 0 & 0 & 0\(^{*}\) & 0 & 20 & 0 & 6 & 8 / 74 \\
\texttt{muse-glimmer-30b} & 0 & 0 & 1.00 & 0 & 20 & 0 & 12 & 32 / 53 \\
\texttt{qwen3.6-35b} & 0 & 0 & 1.00 & 0 & 20 & 6 & 8 & 16 / 41 \\
\texttt{llama-3.3-70b} & 0 & 0 & 1.00 & 0 & 20 & 6 & 16 & 16 / 30 \\
\bottomrule
\end{tabular}
\par\smallskip\raggedright\footnotesize \(^{*}\)The projection moves the digit competition correctly for every donor but the model emits a non-digit token from its full vocabulary; see text.
\end{table}

\begin{figure}[!htbp]
\centering
\includegraphics[width=\textwidth]{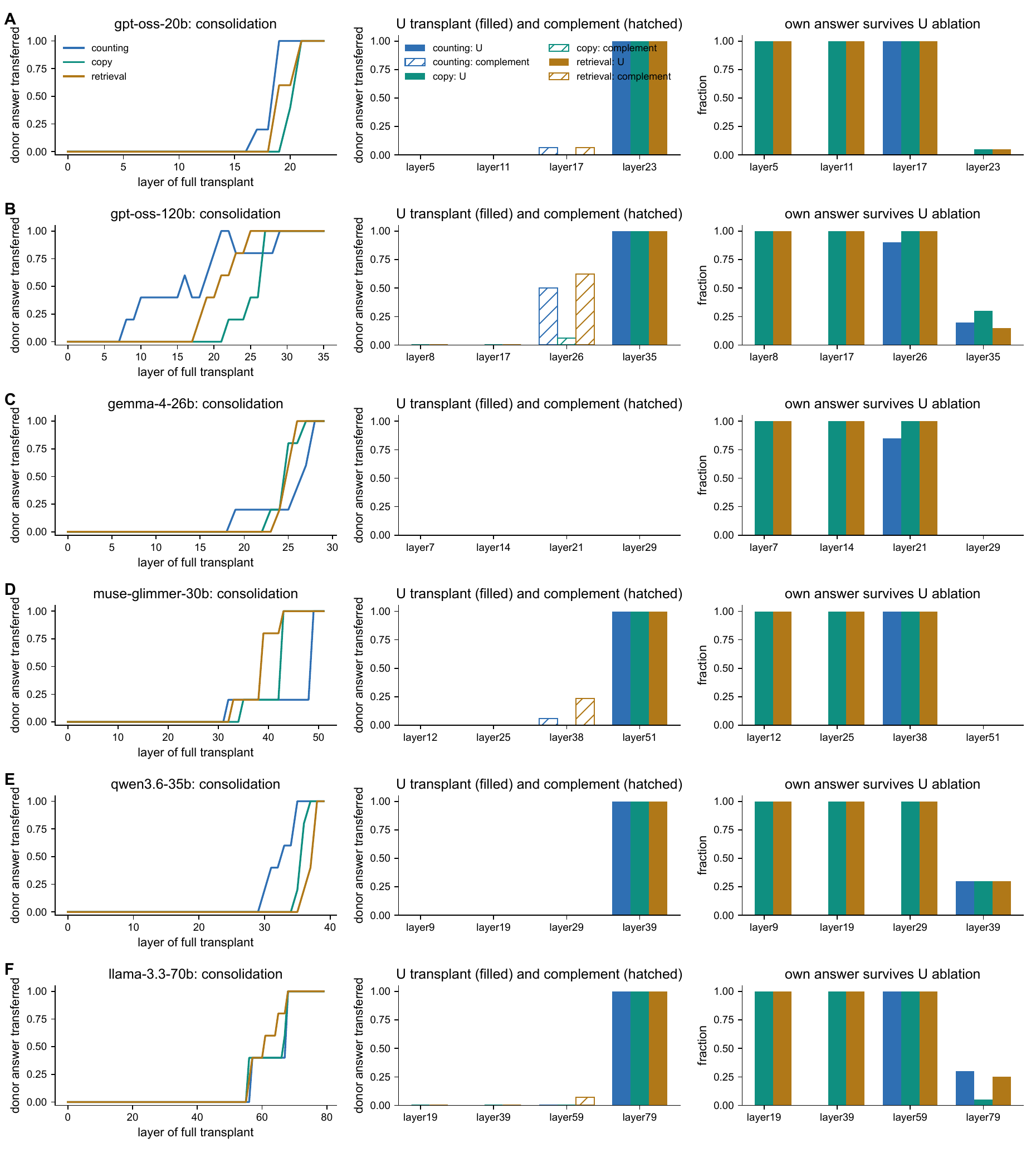}
\caption{\textbf{The output channel by task.} For each model (\textbf{A} to \textbf{F}): the fraction of donors whose answer a full-state transplant carries, by layer, for counting, copy, and retrieval (left); transfer through \(U\) and through its complement at the last site, by task (middle); and survival of the model's own answer under ablation of \(U\) and of the control subspaces, by task (right).}
\label{fig:round2_channel}
\end{figure}

\paragraph{The channel is the same for copying and retrieval.} \(U\) could carry the donor's answer for an algebraic reason: it is spanned by the answer tokens' own unembedding directions, so projecting a state difference onto it moves the answer logits by construction. To separate an output channel from a counting-specific code, two tasks with the same integer answers were run through the same battery: copying a stated number, and retrieving the value of a named key from a short list. At the last hooked site every number is the same for the two tasks as for counting: the \(U\) projection transfers the donor's answer for every informative donor in five models and for none in \texttt{gemma-4-26b}, the complement transfers nothing, ablating \(U\) destroys the model's own answer (0 to 6 of 20 survive for every task), ablating the linear, periodic, or random subspaces destroys nothing, and the interpolation switch falls between \(\alpha=0.5\) and \(0.6\) for every task (Fig.~\ref{fig:round2_channel}). The depth at which the channel fills is not counting-specific either (Table~\ref{tab:round2_channel}): a full donor-state transplant carries the copied or retrieved answer only from a layer within a few layers of the counting consolidation layer, later in three models and earlier in two, never early. \texttt{gpt-oss-120b} consolidates counting from layer 20 through the complement of \(U\) (transfer 0.56 to 0.75 with \(U\) at 0.00 to 0.06 over layers 20 to 26) and switches into \(U\) at layer 27 of 36: a pre-output code converted into the output code in the last quarter. What is counting-specific is therefore not the channel or when it fills, but what fills it.

\begin{table}[H]
\centering
\footnotesize
\caption{\textbf{The output channel fills late for every number-producing task.} First layer at which a full donor-state transplant carries at least 80\% of the donors' answers, by task; the counting column is the consolidation layer of Section~\ref{app:decode}.}
\label{tab:round2_channel}
\begin{tabular}{lrrrr}
\toprule
Model & Layers & Counting & Copy & Retrieval \\
\midrule
\texttt{gpt-oss-20b} & 24 & 19 & 21 & 21 \\
\texttt{gpt-oss-120b} & 36 & 20 & 27 & 23 \\
\texttt{gemma-4-26b} & 30 & 28 & 25 & 26 \\
\texttt{muse-glimmer-30b} & 52 & 49 & 43 & 39 \\
\texttt{qwen3.6-35b} & 40 & 35 & 36 & 38 \\
\texttt{llama-3.3-70b} & 80 & 68 & 68 & 65 \\
\bottomrule
\end{tabular}
\end{table}

\paragraph{The switch does not pass through the integers between the answers.} Interpolating between two count states and reading only the decoded answer cannot distinguish a categorical code from a continuous one, since the argmax turns any linear crossing into a switch. Recording the answer-token logits along the interpolation at the last site shows what happens to the integers between the two answers (Table~\ref{tab:round2_logits}). The base and donor logits cross between \(\alpha=0.5\) and \(0.55\) in every model. In the 79 pairs whose leading digits admit an integer token between the answers, an integer more than two away from both leads in one (a \texttt{gpt-oss-120b} pair), the immediate neighbors of the base answer lead briefly in six (five in \texttt{gpt-oss-20b}, one in \texttt{gpt-oss-120b}), where they hold a third of the answer-token mass near the crossing, and in four models no intermediate of any kind leads.

\begin{table}[H]
\centering
\footnotesize
\caption{\textbf{Logits along the interpolation.} Pairs whose leading digits admit an intermediate token; pairs in which an intermediate integer ever leads, split by whether it lies within two of an endpoint; softmax mass over the answer tokens on intermediate integers at \(\alpha=0.5\) (total, and on integers farther than two from both endpoints); and the \(\alpha\) at which the decoded answer switches.}
\label{tab:round2_logits}
\setlength{\tabcolsep}{3pt}
\begin{tabular}{lrrrrr}
\toprule
Model & Pairs & Neighbour leads & Farther leads & Mass at 0.5 (total / far) & Switch \(\alpha\) \\
\midrule
\texttt{gpt-oss-20b} & 15 & 5 & 0 & 0.33 / 0.04 & 0.55 \\
\texttt{gpt-oss-120b} & 16 & 1 & 1 & 0.33 / 0.10 & 0.55 \\
\texttt{gemma-4-26b} & 12 & 0 & 0 & 0.00 / 0.00 & 0.50 \\
\texttt{muse-glimmer-30b} & 17 & 0 & 0 & 0.08 / 0.02 & 0.55 \\
\texttt{qwen3.6-35b} & 5 & 0 & 0 & 0.01 / 0.00 & 0.55 \\
\texttt{llama-3.3-70b} & 14 & 0 & 0 & 0.00 / 0.00 & 0.50 \\
\bottomrule
\end{tabular}
\end{table}

\subsection{Periodic and digit features are correlates}\label{app:periodic}

The multi-prompt cache also supports readouts of periodic phases \(\cos\) and \(\sin(2\pi n/P)\) for \(P\in\{2,5,10,20\}\), flagged readable when both phases reach held-out \(R^2\ge0.5\) and beyond a smoothness null when they beat the same target computed from the linear readout, and of the last and tens digits by shrinkage discriminant analysis with a count-level permutation null. At the earliest sites a held-out count's last digit is decoded as a neighboring digit 0.27 to 0.79 of the time, as a smooth analog code would produce; in the second half of the network the last digit becomes decodable (0.33 to 0.42 in the two \texttt{gpt-oss} models, 0.61 to 0.85 in \texttt{gemma-4-26b}, \texttt{muse-glimmer-30b} and \texttt{qwen3.6-35b}, against a permutation ceiling of 0.15 to 0.22) and that confusion drops. Period-10 phases are readable at the last site of every model and beyond the smoothness null in two. Where the digit code is strongest, at the third hooked site of \texttt{gemma-4-26b} and \texttt{muse-glimmer-30b}, nothing is causal: no family at any dimension transfers, ablates, or rotates the answer, because those sites precede the consolidation layer. Rotating the period-10 pair at the last site by the phase of \(+d\) counts, \(d=1\ldots9\), moves the answer in 0 to 6 of 180 prompts per model, never in agreement with the target phase and no more often than a random plane of the same span (Fig.~\ref{fig:suite_subspace}, fourth column). The periodic and digit features are correlates, like the linear coordinate, with the one exception of \texttt{gemma-4-26b}'s digit subspace noted above.

\subsection{The accumulation is read at mid-depth across item positions}\label{app:round2}

\begin{figure}[!htbp]
\centering
\includegraphics[width=\textwidth]{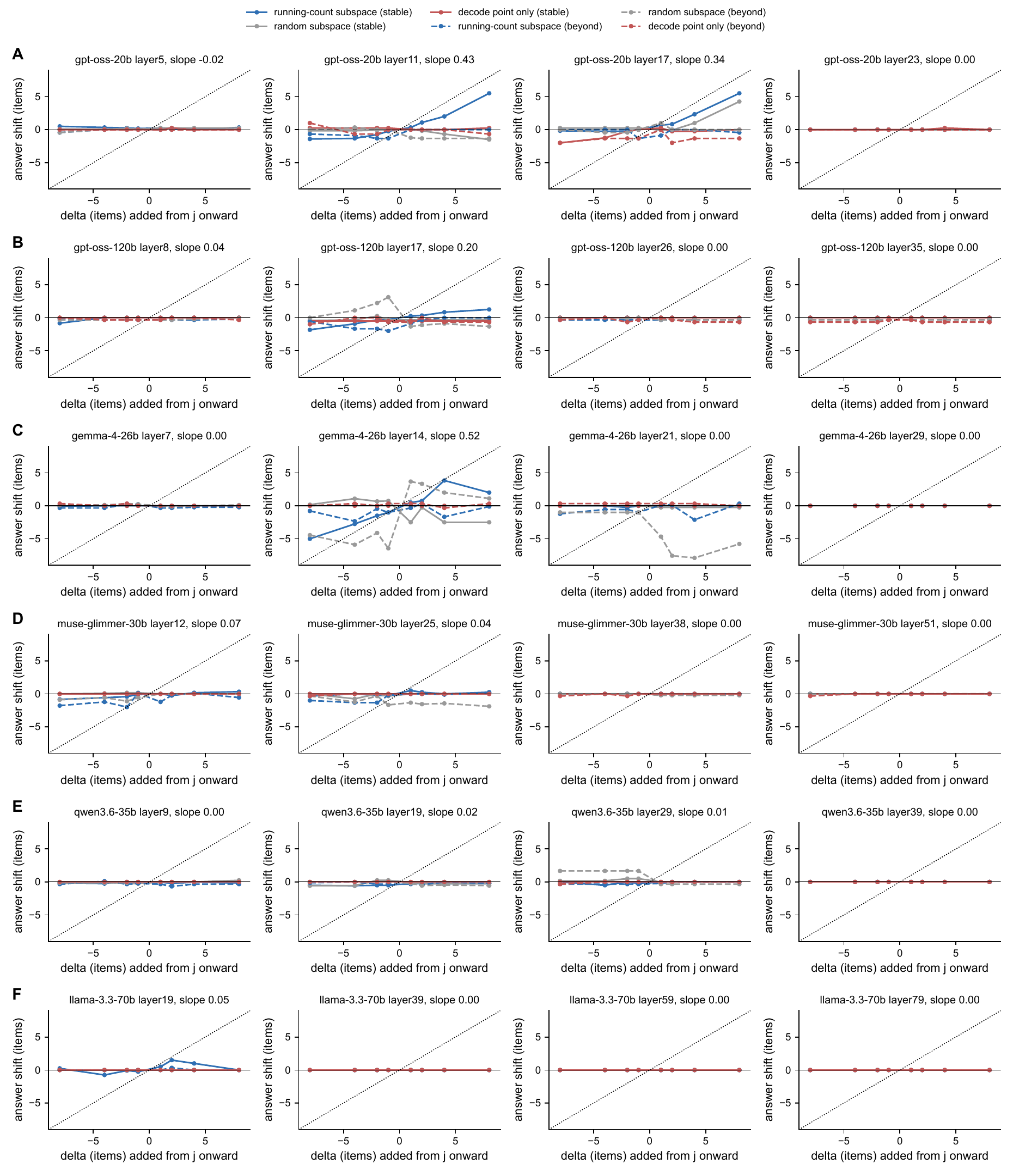}
\caption{\textbf{Shifting the accumulation across item positions.} Answer shift against \(\delta\) for each model (\textbf{A} to \textbf{F}, rows) at the four hooked sites (columns), pooled over prompt lengths and onset positions; the count-subspace edit is drawn against the matched random direction and the decode-point-only edit, with the identity line for reference.}
\label{fig:round2_accumulation}
\end{figure}

The decode-point edits show that the fitted coordinate is not read where the answer is emitted; they say nothing about whether the accumulation is read upstream. We therefore intervened on the running count at the item positions themselves. Because attention is causal and every prompt shares its prefix, one forward pass at 1.25 times CC plus eight items yields the state of every item token for every shorter prompt at all four sites; a ridge readout of the running count from those states holds at every site (leave-one-out \(R^2\) 0.93 to 0.99), and the subspace \(S_k\) spanned by the top trajectory components (\(k\) the dimension that carried 90\% of decode-point transfer) holds 64 to 99.9\% of the item-state variance. The edit replaces the \(S_k\) component of the state at running count \(i\) by the trajectory's own component at count \(i+\delta\), at every item position from a chosen onset (a quarter, half, or three quarters of the way through), for prompts at 0.35 to 0.9 of CC and \(\delta\) from \(-8\) to \(+8\). The controls apply the same edit through a fixed random direction of matched norm, and at the decode point only. The answer shift is regressed on \(\delta\) (Table~\ref{tab:round2_shift}, Fig.~\ref{fig:round2_accumulation}).

In three models the answer follows the shifted accumulation at one or two mid-depth sites, after the first hooked site and before consolidation: \texttt{gpt-oss-20b} at layers 11 and 17 of 24, \texttt{gemma-4-26b} at layer 14 of 30, and \texttt{gpt-oss-120b} at layer 17 of 36, with slopes of 0.20 to 0.52 that reach 0.5 to 0.9 on individual prompts, while at those sites the random direction is flat or of the opposite sign and the decode-point edit is flat, except at \texttt{gpt-oss-20b}'s third hooked site (layer 17), where the random shift reaches 0.20 and the decode-point edit 0.12 against 0.34. At the first hooked site the slope is zero in every model, so the count there is recomputed downstream from the tokens, and after consolidation it is zero, so the decoder no longer reads the item positions once the answer has formed. In \texttt{muse-glimmer-30b}, \texttt{qwen3.6-35b}, and \texttt{llama-3.3-70b} no hooked site exceeds 0.08; their causal window may fall between hooked sites, or their decoders may read the count through a route the edit does not touch. Beyond CC the shift moves nothing at any site of any model, while at \texttt{gemma-4-26b}'s causal site a random shift of the same norm now moves the answer strongly: past the boundary the state is fragile but no longer steerable through the count subspace. Compressing the accumulation so that the last item reads as the in-regime maximum rescues no model except \texttt{gemma-4-26b}, which then answers the compressed count. In half of the models the answer therefore follows the accumulation causally, with a gain below one; in the other half the evidence remains correlational.

\begin{table}[H]
\centering
\footnotesize
\caption{\textbf{Answer shift per item of accumulation shift.} Least-squares slope of the answer change on \(\delta\), pooled over prompt lengths and onset positions, with a bootstrap 95\% interval, at the responsive site of the three responsive models and at the most responsive site of the other three; the two controls are the same edit through a fixed random direction of matched norm and at the decode point only; the last column is the slope for prompts from 1.0 to 1.25 times CC.}
\label{tab:round2_shift}
\begin{tabular}{llrrrr}
\toprule
Model & Site & Slope [95\%] & Random & Decode point & Beyond CC \\
\midrule
\texttt{gpt-oss-20b} & layer 11 of 24 & 0.43 [0.32, 0.53] & \(-0.07\) & \(-0.01\) & 0.08 \\
 & layer 17 of 24 & 0.34 [0.24, 0.45] & 0.20 & 0.12 & \(-0.02\) \\
\texttt{gpt-oss-120b} & layer 17 of 36 & 0.20 [0.13, 0.28] & 0.00 & 0.00 & 0.09 \\
\texttt{gemma-4-26b} & layer 14 of 30 & 0.52 [0.33, 0.74] & \(-0.24\) & 0.00 & 0.06 \\
\texttt{muse-glimmer-30b} & layer 12 of 52 & 0.07 [0.02, 0.14] & \(-0.01\) & 0.00 & 0.10 \\
\texttt{qwen3.6-35b} & layer 19 of 40 & 0.02 [0.00, 0.06] & 0.02 & 0.00 & \(-0.01\) \\
\texttt{llama-3.3-70b} & layer 19 of 80 & 0.05 [0.01, 0.10] & 0.00 & 0.00 & 0.00 \\
\bottomrule
\end{tabular}
\end{table}

\subsection{The boundary state carries the count forward, and the same replacement carries the state of an object history}\label{app:boundary}

A third intervention asks where the running count is handed from one item to the next. A recipient sequence of \(C_B\) items and a donor sequence of \(C_D\) items are run separately; the residual states of the donor's last items are written into the recipient's last item positions at every layer from an entry layer upward, with the recipient's own states restored at every earlier position so that the edit cannot leak backward through the mixture-of-experts batching; the recipient's text is left in the context; then \(r\) further items are appended that neither run has seen, and the model is asked for the count. If the continuation starts from the state the window carries, the answer is \(C_D + r\); if it re-reads the text, \(C_B + r\). Controls are a sham edit (the recipient's own states written back), a same-count donor (\(C_D = C_B\) under different padding), a random-source window (the donor's states from an earlier span of its sequence), a position factorial (donor and recipient paddings crossed), and the whole donor cache as the ceiling. The operator was chosen on calibration pairs by a rule fixed in advance (the smallest window and then the fewest layers among candidates at or above 80\%) and frozen before held-out pairs were run.

\paragraph{Counting.} In \texttt{gpt-oss-20b} a window of the last eight items entered at any layer from 2 to 8 gives the predicted count on every calibration instance, and entered at layer 9 or above on none; windows of one, two, or four items carry nothing, and replacing keys and values instead of residual states carries nothing. The frozen operator (eight items, layers 8 to 23) gives \(C_D + r\) exactly on 10 of 13 strictly eligible held-out instances (bootstrap 95\% interval 0.50 to 1.00) and within one item on 12 of 13; sham edits change none of 13, same-count donors leave 8 of 8 unchanged, and random-source windows give the count their own items carry plus \(r\) rather than \(C_D + r\), so the continuation starts from whatever running count the transplanted items encode. In \texttt{gemma-4-26b} the window must be twelve items; entered at layer 8 it is exact on 9 of 20 held-out instances, on every instance where the transplanted count is at or below the count the recipient's own positions carry (9 of 9), and biased by one to several items above that; shams and same-count donors are unchanged, and random-source windows again follow the count they carry (37 of 45). In \texttt{gpt-oss-120b} the frozen operator enters at layer 5 and gives 7 of 18. The highest entry layer that still carries the count is nearly the same in all three models, layer 8 or 9 of 24, 30, and 36 (entry at layer 9 or 10 and above carries nothing), upstream of the mid-depth sites where the item-position shifts act. The route is position-sensitive: donor keys and values, which carry the donor's positional phase, transfer nothing, the same donor states recomputed at the recipient's positions do, and padding that moves the recipient's positions toward the donor's restores exactness (7 of 11 in the factorial). The running count that the decoder reads is therefore recomputed from the states of the last eight to twelve items through a route that enters at layer 8, not inherited through a single boundary state.

\paragraph{An object-tracking task with the same signature.} To ask whether these states are counting-specific, the same operator was applied, without any retuning, to a task with no count in it, a state-maintenance task in which history length (the number of swaps) and state size (the number of objects) vary independently and access to the record is tested separately from the state. Three to five named objects start at named locations, a history of swaps moves them, and after the history the model is asked, with immediate answering and randomized answer labels, for the location of an object the last swap moved, the location of an object no swap touched, what the last swap said, and, with the correct current state written into the prompt, a location question that needs no tracking. Table~\ref{tab:object_grid} gives \texttt{gemma-4-26b}'s competence. One swap is tracked perfectly at every state size and two swaps at 0.88 to 0.94, while three and four swaps fall to 0.40 to 0.75, with the untouched object misplaced in most histories from three swaps on; access to the last swap's text is 48 of 48 in every cell and the supplied-state answer is 0.83 to 1.00. The failures are in the state update, with the events still readable, which is the signature counting shows.

\begin{table}[H]
\centering
\footnotesize
\caption{\textbf{Object tracking in \texttt{gemma-4-26b}, immediate answering.} Exact-location rate for an object the last swap moved and for an object no swap touched, the rate of reporting the last swap's text correctly, and the rate on the supplied-state question, by number of objects and swaps; 48 histories per cell.}
\label{tab:object_grid}
\begin{tabular}{rrrrrr}
\toprule
Objects & Swaps & Moved object & Untouched object & Last swap reported & Supplied state \\
\midrule
3 & 1 & 1.00 & 1.00 & 1.00 & 1.00 \\
3 & 2 & 0.90 & 0.35 & 1.00 & 1.00 \\
3 & 3 & 0.60 & 0.29 & 1.00 & 1.00 \\
3 & 4 & 0.40 & 0.29 & 1.00 & 0.83 \\
4 & 1 & 1.00 & 1.00 & 1.00 & 1.00 \\
4 & 2 & 0.88 & 0.98 & 1.00 & 1.00 \\
4 & 3 & 0.75 & 0.31 & 1.00 & 1.00 \\
4 & 4 & 0.52 & 0.17 & 1.00 & 0.88 \\
5 & 1 & 1.00 & 1.00 & 1.00 & 1.00 \\
5 & 2 & 0.94 & 1.00 & 1.00 & 1.00 \\
5 & 3 & 0.67 & 0.62 & 1.00 & 0.98 \\
5 & 4 & 0.71 & 0.25 & 1.00 & 0.98 \\
\bottomrule
\end{tabular}
\end{table}

\paragraph{Transplanting the boundary state of an object history.} The recipient is a two-swap history. The donor is the same history with the first swap replaced by another pair of locations or by a swap of a location with itself, so that the two histories reach different states while the second swap's text is identical; that identical line is the window (twelve to thirteen tokens, layers 8 to 29, the earlier positions restored as above, nothing at the decode point, the recipient's full history in the context). From one intervened prefix the model is asked for the location of every object whose true and donor states differ (prediction: the donor's state), of an object whose location agrees (prediction: unchanged), which locations the first swap named, with the donor's pair among the options (prediction: the real text), and the supplied-state question (prediction: unchanged). Controls are the sham edit and the whole donor cache. A first design with the donor differing at the last swap and the window on that swap's own line moved every changed object (144 of 144) but overwrote the event: asked what the last swap said, the model reported the donor's pair 48 of 48 times, so that design replaces text rather than state and was discarded. Table~\ref{tab:object_transplant} and Fig.~\ref{fig:mechanistic_probe}G give the result of the second design.

\begin{table}[H]
\centering
\footnotesize
\caption{\textbf{The counting operator applied to an object history in \texttt{gemma-4-26b}.} Replies to location questions about objects whose true and donor states differ that give the donor's state, under the edit, the sham edit, and the whole donor cache; then, under the edit, the rate of reporting the first swap's real text, the rate at which an object whose location agrees between the true and donor states stays put (``untouched''; this differs from Table~\ref{tab:object_grid}'s never-moved object), and the rate on the supplied-state question. Two-swap histories; 48 per cell; ``replace'' donors differ in the first swap's pair and ``omit'' donors skip it.}
\label{tab:object_transplant}
\begin{tabular}{rlrrrrrr}
\toprule
Objects & Donor & Edit & Sham & Whole cache & Swap 1 real & Untouched & Supplied \\
\midrule
3 & replace & 81/144 & 23/144 & 102/144 & 48/48 & -- & 40/48 \\
3 & omit & 57/96 & 32/96 & 75/96 & 48/48 & 45/48 & 36/48 \\
4 & replace & 44/150 & 19/150 & 98/150 & 48/48 & 42/42 & 43/48 \\
4 & omit & 45/96 & 37/96 & 88/96 & 48/48 & 48/48 & 42/48 \\
5 & replace & 43/156 & 20/156 & 118/156 & 48/48 & 48/48 & 44/48 \\
5 & omit & 42/96 & 31/96 & 87/96 & 48/48 & 48/48 & 41/48 \\
\bottomrule
\end{tabular}
\end{table}

The edit moves the changed objects toward the donor's state in 28 to 59\% of questions against 13 to 39\% for the sham edit and 65 to 92\% for the whole donor cache. What moves is the state: the first swap is still reported correctly in 48 of 48 histories in every cell, an untouched object is displaced in at most 3 of 48, and the supplied-state answer stays correct in 36 to 44 of 48, against 48 of 48 with no edit. The same replacement, at the same layers, therefore carries a state on both tasks: the running count in counting, and the tracked locations in an object history, with the history's text left readable.

\paragraph{Under native reasoning the model re-reads the history.} On 40 two-swap histories (24 with three objects, 16 with four) in which the immediate-answer edit had moved the changed object to the donor's location, the same edit was applied with the model's native reasoning enabled (1{,}536 reasoning tokens; one reply truncated and excluded). The changed object is placed correctly in 23 of 23 and 15 of 16 complete edited replies, in every untouched and sham reply, and the first swap is reported correctly in every reply of every condition. The traces restate the initial assignment and apply both swaps in order before answering; the edited traces are about 40\% longer than the untouched ones (203 against 135 and 210 against 165 reasoning tokens), which is the only footprint the transplanted state leaves. When the history is a sufficient record of the state, reasoning rebuilds the state from it and the transplant is undone. On counting the history is not such a record, because a written value cannot be re-attached to the indistinguishable event where it was written, and generated reasoning does not restore the count (Section~\ref{sec:dynamics}, Appendix~\ref{app:copy}). The same intervention carries the state in both tasks, which is consistent with a shared substrate without identifying one; the tasks differ in whether the text can rebuild it.

\subsection{Count versus length: readable, not routed}\label{app:mixed}

\begin{figure}[!htbp]
\centering
\includegraphics[width=\textwidth]{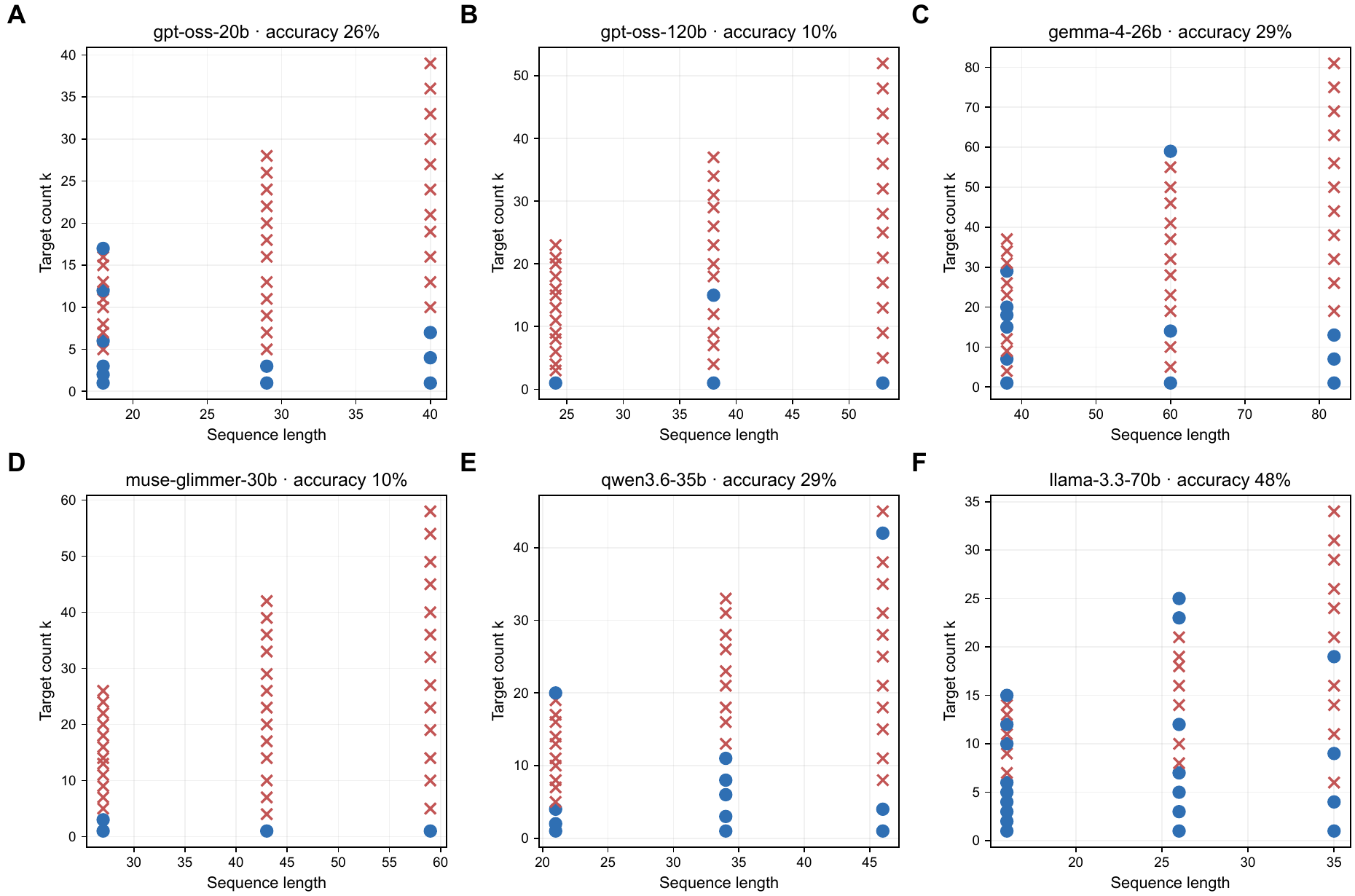}
\caption{\textbf{Mixed-sequence deconfound.} Correctness of the immediate answer over the \(k \times L\) grid for each model (\textbf{A} to \textbf{F}); circles are correct trials and crosses errors. Every probed \(k\) lies far inside the model's homogeneous-task stable regime.}
\label{fig:suite_mixed}
\end{figure}

On mixed sequences, in which the target item appears \(k\) times inside a sequence of total length \(L\) and the two are varied independently within the stable regime, the count and the length are recovered by separate readouts (count \(R^2\) 0.70 to 0.99, length 0.37 to 1.00) along nearly orthogonal directions (\(|\cos|\) 0.01 to 0.42): the accumulator encodes the count separately from the length. Immediate-answer accuracy on the same grids collapses to between 10\% and 48\% (Fig.~\ref{fig:suite_mixed}), with every \(k\) far inside the model's homogeneous-task stable regime. A readout fitted within the grid recovers the count to within one item in 0.43 to 0.86 of trials, above the model's own within-one rate in four of six models by 0.12 to 0.55, and exactly in 0.19 to 0.29, above the model's exact rate in three models (\texttt{gpt-oss-20b}, \texttt{gpt-oss-120b}, and \texttt{muse-glimmer-30b}, by 0.03, 0.09, and 0.16) and below it in the other three (Table~\ref{tab:round2_exact}); the homogeneous-task readout does not transfer to the grids. The claim is therefore about a magnitude estimate within one item, decodable at an early site where the model's own answer is not, with exact decoding ahead of the model in half of the models: count information the model cannot route to its own output.

\subsection{Reasoning modes}\label{app:modes}

\begin{figure}[!htbp]
\centering
\includegraphics[width=\textwidth]{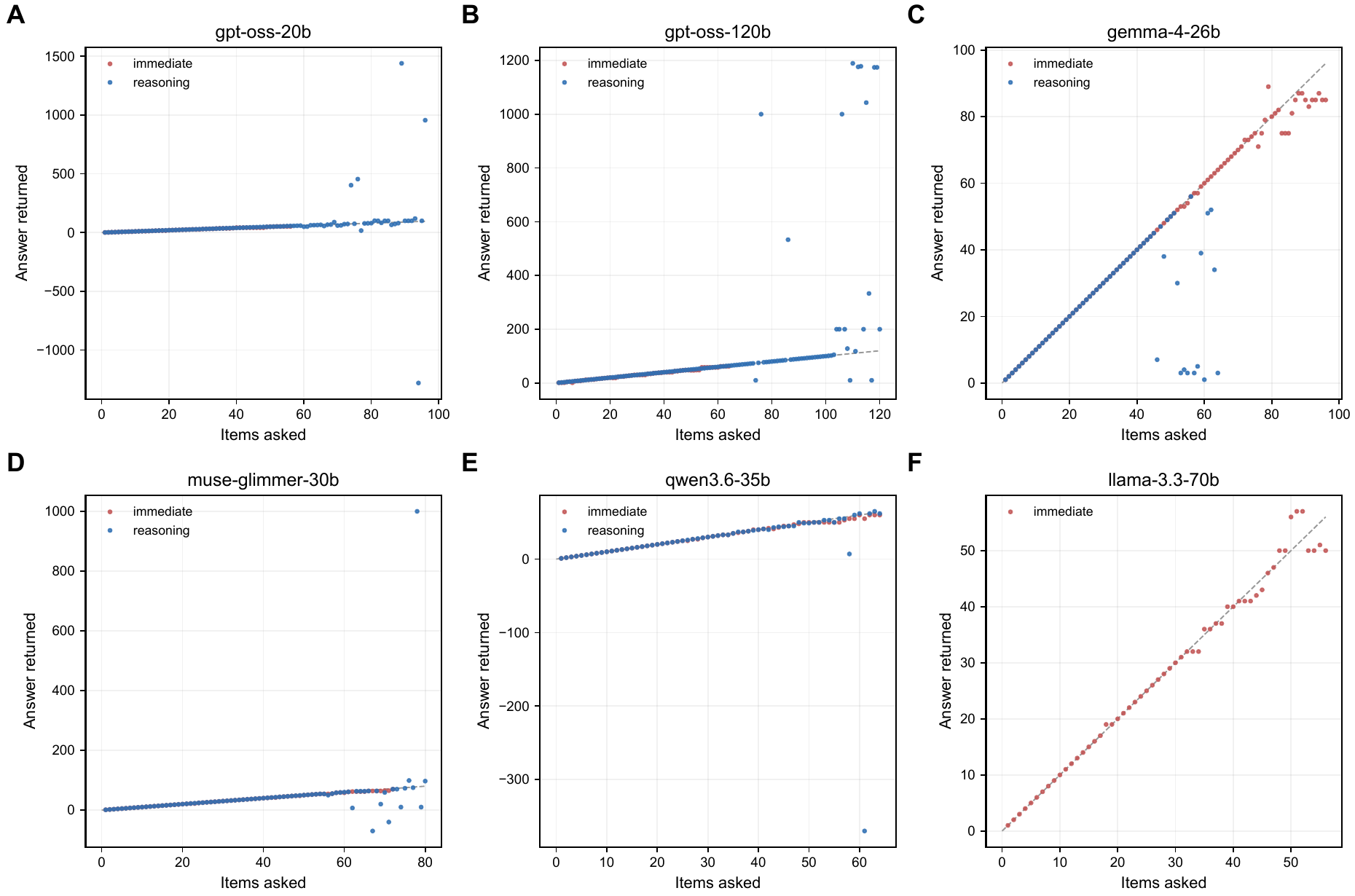}
\caption{\textbf{Immediate-answer and reasoning scans.} Answer returned against items asked in both modes for each model (\textbf{A} to \textbf{F}). Reasoning raises capacity in the two \texttt{gpt-oss} models, lowers it in \texttt{gemma-4-26b}, and leaves \texttt{muse-glimmer-30b} and \texttt{qwen3.6-35b} unchanged.}
\label{fig:suite_modes}
\end{figure}

Immediate-answer and native reasoning scans for all models are shown in Fig.~\ref{fig:suite_modes}. Two models, siblings from one family, gain 64\% and 113\% from reasoning; their analysis channels re-enumerate the sequence with periodic subtotals, and their silent readouts are the noisiest of the six, so externalizing wins. One model loses 40\%: its thinking traces enumerate, distrust the total, re-count in groups, and loop until the token budget truncates them, replacing a better silent mechanism with a worse text-mediated one. Two are unchanged, and the sixth, \texttt{llama-3.3-70b}, has no native reasoning mode. Scaling within the one family raises the text-mediated capacity, not the internal one: the larger sibling's best held-out readout error is five times worse than the deepest model's (\texttt{llama-3.3-70b}), while its reasoning capacity doubles the smaller sibling's.

\FloatBarrier

\section{Cross-benchmark correlations}\label{app:benchmark}

To compare CC with standard task performance, we cross-referenced model results against public benchmark metrics, including GPQA Diamond, ARC-AGI-2, SWE-bench Verified, and OTIS Mock AIME \citep{rein2023gpqa,chollet2025arcagi2,jimenez2024swebench,swebenchverified2024,epoch2025otis}. Correlation coefficients were calculated across intersecting model subsets for 66 public benchmarks (44 sharing at least ten models with the evaluation set).

\begin{figure}[htbp]
\centering
\includegraphics[width=0.5\textwidth]{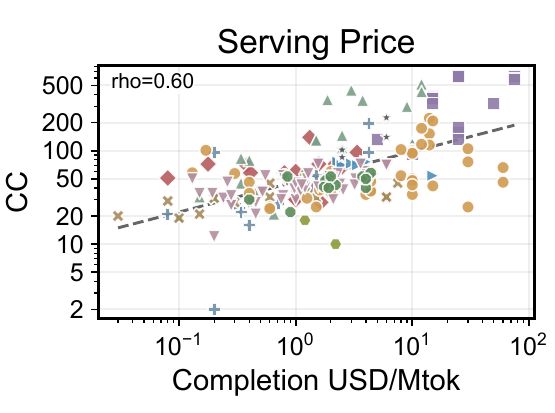}
\caption{\textbf{Serving price against counting capacity.} Completion price per million tokens (2026-08-25 snapshot) against CC, one point per model, colored by provider.}
\label{fig:price}
\end{figure}

\begin{figure}[htbp]
\centering
\includegraphics[width=0.38\textwidth]{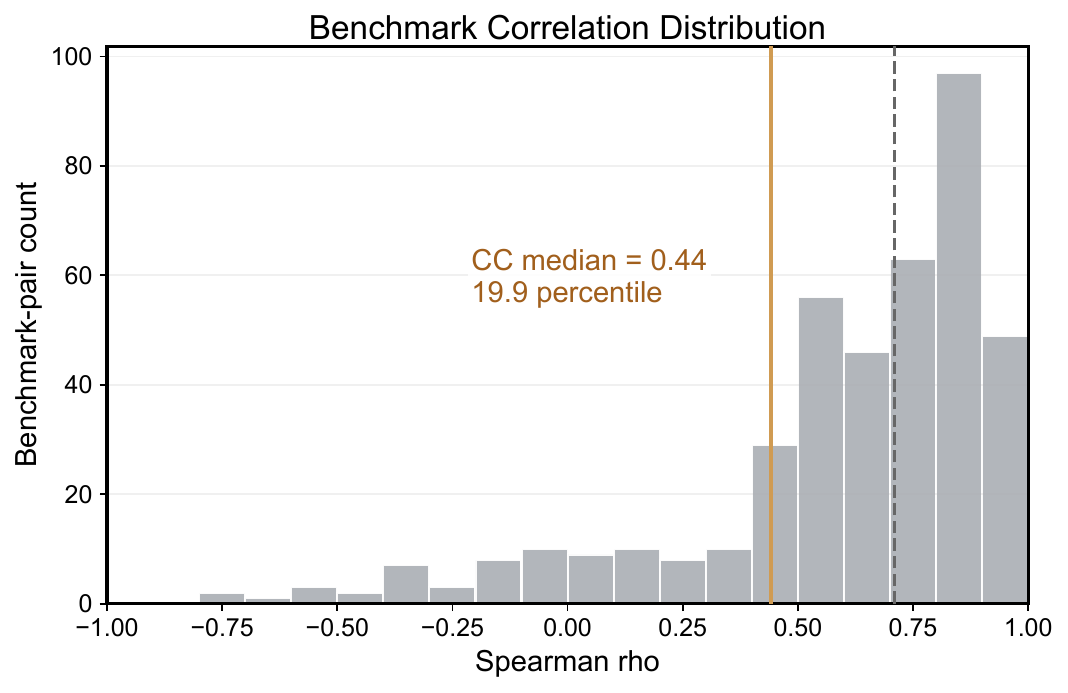}
\caption{\textbf{Distribution of benchmark correlations with counting capacity.} Rank-correlation values across evaluated public benchmark metrics summarize how weakly standard task leaderboards align with task-modeling capacity.}
\label{fig:benchmark_corr_distribution}
\end{figure}

The aggregate comparisons (the METR time horizon and total parameter count) appear alongside the headline benchmarks in the main text (Fig.~\ref{fig:landscape}), and serving price is given below; the Epoch Capabilities Index reaches \(\rho=0.66\) over 103 shared models and appears with every other benchmark in the panel pages below. Figure~\ref{fig:benchmark_corr_distribution} summarizes the individual benchmarks as a distribution of rank correlations. Serving price tracks CC as well (Fig.~\ref{fig:price}): completion price per million tokens from a dated snapshot of live prices (2026-08-25) correlates with CC at \(\rho=0.60\), approximately price \(\propto\) CC\(^{1.1}\), within vendors as well as across them; price is confounded with model size, so this is the market's estimate of usefulness rather than a separate measurement.

 Parameter count is the one scale variable available: 103 models have a public total count (\(\rho=0.69\) with CC, Fig.~\ref{fig:landscape}D), and for the 53 that also report an active count, total parameters predict CC far better than active parameters, with Spearman \(\rho\) of 0.65 against 0.16, a partial correlation of 0.71 for total given active and \(-0.39\) for active given total, and a leave-one-out \(R^2\) of 0.35 from total alone against 0.39 with active added, the added coefficient being negative. Within the 29 mixture-of-experts models the pattern is the same (0.66 against 0.25), and a smaller active fraction goes with a higher CC (\(\rho=-0.72\) against active/total), so the capacity tracks the parameters a model has rather than those it spends per token; the sample is the open-weight part of the catalogue, and the counts come from model cards and slugs, audited against vendor documentation for the mixture models. The three largest rank correlations in the panels belong to Balrog (\(\rho=0.87\) over 16 shared models), WeirdML (0.80 over 49), and the METR time horizon (0.78 over 27). Because frontier models tend to be better at everything, we checked which of these relations survives conditioning on general capability, with the Epoch Capabilities Index as the control. For METR, on the 24 models with both scores, \(\rho(\mathrm{CC},\mathrm{METR})=0.83\); the partial correlation controlling for the index is 0.74, and the reverse partial, index with METR controlling for CC, is 0.86, so the two carry independent information about task duration. For Balrog and WeirdML the same partial correlations are 0.07 (13 models) and 0.30 (43 models): their raw correlations with CC are carried by general capability, with which the two are correlated at \(\rho\) of 0.85 and 0.95. An incremental test agrees. When each score is predicted from the index alone and from the index plus \(\log \mathrm{CC}\), with every vendor held out in turn, adding CC raises out-of-sample \(R^2\) for the log time horizon from 0.81 to 0.90 (0.86 to 0.92 holding out one model at a time) and moves Balrog and WeirdML by at most 0.02 (0.72 to 0.70, and 0.89 to 0.91). Leaving out any one vendor gives \(\rho\) between 0.72 and 0.88 for METR, and the relation holds within the largest vendor group (\(\rho=0.81\) over 13 OpenAI models). The vendor folds behind that gain hold 11, 6, 4, 2, and 1 of the 24 models; a bootstrap over models puts the gain at 0.06 with a 95\% interval from \(-0.24\) to 0.30 (\(P(\text{gain}>0)=0.79\)), so the incremental prediction is suggestive rather than established. The same bootstrap puts the partial correlation at 0.74 with a 95\% interval from 0.39 to 0.87, and dropping the six Anthropic models leaves 0.65 on 18. The three evaluations above were picked out of the whole screen for their raw correlations, so the partial correlation needs a correction for that selection. Computed for every benchmark that shares at least ten models with the catalogue and eight with the index, METR's partial correlation is the largest positive one (the next are 0.51 on ten models and 0.46 on fifty). Its unadjusted \(p\)-value, from the \(t\) approximation with 21 degrees of freedom, is \(4.9\times10^{-5}\), and it stays below 0.004 after Bonferroni adjustment for every benchmark screened. Within the Claude Opus line the horizon rose from 100 to 719 minutes across four releases while CC stayed near 600, and the later Opus releases whose CC fell to 133--179 are not on the leaderboard, so the line itself does not test the relation. The choice of configuration does not drive it: matching the counting configuration to the leaderboard's setting (the reasoning-on capacity for 16K-budget and high-effort rows, the immediate-answer capacity otherwise) gives \(\rho=0.79\), a partial correlation of 0.75, and a gain of 0.09, and taking each model's best-horizon row regardless of setting gives 0.77, 0.71, and 0.07, with the bootstrap interval including zero in every case. Table~\ref{tab:metr_matched} lists the matched models and configurations: entries are matched by the leaderboard's model version string, dated checkpoints to the dated catalogue entry, at every leaderboard setting except max (the medium or default row where several are listed), and where a model has both reasoning configurations the larger capacity is used. Five of the 27 matched capacities, four of them among the 24 with an index score, are budget lower bounds entered at their bound; restricting to resolved capacities leaves 22 models with \(\rho=0.79\) and 20 with an index, on which the partial correlation is 0.71.

\begin{table}[htbp]
\centering
\footnotesize
\caption{\textbf{Models matched between the CC catalogue and the METR time-horizon leaderboard.} Entries are matched by the leaderboard's model version string, with dated checkpoints matched to the dated catalogue entry; the configuration column is the leaderboard's setting for the matched row, and where the leaderboard lists several settings for one model the medium or default row is used. The CC is the larger of the model's two reasoning configurations. The Epoch Capabilities Index (ECI) is given to one decimal; the correlations in the text use unrounded values. Vendor folds for the held-out prediction: OpenAI 11, Anthropic 6, DeepSeek 4, Google 2, Qwen 1.}
\label{tab:metr_matched}
\begin{tabular}{llrrr}
\toprule
Model & Configuration & CC & Horizon (min) & ECI \\
\midrule
\texttt{claude-opus-4} & unknown & 627 & 100.4 & 143.0 \\
\texttt{claude-opus-4.6} & unknown & 622 & 718.8 & 155.0 \\
\texttt{claude-opus-4.5} & default & 620 & 293.0 & 150.0 \\
\texttt{claude-opus-4.1} & default & 580 & 100.5 & 144.4 \\
\texttt{gemini-3.1-pro-preview} & default & 518 & 384.1 & 154.6 \\
\texttt{claude-sonnet-4} & 16K budget & 382 & 74.9 & 142.3 \\
\texttt{claude-sonnet-4.5} & 16K budget & 364 & 121.9 & 146.9 \\
\texttt{gpt-5.3-codex} & default & 223 & 349.5 & 156.2 \\
\texttt{gpt-5.4} & unknown & 208 & 341.7 & 156.5 \\
\texttt{gpt-5.2} & high & 152 & 352.2 & 153.3 \\
\texttt{gemini-2.5-pro-preview} & default & 120 & 38.7 & 145.8 \\
\texttt{o3} & medium & 103 & 91.3 & 147.1 \\
\texttt{gpt-5} & medium & 101 & 137.3 & 150.0 \\
\texttt{gpt-oss-120b} & default & 101 & 42.0 & 140.7 \\
\texttt{deepseek-r1-0528} & default & 71 & 31.2 & 142.0 \\
\texttt{kimi-k2-thinking} & default & 70 & 54.2 & -- \\
\texttt{deepseek-r1} & default & 67 & 26.9 & 139.7 \\
\texttt{o1} & medium & 66 & 38.8 & 142.7 \\
\texttt{o4-mini} & medium & 50 & 76.5 & 146.4 \\
\texttt{gpt-4o-2024-08-06} & default & 48 & 7.0 & 129.1 \\
\texttt{gpt-5.1-codex-max} & default & 43 & 223.7 & -- \\
\texttt{gpt-3.5-turbo-instruct} & default & 36 & 0.6 & -- \\
\texttt{gpt-4o-2024-11-20} & default & 34 & 9.2 & 129.4 \\
\texttt{deepseek-chat} & default & 30 & 18.5 & 146.5 \\
\texttt{deepseek-chat-v3-0324} & default & 30 & 23.1 & 137.1 \\
\texttt{qwen-2.5-72b-instruct} & default & 27 & 5.2 & 129.1 \\
\texttt{gpt-4-turbo} & default & 25 & 6.6 & 127.6 \\
\bottomrule
\end{tabular}
\end{table}

\begin{figure}[htbp]
\centering
\includegraphics[width=1.0\textwidth]{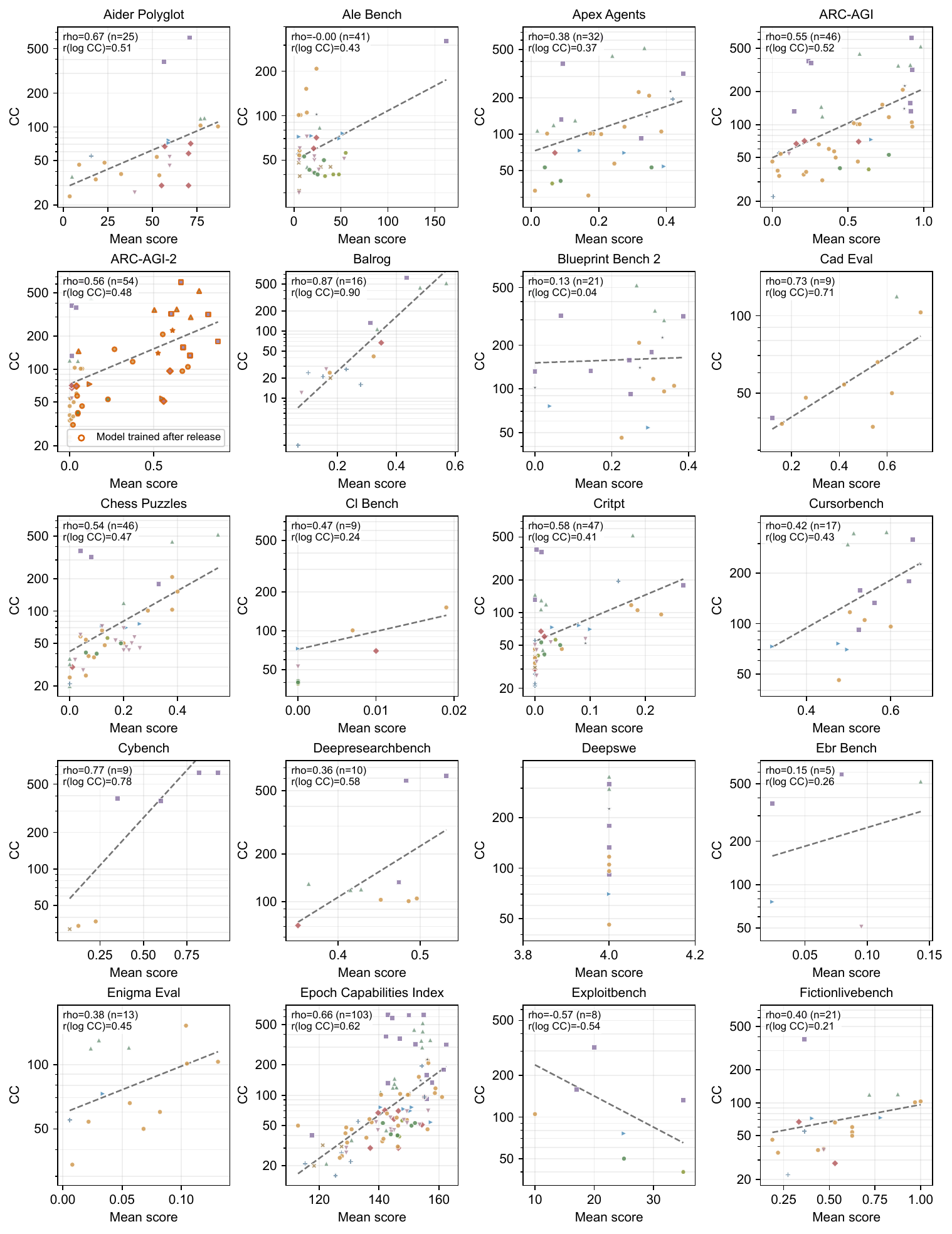}
\caption{\textbf{Comprehensive benchmark correlation panels I.} Pairwise correlation mappings between counting capacity (CC) and established task-based evaluation leaderboards, showing independent capability axes.}
\label{fig:benchmark_corr_1}
\end{figure}

\begin{figure}[htbp]
\centering
\includegraphics[width=1.0\textwidth]{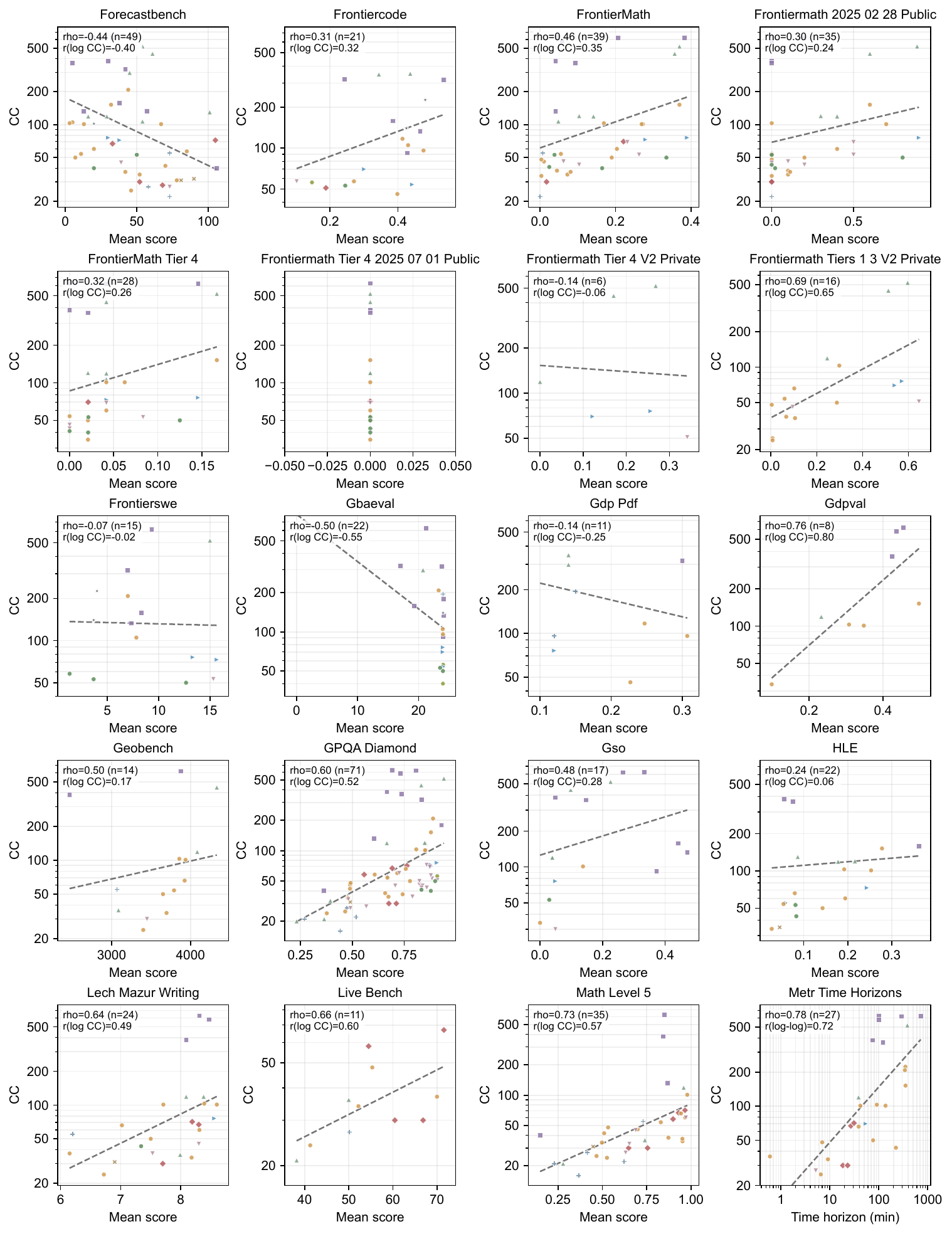}
\caption{\textbf{Comprehensive benchmark correlation panels II.} Additional pairwise correlation mappings.}
\label{fig:benchmark_corr_2}
\end{figure}

\begin{figure}[htbp]
\centering
\includegraphics[width=1.0\textwidth]{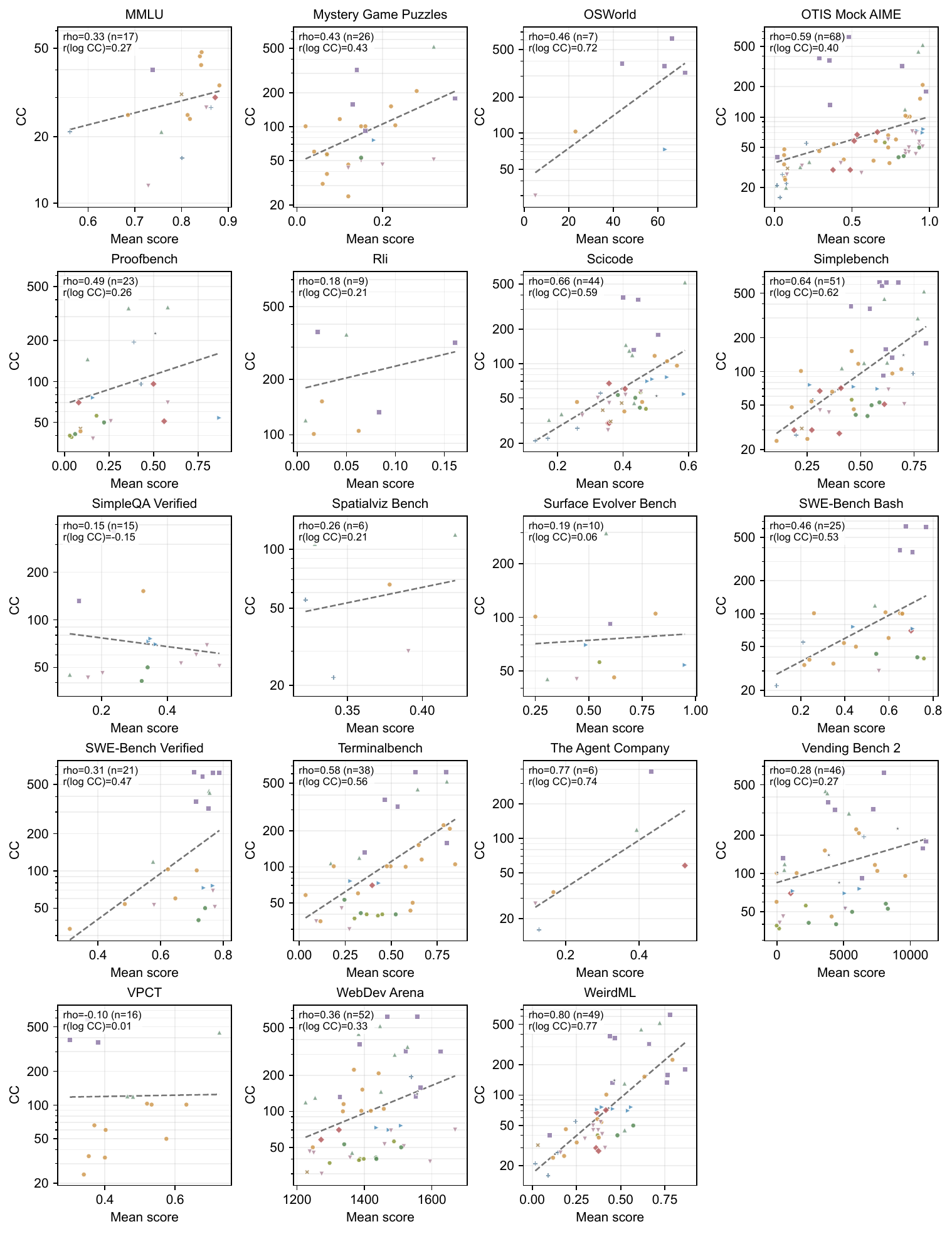}
\caption{\textbf{Comprehensive benchmark correlation panels III.} Additional pairwise correlation mappings.}
\label{fig:benchmark_corr_3}
\end{figure}

 Parameter counts are unavailable for most of the METR set, which is proprietary, so scale cannot be conditioned on. Panels are drawn only for benchmarks sharing at least five models with the evaluation set, and the number of shared models is printed in each panel. The complete set of pairwise correlation panels, extending those shown in the main text (Section~\ref{sec:evaluations}), is collected in Figures~\ref{fig:benchmark_corr_1} to~\ref{fig:benchmark_corr_3} at the end of this appendix.

\section{State-tracking capacity on a fixed model subset}\label{app:tracking}

This appendix reports a prespecified test of whether a state-maintenance task with no count in it fails at a length that follows counting capacity across models, and whether native reasoning loosens that coupling. The plan, the predictions, and the analysis were fixed before any scan ran.

\paragraph{Task.} Four named objects (red, blue, green, yellow) start at four named locations (A to D); a history of \(T\) swaps follows, one per line (``Swap the objects in B and C.''); the question is the location of an object the last swap moved, answered as one integer label among four, the labels drawn at random per probe so that no location has a fixed name. The system prompt is the catalogue's (``Return only one integer''), decoding is the catalogue's (temperature zero unless the model fixes it, 8,192 output tokens), and reasoning is switched on or off through the roster's setting exactly as for CC. Every model sees the same histories: the generator is seeded by length and probe index only.

\paragraph{Schedule and estimator.} The scan is coarse to fine. It begins with two probes at \(T=100\); the step doubles while both are exact and halves when either is not, until the bracket between the last exact and the first inexact length is tight; then a dense scan of about 24 points across the widened bracket, one probe per length, ascending, with the catalogue's stop rule (stop when the two-regime fit declares a transition whose broken regime is established, the Wilson upper bound of the accuracy above it below one half). The dense scan extends upward while the record is still exact at its top or the tail is unresolved, and downward when no stable regime has been seen; a final floor sweep adds two probes at every length from 1 to 8 to every run, so that the stable regime is anchored at short histories. Every probe enters the record, and the tracking capacity TC is the two-regime maximum-likelihood changepoint of Appendix~\ref{app:estimation} with its calibrated interval. Chance is one in four rather than zero, so the broken regime sits near 0.25 instead of near zero.

\paragraph{Scoring.} A reply is exact when its last integer is the expected label. Some models return the location letter instead of the label despite the instruction; a reply that is a single location letter, or whose last line is one, is scored against the expected location and marked as read by letter. A narrated reply is scored by its last integer, as in the counting protocol, so narration that ends on the right label counts and narration that does not counts as inexact. Replies that are API errors count as inexact and are attributed to the budget, as for CC; runs in which every probe returned an error (the o-series with reasoning off, which the catalogue never measured either, and Claude Opus 4, which the router failed on during the scan) are excluded.

\paragraph{Side questions.} At 0.5, 1.0, and 1.5 times the fitted boundary, five probes each are added: the event question (which two locations the last swap named, two options) with the full history present, and the supplied-state question (the same four-label location question with the final state stated and no history), which tests the decision with no tracking involved.

\paragraph{Subset.} The rule was fixed in advance: every model with both a METR horizon and an Epoch index score in Table~\ref{tab:metr_matched} (24 models), the six open-weight probe-suite models, and the five resolved catalogue models with the lowest CC that are still served; every hybrid among them runs in both reasoning modes. The subset is skewed toward recent models, as the METR leaderboard is. Table~\ref{tab:tracking} lists every run.

\begin{figure}[!htbp]
\centering
\includegraphics[width=0.55\textwidth]{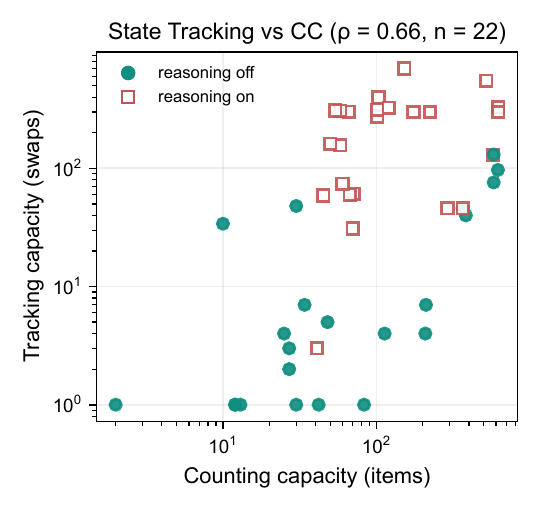}
\caption{\textbf{Tracking capacity against counting capacity.} Each point is one model and configuration: filled, immediate answering; open, native reasoning. Bounded runs are drawn at their bound.}
\label{fig:tracking}
\end{figure}

\paragraph{Results.} Forty-six configurations of 33 models completed (Table~\ref{tab:tracking}, Fig.~\ref{fig:tracking}); the mode of every run was verified from its own replies, and no run contradicted its roster setting. Twenty runs resolve a transition with its tail established, seven declare a transition whose tail is not established, eleven are bounded by the output budget (all with reasoning on), one is still exact at the top of its range, and seven show no stable regime at any length; the last enter the correlations at a capacity of one and are dropped in the sensitivity analysis. With reasoning off, on the 22 models whose counting capacity was measured in the same configuration, tracking capacity follows counting capacity: Spearman \(\rho=0.66\) (bootstrap 95\% interval 0.28 to 0.90), with a log-log slope of 0.75 (0.37 to 1.09), so the two capacities scale nearly proportionally; the ratio of tracking to counting capacity has a median of 0.10 with an interquartile range of 0.05 to 0.16. Dropping the seven floor runs leaves \(\rho=0.67\) (0.21 to 0.92) on 15 models, and keeping only fully resolved runs leaves 0.63 (0.09 to 0.92) on 14. The coupling is not carried by general capability: the partial correlation given the Epoch index is 0.65 on the 18 models with a score, while the index itself correlates with tracking capacity at 0.18. Total parameters are available for only 11 open-weight models in this configuration, on which the partial given size is 0.01 and size alone correlates at 0.38, too few to separate the two; the parameter comparison rests on the catalogue-wide analysis of Appendix~\ref{app:benchmark}. Predicting tracking capacity from the index with vendors held out does worse than the mean, so the incremental test used for METR is uninformative here. The association is largely between vendor families: dropping the five Anthropic models leaves \(\rho=0.31\) on 17, dropping the floor runs as well leaves \(-0.01\) on 11, and the correlation after removing each vendor's mean from both variables is 0.24; within families the samples are small (0.72 on the five Anthropic models, 0.15 on six OpenAI, 0.06 on five DeepSeek), and OpenAI's reasoning-off models track four to seven swaps across a CC range of 25 to 210. Leaving out any other vendor leaves \(\rho\) between 0.64 and 0.82. The cross-model coupling is therefore a between-family regularity that a wider within-family sample would have to confirm. Reasoning-off tracking capacity correlates with the METR horizon at 0.44 on the 14 models shared with the leaderboard (0.39 with reasoning on, 17 models). With reasoning on (24 configurations) the correlation with the same models' reasoning-on counting capacity is 0.31 (\(-0.13\) to 0.66). On the 13 hybrids measured in both modes, native reasoning raises tracking capacity in 12, by a median factor of 4.6 (bootstrap interval 2.5 to 59), while raising counting capacity by 1.07. The reasoning-on runs are not all settled: five are bounded by the output budget and one is still exact at its ceiling, so their entries are lower bounds; five declared a transition whose tail was not established, so their entries are uncertain rather than bounds; two are fully resolved. On the eight hybrids whose reasoning-on run is resolved or right-censored the median gain is 3.0, and on the one hybrid fully resolved in both modes, \texttt{claude-sonnet-4.5}, it is 1.0 (1.1 for \texttt{claude-sonnet-4}, whose reasoning-off tail is unestablished), the immediate capacity already matching the reasoning one. On these 13 hybrids the correlation with counting capacity falls from 0.81 with reasoning off to 0.43 with it on, but the interval on that change (\(-1.08\) to 0.24) includes zero. The history is a written record of the state, so a chain of thought replaces the silent capacity, as the second prediction stated. At and beyond each boundary the last swap is reported correctly in 0.89 of probes and the supplied-state question is answered in 0.96, while tracking accuracy above the boundary is 0.20, at the chance level of one in four: the failure is in the state update, with the text readable and the decision intact. The seven runs with no stable regime are all reasoning-off (\texttt{deepseek-chat}, \texttt{deepseek-v3.1-terminus}, \texttt{deepseek-v3.2}, \texttt{gemma-4-26b}, \texttt{llama-3.2-1b}, \texttt{qwen-2.5-7b}, \texttt{qwen3.6-35b}); \texttt{qwen3.6-35b} answers with the object's initial location in 9 of 15 probes and never applies a swap. One discrepancy is unresolved: \texttt{gemma-4-26b} with reasoning off shows no stable regime here, whereas the same task run locally through the model's own checkpoint in Appendix~\ref{app:boundary} tracks one swap perfectly and two at 0.88 with four objects; the prompts, labels, and decoding settings are the same, so the difference lies in the served route or in the extra floor probes at lengths 5 to 8, which the local grid did not include, and it is reported rather than resolved.

\begin{table}[htbp]
\centering
\footnotesize
\setlength{\tabcolsep}{4pt}
\caption{\textbf{State-tracking capacity against counting capacity.} Tracking capacity TC (four objects, swaps until the location of a moved object is no longer the majority answer) and CC for the same model and the same configuration; the event-access and supplied-state rates at the fitted boundary (five probes each). Reasoning is the verified setting: on when the run's replies carried reasoning tokens in more than half of the scan probes. Marks: \(b\), bounded by the output budget, entered at the bound; \(u\), transition declared but its tail not established; \(c\), still exact at the top of the probed range; 0, no stable regime at any length, entered at one in the correlations and dropped in the sensitivity analysis. The subset is fixed by rule (every model with a METR horizon and an index score, the six probe-suite models, and the five lowest-capacity served models) and is skewed toward recent models, as the METR leaderboard is.}
\label{tab:tracking}
\begin{tabular}{llrrrrr}
\toprule
Model & Reasoning & CC & TC & TC interval & Event at TC & Supplied at TC \\
\midrule
\texttt{claude-opus-4.6} & on & 622 & 300$^{b}$ & [83, 300] & 5/5 & 5/5 \\
\texttt{claude-opus-4.5} & on & 620 & 328$^{b}$ & [328, 328] & 5/5 & 5/5 \\
\texttt{claude-opus-4.6} & off & 618 & 97 & [93, 99] & 5/5 & 5/5 \\
\texttt{claude-opus-4.1} & off & 580 & 76 & [76, 76] & 5/5 & 5/5 \\
\texttt{claude-opus-4.5} & off & 580 & 131 & [117, 131] & 5/5 & 5/5 \\
\texttt{claude-opus-4.1} & on & 573 & 129$^{u}$ & [129, 200] & -- & -- \\
\texttt{gemini-3.1-pro-preview} & on & 518 & 550 & [508, 550] & 5/5 & 5/5 \\
\texttt{claude-sonnet-4} & off & 382 & 40$^{u}$ & [37, 50] & -- & -- \\
\texttt{claude-sonnet-4.5} & off & 364 & 46 & [46, 46] & 5/5 & 5/5 \\
\texttt{claude-sonnet-4.5} & on & 364 & 46 & [46, 46] & 5/5 & 5/5 \\
\texttt{claude-sonnet-4} & on & 290 & 46 & [46, 46] & 5/5 & 5/5 \\
\texttt{gpt-5.3-codex} & on & 223 & 300$^{u}$ & [100, 1040] & -- & -- \\
\texttt{gpt-5.3-codex} & off & 210 & 7 & [1, 9] & -- & -- \\
\texttt{gpt-5.4} & off & 208 & 4 & [1, 6] & -- & -- \\
\texttt{gpt-5.4} & on & 174 & 300$^{u}$ & [8, 950] & -- & -- \\
\texttt{gpt-5.2} & on & 152 & 700$^{b}$ & [700, 750] & 5/5 & 5/5 \\
\texttt{gemini-2.5-pro-preview} & on & 120 & 325 & [302, 325] & 5/5 & 5/5 \\
\texttt{gpt-5.2} & off & 113 & 4 & [1, 8] & -- & -- \\
\texttt{o3} & on & 103 & 400$^{b}$ & [379, 700] & -- & -- \\
\texttt{gpt-5} & on & 101 & 275$^{b}$ & [256, 300] & 5/5 & 5/5 \\
\texttt{gpt-oss-120b} & on & 101 & 313$^{b}$ & [100, 324] & -- & -- \\
\texttt{gemma-4-26b-a4b-it} & off & 83 & 0$^{0}$ & no stable regime & -- & -- \\
\texttt{deepseek-r1-0528} & on & 71 & 61$^{b}$ & [56, 82] & 4/5 & 4/5 \\
\texttt{deepseek-v3.2} & on & 70 & 31$^{u}$ & [31, 50] & -- & -- \\
\texttt{deepseek-r1} & on & 67 & 60$^{b}$ & [48, 100] & -- & -- \\
\texttt{o1} & on & 66 & 302$^{b}$ & [247, 313] & 5/5 & 4/5 \\
\texttt{deepseek-v3.1-terminus} & on & 60 & 74$^{c}$ & [74, 100] & 1/5 & 4/5 \\
\texttt{deepseek-v3.2-exp} & on & 58 & 157$^{b}$ & [7, 423] & -- & -- \\
\texttt{gpt-oss-20b} & on & 58 & 307 & [111, 353] & 2/5 & 5/5 \\
\texttt{muse-glimmer-30b} & on & 54 & 309$^{u}$ & [309, 471] & -- & -- \\
\texttt{o4-mini} & on & 50 & 162 & [112, 162] & 5/5 & 5/5 \\
\texttt{gpt-4o-2024-08-06} & off & 48 & 5 & [3, 5] & -- & -- \\
\texttt{qwen3.6-35b-a3b} & on & 45 & 59$^{u}$ & [8, 80] & -- & -- \\
\texttt{qwen3.6-35b-a3b} & off & 42 & 0$^{0}$ & no stable regime & 5/5 & 5/5 \\
\texttt{gemma-4-26b-a4b-it} & on & 41 & 3$^{b}$ & [2, 25] & -- & -- \\
\texttt{gpt-4o-2024-11-20} & off & 34 & 7 & [2, 19] & -- & -- \\
\texttt{deepseek-chat} & off & 30 & 0$^{0}$ & no stable regime & -- & -- \\
\texttt{deepseek-chat-v3-0324} & off & 30 & 48 & [41, 52] & 5/5 & 5/5 \\
\texttt{llama-3.3-70b-instruct} & off & 27 & 3 & [2, 3] & 5/5 & 5/5 \\
\texttt{qwen-2.5-72b-instruct} & off & 27 & 2 & [1, 2] & -- & -- \\
\texttt{gpt-4-turbo} & off & 25 & 4 & [2, 5] & -- & -- \\
\texttt{deepseek-v3.1-terminus} & off & 13 & 0$^{0}$ & no stable regime & 4/5 & 4/5 \\
\texttt{deepseek-v3.2} & off & 12 & 0$^{0}$ & no stable regime & 3/5 & 5/5 \\
\texttt{qwen-2.5-7b-instruct} & off & 12 & 0$^{0}$ & no stable regime & 3/5 & 3/5 \\
\texttt{deepseek-v3.2-exp} & off & 10 & 34 & [34, 34] & 5/5 & 4/5 \\
\texttt{llama-3.2-1b-instruct} & off & 2 & 0$^{0}$ & no stable regime & -- & -- \\
\bottomrule
\end{tabular}
\end{table}

\FloatBarrier

\end{document}